%% file: main.tex
\pdfoutput=1
\documentclass{article}

\usepackage[main, final]{neurips_2026}

\usepackage{times}
\usepackage{amsmath,amssymb,amsfonts}
\usepackage{textcomp}
\usepackage{graphicx}
\usepackage{booktabs}
\usepackage{multirow}
\usepackage{algorithm}
\usepackage{algpseudocode}
\usepackage{placeins}
\usepackage{hyperref}
\usepackage{url}
\usepackage{bm}
\usepackage{microtype}
\usepackage{xcolor}
\usepackage{adjustbox}
\usepackage{amsthm}
\newtheorem{remark}{Remark}

\makeatletter
\renewcommand{\paragraph}{%
  \@startsection{paragraph}{4}{\z@}%
  {1.2ex \@plus.5ex \@minus.2ex}%
  {-1em}%
  {\normalfont\normalsize\bfseries}}
\makeatother
\usepackage{kotex}

\newcommand{\fpotk}[1]{FPoT$_{k{=}#1}$}

\begin{document}

\title{Fractional Power-of-Two Quantization for\\Efficient and Effective Multiplier-Free LLM Inference}

\author{
  Sunghyun Wee \\
  Seoul National University \\ LG Electronics \\
  \texttt{wsh05@snu.ac.kr} 
  \And
  Geunjae Choi\thanks{Equal contribution} \\
  LG Electronics \\
  \texttt{cgj.choi@lge.com}
  \And
  Hyeonjin Kim\footnotemark[1] \\
  Seoul National University \\
  \texttt{peaceful1@snu.ac.kr}
  \And
  Suyoung Kim\footnotemark[1] \\
  Seoul National University \\
  \texttt{ksyo96@snu.ac.kr}
  \And
  Nojun Kwak\thanks{Corresponding author} \\
  Seoul National University \\
  \texttt{nojunk@snu.ac.kr}
}

\maketitle
\setcounter{footnote}{0}

\begin{abstract}
Large Language Models (LLMs) demand efficient inference, where the multiply-accumulate (MAC) operations in linear projections dominate compute and energy cost. Post-training quantization (PTQ) reduces this cost by mapping weights and activations to low-bit integers. Power-of-Two (PoT) weight quantization promises efficient multiplier-free LLM inference via shift-add logic, but suffers significant accuracy degradation at low bit-widths due to its coarse base-2 quantization grid. We introduce \textbf{Fractional PoT (FPoT)}, a base-$\sqrt{2}$ (half-octave) fractional PoT grid that achieves practical 4-bit weight quantization while preserving a fully integer shift-add datapath. The method combines (i)~the base-$\sqrt{2}$ weight grid with uniform integer activations, (ii)~a calibration-time grid-alignment refinement absorbed into the static per-channel scale, and (iii)~a lightweight 1-adder $\sqrt{2}$ approximation whose error is absorbed during weight calibration via approximate-grid quantization, yielding a processing element (PE) with significantly fewer gates than a 4-bit integer multiplier. On Llama-3-8B with 4-bit weight and activation quantization, FPoT improves zero-shot commonsense reasoning performance by $8.72\%$ relative to the PoT baseline, while maintaining performance close to the uniform integer baseline, with only a negligible $0.4\%$p degradation. These results hold across various configurations spanning seven models and three bit-widths. Furthermore, FPoT can be integrated with state-of-the-art LLM PTQ methods, demonstrating framework-agnostic applicability.
\end{abstract}

\input{sections/01_introduction}
\input{sections/02_related_work}
\input{sections/03_methodology}
\input{sections/04_analysis}
\input{sections/05_experimental_results}
\input{sections/06_discussion}
\input{sections/07_conclusion}

\FloatBarrier

\begin{ack}
This work was supported by the Korean Government through grants from IITP (RS-2021-II211343 and RS-2025-25442338) and NRF (RS-2026-25500579).
The authors declare no competing interests.
\end{ack}

\bibliographystyle{plainnat}
\bibliography{references}

\newpage
\appendix
\input{sections/A_appendix_preliminaries}          %
\FloatBarrier
\input{sections/B_appendix_grid_matching_theory}    %
\FloatBarrier
\input{sections/C_appendix_empirical_bin_occupancy} %
\FloatBarrier
\input{sections/D_appendix_per_projection_sensitivity_analysis} %
\FloatBarrier
\input{sections/E_appendix_hardware_efficiency_analysis_details} %
\FloatBarrier
\input{sections/F_appendix_fused_triton_kernel}     %
\FloatBarrier
\input{sections/G_appendix_additional_ablations}    %
\input{sections/G_appendix_additional_ablations_hyperparams} %
\FloatBarrier
\input{sections/H_appendix_per_task_downstream_results} %
\FloatBarrier
\input{sections/I_appendix_logart_comparison}       %
\FloatBarrier
\input{sections/J_appendix_cross_framework_portability} %
\FloatBarrier
\input{sections/K_appendix_factorial}             %
\FloatBarrier
\input{sections/L_appendix_generation}            %
\FloatBarrier

\end{document}

%% file: sections/01_introduction.tex
\section{Introduction}

The rapid proliferation of Large Language Models (LLMs)~\citep{touvron2023llama2,llama3, qwen25, jiang2023mistral7b} has transformed natural language processing, yet their deployment on resource-constrained edge devices remains challenging.
One of the dominant costs in LLM inference is the multiply-accumulate (MAC) operation within Transformer~\citep{Vaswani2017} linear projections, which account for virtually all parameters and a majority of inference energy~\citep{wang2020linformer, sze2017efficientprocessingdeepneural}. 
Recent post-training quantization (PTQ) ~\citep{ashkboos2024quarot, liu2024spinquant, ostquant2025, duquant2024} has pushed integer (INT) quantization to 4-bit weights, activations, and key-value (KV) cache at near-lossless accuracy, making 4-bit the practical operating point: 8-bit pays with costly multipliers, while lower bits collapse LLM accuracy.

Power-of-Two (PoT) quantization, also known as logarithmic quantization~\citep{elhoushi2021deepshift, li2020apot}, offers a multiplier-free pathway and further reduces computation: constraining weights to $\pm 2^n$ allows MAC operations to be substituted with bit-wise shifts and an integer addition, substantially reducing the per-MAC energy. Register-Transfer Level (RTL) synthesis on Samsung 5\,nm shows a PoT weight MAC at $0.42{\times}$ the energy of an equivalent INT MAC~\citep{przewlocka2022pot}.

Though efficient, accuracy remains the central obstacle of PoT quantization. Standard PoT places representable levels at geometric intervals of factor~2 (one octave), yielding a worst-case rounding error of $41.4\%$ ($2^{1/2}-1$) at the midpoint under a 4-bit round-to-nearest strategy. While this coarseness is tolerable for Convolutional Neural Networks (CNNs)~\citep{elhoushi2021deepshift,li2020apot,li2023denseshift}, it causes catastrophic degradation in LLMs: perplexity (PPL) degradation of $+9.5$ to $+107.5$ when PoT is applied to both weights and activations, or $+1.43$ when restricted to weights alone. \emph{Standard PoT is not a viable drop-in replacement for multiplication operations in LLMs.}

In this work, we show that the accuracy barriers of 4-bit PoT stem from a grid mismatch with the LLM weight statistics, and introduce \textbf{Fractional PoT (FPoT)} to address this. On Llama-3-8B at 4-bit, applying FPoT to weights reaches on-par performance with the uniform INT baseline (only $0.4\%$p 0-shot degradation), while delivering an $8.72\%$ relative gain over standard PoT.
Throughout this paper, we use \emph{near-lossless} to denote configurations whose perplexity increase remains within $+0.5$ and whose downstream accuracy drop remains within $1.0$\%p of the uniform INT baseline.

Our contributions are as follows.
\begin{itemize}
    \item \textbf{Theoretically grounded base-$\sqrt{2}$ grid for 4-bit LLMs}: We conduct a Gaussian-proxy analysis to identify base-$\sqrt{2}$ as the optimal integer-resolution PoT grid at 4-bit, halving the worst-case relative error of standard PoT (41.4\%${\rightarrow}$18.9\%).
    \item \textbf{Approximate-grid quantization with low computational overhead}: FPoT quantizes weights to the actual hardware grid using a 1-adder $\sqrt{2}$ approximation. The resulting processing element (PE) requires significantly fewer gates than a 4-bit integer multiplier.
    \item \textbf{Near-lossless performance over uniform INT baselines}: FPoT integrates with state-of-the-art LLM PTQ frameworks at comparable overhead (Appendix~\ref{app:duquant}), achieving performance on par with uniform INT baselines. 
\end{itemize}

%% file: sections/02_related_work.tex
\section{Related Work}
\label{sec:rw}

\subsection{Post-Training Quantization (PTQ) for LLMs}

Uniform INT weight quantization methods such as GPTQ~\citep{frantar2022gptq} and AWQ~\citep{lin2023awq} achieve competitive accuracy at 4-bit via Hessian-weighted error compensation and activation-aware weight scaling.
Based on these works, a line of research on quantizing activations to low bit-widths emerged for more efficient calculation.
However, activation quantization is more challenging due to the high-magnitude outliers concentrated in a small subset of channels~\citep{dettmers2022llmint8}. Extreme outliers cause uniform INT4 activation quantization to collapse most activations to zero under a per-tensor scale dominated by the outlier channels. Gradually, attempts to mitigate such outliers emerged. Early approaches such as SmoothQuant~\citep{xiao2023smoothquant} migrate the quantization difficulty from activations to weights via per-channel scaling. Recent methods resolve this via rotation-based preprocessing that redistributes outliers across dimensions: QuaRot~\citep{ashkboos2024quarot} applies randomized Hadamard rotations, SpinQuant~\citep{liu2024spinquant} learns globally optimal rotations, OSTQuant~\citep{ostquant2025} jointly trains rotation with per-channel smooth scales for tighter control, and DuQuant~\citep{duquant2024} adds dual block-diagonal transforms and channel permutation. These approaches collectively enable 4-bit weights, 4-bit activations, 4-bit KV cache~(W4A4KV4) uniform quantization at near-lossless accuracy.

While the aforementioned methods enable practical 4-bit activation quantization, they mainly focus on uniform quantization grids. In this work,
we apply logarithmic weight quantization to the state-of-the-art quantization methods and demonstrate that it can reach comparable performance with better efficiency.

\subsection{Power-of-Two (PoT) and Shift-based Quantization}

PoT quantization~\citep{elhoushi2021deepshift} is a family of methods that constrains weights to $\pm 2^{n}$ values and replaces multipliers with barrel shifters. Prior works on CNN include DeepShift~\citep{elhoushi2021deepshift}, APoT~\citep{li2020apot}, and DenseShift~\citep{li2023denseshift}.
These methods are either not directly applicable to LLMs, or represent weights as sums of multiple PoT terms, requiring $n$ sequential shift-adds per MAC.
For LLMs, PoT-PTQ~\citep{potpourri2025} is a base-2 quantization that achieves 2--3-bit via data-agnostic scale initialization plus residual fine-tuning. Concurrent with our work, LogART~\citep{logart2026} extends logarithmic PTQ with learnable rounding and dynamic base allocation between base-$2$ and base-$\sqrt{2}$, realizing $\sqrt{2}$ with the shift-add approximation $\sqrt{2} \approx 2^0 + 2^{-1}$ of earlier logarithmic accelerators~\citep{xu2018multibase, xu2023aslog}. However, both methods validate their LLM accuracy only with floating-point (FP) activations at 16-bit precision (FP16). 

In that setting, the main operation is $\texttt{FP16} \times \texttt{PoT}$, requiring FP16 accumulators that cannot be simplified to an integer shift-add unit.
Our FPoT pairs a fixed half-octave grid with uniform INT activations, enabling a fully integer shift-add PE.%

\paragraph{Position of this work.}
Between the two distinct lines of work---activation quantization, which enables fully INT-only arithmetic, and PoT quantization, which replaces multipliers with shift-add PE---a gap remains.
Our method, FPoT, aims to address this gap: a multiplier-free LLM quantization method that retains uniform INT activations and composes with recent PTQ frameworks. State-of-the-art uniform INT quantization methods~\citep{ashkboos2024quarot, liu2024spinquant, ostquant2025, duquant2024} achieve near-lossless W4A4 but retain integer multipliers. PoT-based methods are either CNN-targeted~\citep{elhoushi2021deepshift, li2020apot, li2023denseshift}, or validate LLM accuracy only with FP activations~\citep{potpourri2025, logart2026}. FPoT can pair base-$\sqrt{2}$ weight quantization with state-of-the-art uniform INT activation quantization to achieve efficient and effective W4A4 quantization. $\sqrt{2}$ is approximated with a single adder, where the approximation cost is absorbed via approximate-grid quantization (\S\ref{sec:approx-grid}). FPoT realizes a fully integer shift-add PE at near-lossless W4A4KV4 accuracy.

%% file: sections/03_methodology.tex
\section{Methodology}
\label{sec:method}

Our method applies base-$\sqrt{2}$ quantization to all linear projection weights of each Transformer block while maintaining uniform INT quantization for activations and the key-value (KV) cache. This design isolates the logarithmic grid to static weights, where calibration can absorb quantization error, while preserving hardware-friendly uniform INT formats for dynamic activations and KV cache. Weight distributions are redistributed by state-of-the-art LLM PTQ preprocessing (Appendix~\ref{app:prelim}) prior to GPTQ calibration, and a calibration-time weight grid-alignment further refines the per-channel scale. In brief, such methods apply learned rotations and per-channel scaling to redistribute channel-concentrated outliers, producing smoother distributions amenable to the logarithmic grid. The transform adds negligible overhead (${\approx}0.02\%$ of weight MACs).

\paragraph{Grid resolution and formulation.}
\label{sec:fpot-formulation}
Standard PoT quantization places representable levels at geometric intervals of factor~$2$ (one octave), yielding a worst-case relative quantization error of $2^{1/2} - 1 \approx 41.4\%$ at the midpoint between adjacent levels. This coarse resolution is a fundamental accuracy bottleneck for LLM weight quantization using PoT.

We introduce a \emph{fractional exponent resolution parameter} $k \in \mathbb{Z}_{>0}$ (FPoT), an extension of the PoT grid. For a per-channel scale $s > 0$ and $n$ levels per sign, the FPoT quantized value is:
\begin{equation}
  Q_{\mathrm{FPoT}}(w;\,s,n,k) = \mathrm{sgn}(w) \cdot s \cdot 2^{\,\mathrm{clip}\!\left(\left\lfloor k \cdot \log_2\tfrac{|w|}{s} \right\rceil,\,0,\,n{-}1\right) / k},
  \label{eq:fpot}
\end{equation}
where $\lfloor \cdot \rceil$ denotes rounding to the nearest integer, and the exponent index is clipped to $[0,\, n{-}1]$ before division by $k$. 
For $k{=}1$, Eq.~\eqref{eq:fpot} reduces to standard PoT. For $k{=}2$ (half-octave grid), representable values are spaced at intervals of $2^{1/2} \approx 1.414$, reducing the worst-case relative error to $2^{1/4}-1 \approx 18.9\%$---roughly a $2.2{\times}$ improvement in grid resolution. We analyze the optimality of $k{=}2$ for 4-bit LLM weight quantization in \S\ref{sec:grid-matching}.

\paragraph{Signed quantization with $k$-adaptive zero threshold.}
\label{sec:fpot-zero}

The zero threshold follows naturally from the grid geometry. For a FPoT grid with minimum exponent index $e{=}0$, the log-domain decision boundary between ``round to $e{=}0$'' and ``round to zero'' yields:
\begin{equation}
    \tau_k = 2^{-1/(2k)},
    \label{eq:zero-thresh}
\end{equation}
so that $|w|/s < \tau_k$ maps to zero. At $k{=}2$, $\tau_2 \approx 0.841$; the resulting fraction of weights mapped to zero and its accuracy implications are analyzed in \S\ref{sec:grid-matching}.

The complete FPoT quantization procedure, including signed representation, $k$-adaptive zero thresholding, and the inference-time hardware decomposition, is given in the pseudocode in Appendix~\ref{app:triton-kernel}.

\paragraph{Hardware implications.}
\label{sec:fpot-hw}

The stored weight is encoded as $b$~bits with one sign bit and a magnitude code $m \in \{0, \ldots, 2^{b-1}{-}1\}$. When $m{=}0$ the MAC is skipped (zero-gating); otherwise the exponent index $e = m{-}1$ yields the dequantized value $\hat{w} = \mathrm{sgn}(w) \cdot s \cdot 2^{e/k}$. For $k{=}2$, $\lfloor e/2 \rfloor$ and $e \bmod 2$ are direct wire selections from the magnitude bits (Appendix~\ref{app:triton-kernel}). Combining FPoT weights with uniform INT activations, the shift-accumulate MAC becomes:
\begin{equation}
    y = \sum_i \mathrm{sgn}(w_i) \cdot (x_i \cdot 2^{\lfloor e_i/2 \rfloor}) \cdot (\sqrt{2})^{e_i \bmod 2},
\end{equation}
which replaces integer multipliers with barrel shifters plus a lightweight $\times\!\sqrt{2}$ unit for odd exponents (${\sim}50\%$ of MACs). Activations remain in uniform INT format (Appendix~\ref{app:sensitivity} provides the supporting per-projection analysis). The entire FPoT compute unit shares a \textbf{single barrel shifter} with PoT; the only additional logic is a 1-adder $\sqrt{2}$ approximation unit and a multiplexer (MUX) selected by the least significant bit (LSB) of the weight (Fig.~\ref{fig:pe-arch}). This shift-accumulate structure replaces all 4-bit integer multipliers in every linear projection of each Transformer block with a uniform FPoT PE.

\input{figures/fig_pe_arch}

\paragraph{Approximate-grid quantization.}
\label{sec:approx-grid}

The odd-$e$ path above requires a $\times\!\sqrt{2}$ coefficient. A 1-adder realization ($x + x{\gg}1 = 1.5x$) deviates from $\sqrt{2} \approx 1.414$ by $6.1\%$, which would normally degrade accuracy under na\"{\i}ve post-hoc substitution---quantizing to the exact FPoT grid and only replacing the coefficient at dequantization. Our solution is \emph{approximate-grid quantization}: we calibrate weights on the \emph{actual hardware grid} (i.e., using $1.5$ for odd $e$ instead of $\sqrt{2}$), so that weight calibration absorbs the approximation as part of the quantization grid rather than as post-hoc distortion.
Empirical validation of this scheme is reported in \S\ref{sec:sqrt2-approx}, with the resulting hardware cost analyzed in \S\ref{sec:hw-analysis}.

\paragraph{Weight grid-alignment.}
\label{sec:w-grid-align}
We additionally evaluate a calibration-time grid-alignment refinement applicable to both PoT and FPoT grids. Since weights are static after calibration, the optimal phase offset can be absorbed into the per-channel scale at \textbf{zero inference overhead}. We add a phase-shift refinement evaluating $G$ candidate scales
\begin{equation}
  s_g \;=\; s_{\mathrm{base}} \cdot 2^{g/G}, \quad g = 0,\,1,\,\ldots,\,G{-}1,
  \label{eq:w-grid-align}
\end{equation}
and pick per output-channel the $g^{*}$ minimizing MSE between raw weights and their FPoT reconstruction. The winning $s_{g^{*}}$ replaces $s_{\mathrm{base}}$ as a static per-channel weight scale, so the deployed architecture is identical to unaligned FPoT.

%% file: figures/fig_pe_arch.tex
\begin{figure}[t]
    \centering
    \includegraphics[width=\linewidth]{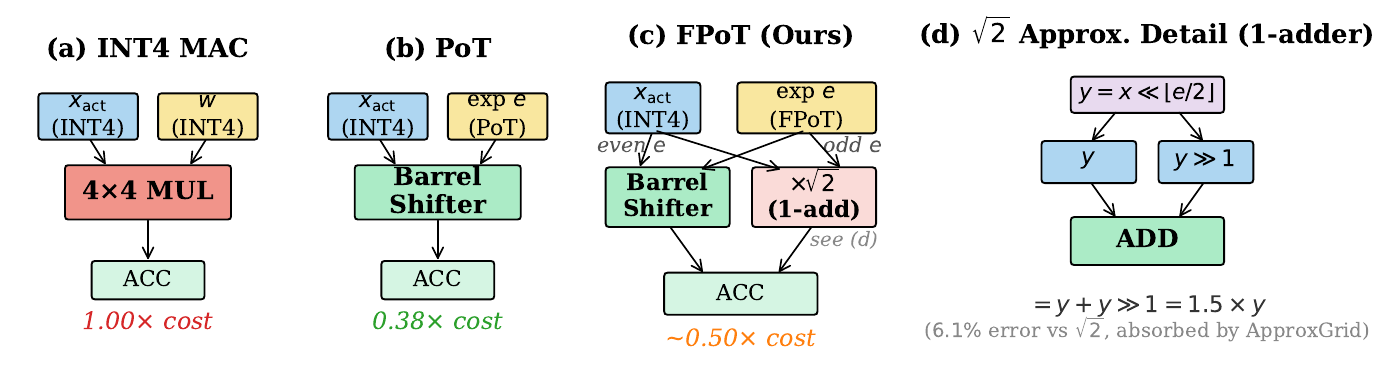}
    \caption{PE architecture: INT4 multiplier, PoT barrel shifter, and FPoT (Ours). FPoT adds one adder and a MUX; $\sqrt{2}\,y \approx y+y{\gg}1 = 1.5y$ ($6.1\%$ error, absorbed by approximate-grid quantization, \S\ref{sec:approx-grid}).}
    \label{fig:pe-arch}
\end{figure}

%% file: sections/04_analysis.tex
\section{Analysis}
\label{sec:analysis}

We analyze why the FPoT base-$\sqrt{2}$ grid is optimal for 4-bit LLM weight quantization (\S\ref{sec:grid-matching}) and quantify its hardware cost relative to INT and standard PoT baselines (\S\ref{sec:hw-analysis}).

\subsection{Grid-Matching Analysis}
\label{sec:grid-matching}

\paragraph{Gaussian proxy.}
Raw LLM weights are far from Gaussian: they exhibit heavy-tailed, leptokurtic distributions with high-magnitude outliers concentrated in a few channels. Recent LLM PTQ preprocessing \emph{redistributes} channel-concentrated outliers across dimensions, reducing kurtosis and equalizing per-channel dynamic range. We analyze a zero-mean symmetric distribution whose log-magnitude is unimodal---a property that holds for the preprocessed weights, with the Gaussian $w \sim \mathcal{N}(0,\sigma^2)$ serving as a tractable analytical proxy. Under this proxy, $u = \log_2|w|$ is approximately Gaussian in log-space with standard deviation $1/(\sqrt{2}\ln 2) \approx 1.02$ octaves (derivation in Appendix~\ref{app:log-magnitude}), so ${\sim}95\%$ of probability mass concentrates within ${\sim}4$ octaves around the mode. Empirically, the preprocessed weight distributions of all seven models studied closely match this Gaussian-proxy prediction (Appendix~\ref{app:empirical}), validating its use for 4-bit grid design.

\paragraph{Optimal base under the Gaussian proxy.}
\label{sec:fpot-motivation}
A PoT quantizer with step ratio $\beta = 2^{1/k}$ and $n$ levels per sign covers a dynamic range $\beta^{n-1}$; with $n{=}7$, $k{=}1$ yields a $64{\times}$ range that exceeds the ${\sim}4$-octave preprocessed weight spread, $k{=}2$ yields $8{\times}$ (aligned), and $k{=}4$ yields $2.8{\times}$ (tail clipping). Numerically minimizing the MSE on $w \sim \mathcal{N}(0,1)$ with 15~levels while jointly optimizing the scale $s$ yields $\beta^{\star} \approx 2^{0.55}$ (effective $k^{\star} \approx 1.8$), very close to $k{=}2$ (Fig.~\ref{fig:mse-nbe-vs-base}): \fpotk{2} achieves MSE $= 1.27{\times}$ INT4, while PoT wastes levels near zero ($3.21{\times}$) and $k{\geq}3$ suffers dynamic range starvation ($2.87{\times}$ at $k{=}3$, $5.37{\times}$ at $k{=}4$). This optimality is bit-width specific---at 8-bit the optimum shifts to $k^{\star} \approx 19$---so the $k{=}2$ choice is specifically motivated by the 4-bit constraint (8-bit analysis and full numerical table in Appendix~\ref{app:mse-nbe-table}).

\input{figures/fig_combined_mse_binoccupancy_float}

\paragraph{Normalized Bin Entropy.}
To quantify how effectively a grid utilizes its fixed bit budget under a given weight distribution, we introduce the \emph{Normalized Bin Entropy} (NBE). For a quantizer $Q$ with $N$ representable levels and level-occupancy probabilities $p_i = \Pr[Q(W) = \ell_i]$, we define
\begin{equation}
  \mathrm{NBE}(Q; p_W) \;=\; \frac{H(\mathbf{p})}{\log_2 N},
  \quad H(\mathbf{p}) = -\textstyle\sum_{i=1}^{N} p_i \log_2 p_i,
  \label{eq:nbe}
\end{equation}
where $N$ is the budget-determined number of representable levels (e.g., $N{=}15$ for the 4-bit symmetric grids studied here: $7$~positive + $7$~negative + $1$~zero). NBE $= 1$ iff data mass is uniform across all levels; NBE $\to 0$ as mass concentrates on a single level. Information-theoretically, $\log_2 N \cdot \mathrm{NBE}$ is the expected Shannon code length per weight, so NBE measures the fraction of the fixed bit budget that carries information (Appendix~\ref{app:nbe-properties}). Unlike volumetric metrics such as QSUR~\citep{ostquant2025}, NBE distinguishes grid structures sharing the same dynamic range (Property~4 in Appendix~\ref{app:nbe-properties}).

Under the Gaussian proxy, \fpotk{2} achieves $\mathrm{NBE} = 0.9364$, surpassing both uniform INT4 ($0.9026$) and PoT ($0.8773$): the half-octave log-grid places denser levels where probability mass concentrates near the mode $|w| \approx \sigma$, while PoT wastes levels in the sparsely populated tail ($64{:}1$ range); Fig.~\ref{fig:bin-occupancy-gaussian-body} visualizes this contrast. Empirical bin occupancy on real LLM weights (Appendix~\ref{sec:appendix-bin-occupancy}) closely matches the Gaussian-proxy prediction.

\paragraph{Implicit magnitude pruning.}
The $k$-adaptive zero threshold $\tau_k = 2^{-1/(2k)}$ (Eq.~\eqref{eq:zero-thresh}) interacts with the Gaussian-proxy distribution to map a $k$-dependent fraction of weights to zero. At $k{=}2$, $\tau_2 \approx 0.841$ causes ${\sim}22\%$ of preprocessed weights to be zeroed---\textbf{implicit magnitude pruning} that is fully compensated by the $2{\times}$ finer grid resolution. Empirical measurement on real LLM weights confirms this ratio (Appendix~\ref{sec:appendix-bin-occupancy}). 
These zeroed weights further skip their MACs entirely via zero-gating (\S\ref{sec:fpot-hw}), providing additional compute reduction beyond the shift-add substitution.
Prior PoT methods (DeepShift, APoT) adopt ad-hoc zero strategies without recognizing this $k$-dependent effect.

\subsection{Hardware Cost Analysis}
\label{sec:hw-analysis}

We evaluate the hardware benefits of FPoT weight quantization with two complementary metrics: compute-unit (CU) cost from a na\"{\i}ve component model and energy from published RTL synthesis data. FPoT is applied to all linear projection layers, converting all weight MACs (6.98\,B per token) into shift-accumulate operations (per-projection breakdown in Appendix~\ref{app:mac-breakdown}).

\paragraph{Compute unit cost.}
\label{sec:hw_cost}
We model hardware cost with a na\"{\i}ve component-based analysis (Appendix~\ref{app:hw_derivation}): an INT multiplier grows quadratically ($W{\times}A$ gate-equivalents), while a PoT barrel shifter and FPoT shifter\,+\,adder grow linearly. FPoT adds one adder for the $\sqrt{2}$ approximation ($x{+}x{\gg}1 = 1.5x$, 6.1\% error absorbed by approximate-grid quantization, \S\ref{sec:approx-grid}). At W4A4, PoT costs $0.38{\times}$ and FPoT costs ${\sim}0.50{\times}$ the INT multiplier; the $O(WA)$-vs.-$O(W{+}A)$ scaling gap widens at higher bit-widths (Table~\ref{tab:hw_analytical}). Critically, the additional 1-adder over standard PoT costs only ${\sim}0.12{\times}$ extra CU at W4A4 while reducing perplexity overhead by $67\%$--$94\%$, placing FPoT at the knee of the accuracy--efficiency Pareto frontier (Fig.~\ref{fig:tradeoff}) in the compute-bound prefill regime. Open-source logic synthesis (Yosys~\citep{wolf2013yosys}) is consistent with the structural ordering INT\,$>$\,FPoT\,$>$\,PoT and the bit-width scaling predicted above (Tables~\ref{tab:hw_analytical}--\ref{tab:yosys_gates}, Appendix~\ref{app:hw_yosys}). 

\paragraph{Energy.}
\label{sec:hw_energy}
We anchor the energy comparison to RTL synthesis by \citet{przewlocka2022pot} on Samsung 5\,nm (250\,MHz, TT corner, 0.6\,V, 25\textdegree C, 50\% toggle rate), which reports a PoT-to-INT energy ratio of $\mathbf{0.42{\times}}$ at W4A8---a $2.4{\times}$ reduction per weight MAC. FPoT with the 1-adder $\sqrt{2}$ unit adds minimal overhead (one adder, active for ${\sim}50\%$ of MACs); we report the FPoT energy ratio as $\mathbf{0.42\text{--}0.46{\times}}$ at W4A4 ($0.84$--$1.00{\times}$ at W4A8; $1.68$--$2.00{\times}$ at W8A8)---a $2.2$--$2.4{\times}$ energy reduction per weight multiply versus INT, with the $\sqrt{2}$ unit contributing at most $0.04{\times}$ overhead for a $50\%$ duty cycle on odd exponents. Appendix~\ref{app:hw_tradeoff} provides per-model accuracy--cost trade-off curves.

\paragraph{Scope of hardware claims.}
The $0.42$--$0.46{\times}$ energy and $0.50{\times}$ CU ratios are \emph{per weight MAC}, anchored to cell-level RTL synthesis~\citep{przewlocka2022pot}; end-to-end inference energy reduction scales with the fraction of total energy attributable to weight MACs, which is largest in the compute-bound prefill regime and diminishes under memory-bound decode (Limitations, \S\ref{sec:discussion}).

\input{figures/fig_tradeoff_float}

%% file: figures/fig_combined_mse_binoccupancy_float.tex
\begin{figure*}[t]
\centering
\includegraphics[width=\linewidth]{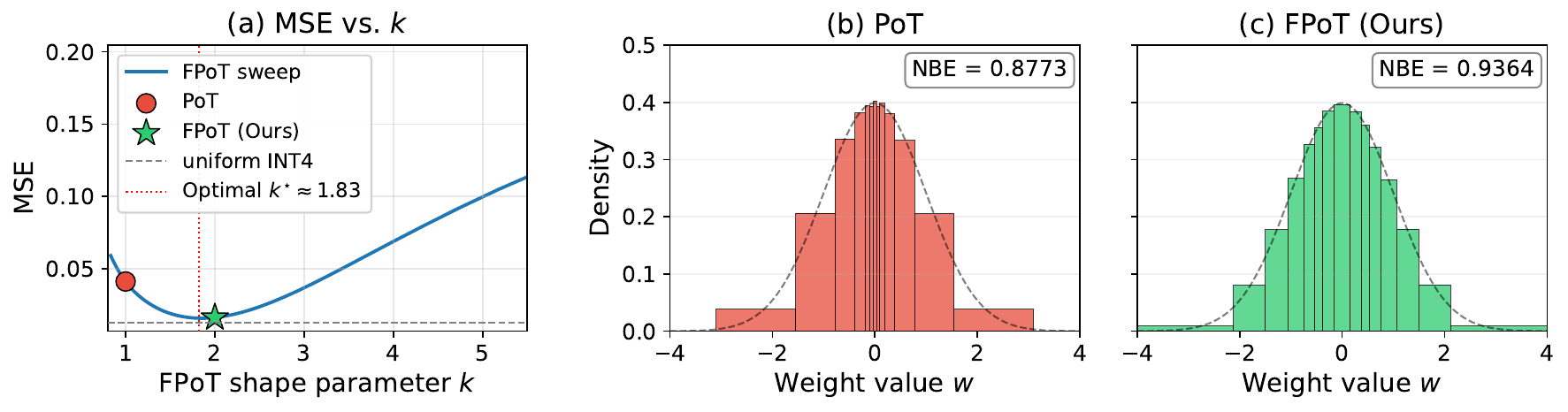}
\caption{%
(a)~MSE vs.\ shape parameter $k$ for $\mathcal{N}(0,1)$ weights at 4-bit ($N{=}15$ levels): the MSE is minimized at $k^\star \approx 1.8$, so the integer choice $k{=}2$ is near-optimal.
(b,\,c)~Bin occupancy under the Gaussian proxy. 
FPoT concentrates levels near the high-density mode, while PoT spreads levels across an excessively wide range, leaving most of them in the sparsely populated tail.
Dashed curves: input density $p_W$.
}
\label{fig:mse-nbe-vs-base}
\label{fig:bin-occupancy-gaussian-body}
\end{figure*}

%% file: figures/fig_tradeoff_float.tex
\begin{figure}[t]
    \centering
    \includegraphics[width=\linewidth]{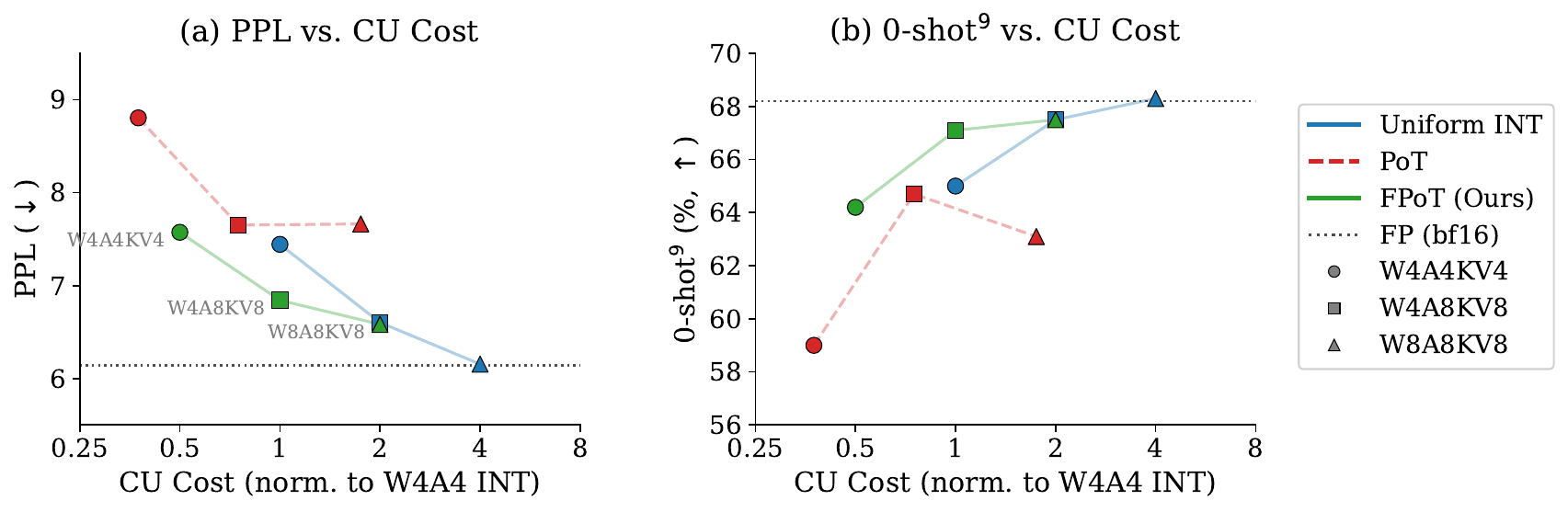}
    \caption{Accuracy--cost Pareto trade-off for Llama-3-8B. FPoT achieves near-baseline PPL at ${\sim}0.50{\times}$ CU cost across all bit-widths. Per-model plots in Appendix~\ref{app:hw_tradeoff}.}
    \label{fig:tradeoff}
\end{figure}

%% file: sections/05_experimental_results.tex
\section{Experimental Results}
\label{sec:results}

\subsection{Experimental Setup}

FPoT is framework-agnostic with respect to the choice of state-of-the-art LLM PTQ preprocessing; for the main results we use the OSTQuant framework~\citep{ostquant2025} for preprocessing, with rotation parameters jointly optimized with smooth scaling factors via Riemannian gradient descent for 400 steps with straight-through estimator (STE) gradients through discrete PoT quantization. Final weights are obtained via GPTQ~\citep{frantar2022gptq} with PoT targets. Cross-framework portability with DuQuant is validated in Appendix~\ref{app:duquant}. We evaluate WikiText-2~\citep{merity2017pointer} perplexity (PPL$\downarrow$) and the unweighted mean of zero-shot accuracy across nine benchmarks (0-shot$^9$$\uparrow$, in \%): ARC-Challenge, ARC-Easy~\citep{clark2018arc}, HellaSwag~\citep{zellers2019hellaswag}, WinoGrande~\citep{sakaguchi2021winogrande}, PIQA~\citep{bisk2020piqa}, BoolQ~\citep{clark2019boolq}, OpenBookQA~\citep{mihaylov2018openbookqa}, LAMBADA~\citep{paperno2016lambada}, and Social IQA~\citep{sap2019socialiqa}. All zero-shot evaluations are conducted via the lm-eval-harness framework~\citep{gao2024lmevalharness}. Table~\ref{tab:main_results} reports mean$\pm$std over three calibration seeds (0, 42, 123); all other experiments use a single calibration seed (seed 42) unless the caption explicitly states otherwise. To scale FPoT training to 14B models on a single A100-80GB GPU, we implement a fused Triton~\citep{tillet2019triton} kernel that merges the FPoT quantize--dequantize--STE pipeline into a single launch, reducing per-projection intermediate memory by $82\%$.

\subsection{Main Results}
\label{sec:main-results}

Table~\ref{tab:main_results} presents the main comparison of our proposed FPoT method against the non-PoT (uniform INT) baseline and standard PoT across seven models and three bit-width configurations. Our default configuration uses FPoT with $G{=}16$ grid-alignment and 1-adder approximate-grid quantization ($\sqrt{2} \approx 1.5$); activations and KV cache use uniform INT quantization throughout.

\input{tables/tab_main_results}

\textbf{FPoT achieves near-lossless PoT quantization in 20 of 21 settings} (per the $\Delta\text{PPL}{<}{+}0.5$, $\Delta\text{0-shot}^9 \geq -1.0$\%p criterion stated in the introduction). On Llama-3-8B W4A4KV4, FPoT reduces PPL overhead from $+1.43$ (PoT) to $+0.24$---an $83\%$ reduction---while downstream accuracy degradation shrinks from $-5.6$\%p to $-0.4$\%p. Since $k{=}2$ is specifically optimized for the 4-bit regime (\S\ref{sec:grid-matching}), we analyze results primarily within the W4 operating points (W4A4KV4 and W4A8KV8, 14~configurations across seven models): $\Delta$PPL ranges from $+0.05$ (Qwen2.5-14B W4A4KV4) to $+0.26$ (Qwen2.5-1.5B W4A8KV8)---\emph{strictly below $+0.3$ at the designed operating point}---with PoT overhead reductions of $71\%$--$94\%$ ($67\%$--$94\%$ across all twenty-one configurations including W8A8KV8). The W8A8KV8 configurations lie outside the 4-bit-specific design range yet still satisfy the $+0.5$ near-lossless threshold ($\Delta$PPL up to $+0.42$), demonstrating that FPoT$_{k{=}2}$ generalizes beyond its design point.

\subsection{\texorpdfstring{$\sqrt{2}$}{√2} Approximation Quality}
\label{sec:sqrt2-approx}

The FPoT PE requires a fixed $\times\!\sqrt{2}$ for odd-exponent weights (${\sim}50\%$ of MACs). Table~\ref{tab:sqrt2-approx} compares canonical signed digit (CSD) shift-add approximations on Llama-3-8B (W4A4KV4) under three strategies: na\"{\i}ve post-hoc substitution (``Exact Grid'': quantize to exact grid, dequantize with approximate coefficient), \emph{approximate-grid quantization} (``Approx Grid'': calibrate on the actual hardware grid), and \emph{Aware} training (use the approximate $\sqrt{2}$ during both OSTQuant training and GPTQ/eval). CU cost is evaluated under the same na\"{\i}ve component model as \S\ref{sec:hw_cost}, revealing that additional CSD adders rapidly consume the FPoT hardware advantage: the 3-adder variant already matches INT4 cost ($1.00{\times}$), leaving only the 1-adder at the knee of the Pareto frontier.

\input{tables/tab_sqrt2_approx}

With na\"{\i}ve substitution, the 1-adder ($y + y{\gg}1 = 1.5y$ in Fig.~\ref{fig:pe-arch}(d), 6.1\% error) degrades PPL by $+0.17$. However, with approximate-grid quantization, the same 1-adder achieves PPL\,$= 7.574$---within $+0.01$ of exact $\sqrt{2}$---because weight calibration absorbs the approximation error as part of the quantization grid rather than as post-hoc distortion. The hierarchy \textbf{Approx Grid $>$ Aware $>$ Exact Grid} holds for the 1-adder case (PPL $7.574 < 7.625 < 7.729$): aware training ($7.625$) lies between the two extremes since the STE gradient through approximate dequantization is less accurate than through exact dequantization, while smaller approximation errors (2-adder, 3-adder) make all three strategies converge. This yields a simple PE whose compute unit is ${\sim}0.50{\times}$ that of an INT4 multiplier with no measurable accuracy loss---a Pareto-optimal point among CSD variants (Table~\ref{tab:sqrt2-approx}): the 2-adder and 3-adder alternatives pay $1.5{\times}$ and $2{\times}$ more CU, respectively, for at most a $0.005$ PPL improvement, while the 0-adder PoT ($0.38{\times}$) incurs an unusable $+1.23$ PPL penalty.

\subsection{Additional Experiments}
\label{sec:ost-ablation}

\paragraph{Preprocessing.}
Both rotation and per-channel scaling are necessary: without any transform FPoT yields catastrophic PPL\,$= 181.3$; only the full combination (learned rotation + scale) reaches $7.56$, within $+0.12$ of the non-PoT baseline (Table~\ref{tab:rotation_ablation}, Appendix~\ref{app:preprocessing}).

\paragraph{Preprocessing $\times$ quantizer factorial.}
Training the preprocessing for one grid and quantizing to another separates the two factors. The weight quantizer moves PPL about ten times more than the preprocessing target ($1.51$ versus $0.15$ across means), and the order INT, FPoT, PoT holds under every preprocessing (Appendix~\ref{app:factorial}).

\paragraph{Full FPoT (weight + activation).}
Applying the logarithmic grid to both weights and activations degrades accuracy severely (PPL $16.9$--$114.1$ at $k{=}1$, $8.21$--$52.49$ at $k{=}2$, versus $7.56$ for weight-only; Appendix~\ref{app:wonly-vs-full}). Uniform INT activations remain the practical choice.

\paragraph{Weight grid-alignment.}
Grid-alignment (Eq.~\eqref{eq:w-grid-align}) reduces PPL by up to $-0.051$ at $k{=}1$ at zero inference cost; at $k{=}2$ the finer grid self-corrects ($|\Delta\text{PPL}| \leq 0.034$; Appendix~\ref{app:wgrid-align}).

\paragraph{Hyperparameter sensitivity.}
The zero threshold $\tau$ and log-vs-linear rounding mode have negligible impact ($\Delta$PPL $< 0.03$); see Appendices~\ref{app:zero-thresh} and~\ref{app:round-mode}.

\paragraph{Calibration robustness.}
FPoT's data-independent grid boundaries yield robust calibration: $|\Delta\text{PPL}| \leq 0.065$ across $n \in \{128, 256, 512\}$ and only $+0.06$--$+0.10$ shift from WikiText-2 to C4 (Appendix~\ref{app:cal-sensitivity}).

\paragraph{Cross-framework portability.}
A different PTQ framework yields comparable FPoT overhead ($+0.05$--$+0.34$ PPL; Appendix~\ref{app:duquant}, Table~\ref{tab:duquant_comparison}).

\paragraph{Generation tasks.}
On two instruction-tuned models, standard PoT trails uniform INT by about 15 to 25 points on math and code, and FPoT recovers most of that gap. On Llama-3.1-8B-Instruct, for example, HumanEval rises from $27.6$ with PoT to $47.8$ with FPoT, against $52.4$ for INT (Appendix~\ref{app:generation}).

\paragraph{LogART comparison.}
In the weight-only setting (W4A16), LogART achieves lower PPL but degrades catastrophically (PPL\,$>$\,3900) under post-hoc INT4 activation without PTQ preprocessing, while FPoT remains stable (Appendix~\ref{app:logart}, Table~\ref{tab:logart-comparison}).

%% file: tables/tab_main_results.tex
\begin{table*}[ht]
\centering
\caption{%
Main results across models and quantization configurations (mean$_{\pm\text{std}}$ over 3 calibration seeds).
Uniform: OSTQuant uniform INT baseline; PoT (base 2) and FPoT (base $\sqrt{2}$, Ours) replace the weight grid on the same pipeline; default configuration in \S\ref{sec:main-results}.
PPL: WikiText-2 perplexity ($\downarrow$).
0-shot$^9$: mean zero-shot accuracy across nine benchmarks ($\uparrow$, \%).
}
\label{tab:main_results}
\setlength{\tabcolsep}{3.5pt}
\begin{adjustbox}{max width=\textwidth}
\begin{tabular}{ll cccccc}
\toprule
 & & \multicolumn{2}{c}{W4A4KV4} & \multicolumn{2}{c}{W4A8KV8} & \multicolumn{2}{c}{W8A8KV8} \\
\cmidrule(lr){3-4} \cmidrule(lr){5-6} \cmidrule(lr){7-8}
Model & Weight Quant & PPL${\downarrow}$ & 0-shot$^9{\uparrow}$ & PPL${\downarrow}$ & 0-shot$^9{\uparrow}$ & PPL${\downarrow}$ & 0-shot$^9{\uparrow}$ \\
\midrule
\multirow{3}{*}{Llama-3-8B}
 & Uniform      & 7.36$_{\pm.07}$ & 65.2$_{\pm.4}$ & 6.60$_{\pm.01}$ & 67.5$_{\pm.1}$ & 6.16$_{\pm.00}$ & 68.2$_{\pm.1}$ \\
 & PoT          & 8.79$_{\pm.04}$ & 59.6$_{\pm.6}$ & 7.64$_{\pm.02}$ & 64.0$_{\pm.7}$ & 7.52$_{\pm.12}$ & 63.8$_{\pm.7}$ \\
 & \textbf{FPoT (Ours)} & \textbf{7.60}$_{\pm.02}$ & \textbf{64.8}$_{\pm.6}$ & \textbf{6.85}$_{\pm.01}$ & \textbf{67.3}$_{\pm.5}$ & \textbf{6.58}$_{\pm.00}$ & \textbf{67.7}$_{\pm.2}$ \\
\midrule
\multirow{3}{*}{Llama-2-7B}
 & Uniform      & 5.97$_{\pm.01}$ & 62.1$_{\pm.3}$ & 5.63$_{\pm.04}$ & 63.8$_{\pm.2}$ & 5.48$_{\pm.00}$ & 65.1$_{\pm.0}$ \\
 & PoT          & 6.93$_{\pm.00}$ & 58.0$_{\pm.1}$ & 6.27$_{\pm.03}$ & 61.4$_{\pm.7}$ & 6.05$_{\pm.04}$ & 62.3$_{\pm.2}$ \\
 & \textbf{FPoT (Ours)} & \textbf{6.11}$_{\pm.04}$ & \textbf{61.9}$_{\pm.2}$ & \textbf{5.75}$_{\pm.02}$ & \textbf{63.8}$_{\pm.2}$ & \textbf{5.65}$_{\pm.03}$ & \textbf{64.2}$_{\pm.3}$ \\
\midrule
\multirow{3}{*}{Llama-2-13B}
 & Uniform      & 5.26$_{\pm.01}$ & 65.6$_{\pm.2}$ & 5.03$_{\pm.01}$ & 67.1$_{\pm.3}$ & 4.89$_{\pm.00}$ & 67.7$_{\pm.0}$ \\
 & PoT          & 5.90$_{\pm.02}$ & 62.7$_{\pm.5}$ & 5.48$_{\pm.01}$ & 65.2$_{\pm.3}$ & 5.31$_{\pm.03}$ & 65.2$_{\pm.7}$ \\
 & \textbf{FPoT (Ours)} & \textbf{5.37}$_{\pm.02}$ & \textbf{65.2}$_{\pm.1}$ & \textbf{5.11}$_{\pm.02}$ & \textbf{66.6}$_{\pm.1}$ & \textbf{5.03}$_{\pm.01}$ & \textbf{67.2}$_{\pm.4}$ \\
\midrule
\multirow{3}{*}{Qwen2.5-1.5B}
 & Uniform      & 11.10$_{\pm.14}$ & 38.7$_{\pm.4}$ & 9.75$_{\pm.03}$ & 42.7$_{\pm.6}$ & 9.32$_{\pm.00}$ & 44.0$_{\pm.2}$ \\
 & PoT          & 13.23$_{\pm.11}$ & 37.2$_{\pm.8}$ & 11.35$_{\pm.09}$ & 39.3$_{\pm.4}$ & 11.33$_{\pm.05}$ & 38.8$_{\pm1.4}$ \\
 & \textbf{FPoT (Ours)} & \textbf{11.23}$_{\pm.03}$ & \textbf{39.9}$_{\pm.4}$ & \textbf{10.01}$_{\pm.03}$ & \textbf{42.4}$_{\pm1.0}$ & \textbf{9.73}$_{\pm.01}$ & \textbf{43.0}$_{\pm.7}$ \\
\midrule
\multirow{3}{*}{Qwen2.5-7B}
 & Uniform      & 7.56$_{\pm.01}$ & 66.4$_{\pm.5}$ & 7.04$_{\pm.01}$ & 68.1$_{\pm.1}$ & 6.87$_{\pm.00}$ & 68.6$_{\pm.1}$ \\
 & PoT          & 8.22$_{\pm.01}$ & 62.9$_{\pm.6}$ & 7.56$_{\pm.02}$ & 65.5$_{\pm.2}$ & 7.57$_{\pm.02}$ & 66.2$_{\pm1.0}$ \\
 & \textbf{FPoT (Ours)} & \textbf{7.62}$_{\pm.02}$ & \textbf{66.3}$_{\pm.3}$ & \textbf{7.14}$_{\pm.01}$ & \textbf{68.6}$_{\pm.6}$ & \textbf{7.03}$_{\pm.02}$ & \textbf{68.5}$_{\pm.6}$ \\
\midrule
\multirow{3}{*}{Qwen2.5-14B}
 & Uniform      & 6.33$_{\pm.01}$ & 68.7$_{\pm.4}$ & 5.65$_{\pm.01}$ & 71.1$_{\pm.4}$ & 5.31$_{\pm.00}$ & 71.1$_{\pm.2}$ \\
 & PoT          & 7.21$_{\pm.05}$ & 64.8$_{\pm1.1}$ & 6.48$_{\pm.05}$ & 68.8$_{\pm.3}$ & 6.40$_{\pm.01}$ & 68.7$_{\pm.4}$ \\
 & \textbf{FPoT (Ours)} & \textbf{6.38}$_{\pm.04}$ & \textbf{68.9}$_{\pm.3}$ & \textbf{5.85}$_{\pm.08}$ & \textbf{71.3}$_{\pm.8}$ & \textbf{5.60}$_{\pm.00}$ & \textbf{72.0}$_{\pm.1}$ \\
\midrule
\multirow{3}{*}{Mistral-7B}
 & Uniform      & 5.62$_{\pm.01}$ & 66.5$_{\pm.3}$ & 5.41$_{\pm.02}$ & 67.8$_{\pm.1}$ & 5.32$_{\pm.00}$ & 68.3$_{\pm.1}$ \\
 & PoT          & 6.00$_{\pm.03}$ & 64.6$_{\pm.4}$ & 5.69$_{\pm.02}$ & 65.5$_{\pm.7}$ & 5.71$_{\pm.02}$ & 66.4$_{\pm.3}$ \\
 & \textbf{FPoT (Ours)} & \textbf{5.71}$_{\pm.05}$ & \textbf{66.1}$_{\pm.5}$ & \textbf{5.49}$_{\pm.04}$ & \textbf{67.3}$_{\pm.4}$ & \textbf{5.45}$_{\pm.07}$ & \textbf{67.7}$_{\pm.4}$ \\
\bottomrule
\end{tabular}
\end{adjustbox}
\end{table*}

%% file: tables/tab_sqrt2_approx.tex
\begin{table}[ht]
\centering
\caption{
  $\sqrt{2}$ approximation strategies on Llama-3-8B W4A4KV4 (FPoT; single calibration seed). Exact Grid / Approx Grid (Ours) / Aware: see \S\ref{sec:sqrt2-approx} for definitions. CU/${\times}$: same na\"{\i}ve component model (\S\ref{sec:hw_cost}); $^\ddagger$exact $\sqrt{2}$ has no finite CSD realization. 0-shot$^9$ in \%.
}
\label{tab:sqrt2-approx}
\setlength{\tabcolsep}{2.5pt}
\begin{tabular}{lccccccccc}
\toprule
 & & & & \multicolumn{2}{c}{Exact Grid} & \multicolumn{2}{c}{Approx Grid} & \multicolumn{2}{c}{Aware} \\
\cmidrule(lr){5-6} \cmidrule(lr){7-8} \cmidrule(lr){9-10}
Impl. & Value (Err.) & Add. & CU/${\times}$ & PPL${\downarrow}$ & 0-shot$^9{\uparrow}$ & PPL${\downarrow}$ & 0-shot$^9{\uparrow}$ & PPL${\downarrow}$ & 0-shot$^9{\uparrow}$ \\
\midrule
Exact      & 1.414 (0\%)   & ---    & N/A$^\ddagger$  & 7.560 & 64.7 & ---   & ---  & ---   & --- \\
3-adder    & 1.438 (1.6\%) & 3      & $1.00{\times}$  & 7.580 & 64.4 & 7.569 & 65.0 & 7.611 & 64.3 \\
2-adder    & 1.375 (2.8\%) & 2      & $0.75{\times}$  & 7.575 & 64.5 & 7.584 & 64.4 & 7.584 & 64.8 \\
\textbf{1-adder} & \textbf{1.500 (6.1\%)} & \textbf{1} & $\bm{0.50{\times}}$ & 7.729 & 63.4 & \textbf{7.574} & \textbf{64.2} & 7.625 & 63.8 \\
\midrule
$k{=}1$    & ---           & 0      & $0.38{\times}$  & 8.806 & 59.0 & ---   & ---  & ---   & --- \\
\bottomrule
\end{tabular}

\vspace{3pt}
{\footnotesize
$^\ddagger$~Exact $\sqrt{2}$ requires a full multiplier-grade unit (no finite CSD realization); not a shift-add alternative.
}
\end{table}

%% file: sections/06_discussion.tex
\section{Discussion}
\label{sec:discussion}

\paragraph{Design rationale and comparison with prior PoT methods.}
FPoT's three design choices---a regular geometric grid amenable to GPTQ (vs.\ APoT's irregular additive decompositions~\citep{li2020apot}), uniform INT activations eliminating runtime $\log_2$ conversion~\citep{elhoushi2021deepshift,li2023denseshift}, and 4-bit grid-resolution targeting (vs.\ PoT-PTQ's 2--3-bit scale-calibration focus~\citep{potpourri2025})---together let FPoT reach near-lossless W4A4KV4 with a multiplier-free datapath. LogART~\citep{logart2026} reports strong W4A16 accuracy but lacks LLM W4A4KV4 evaluation due to absent PTQ preprocessing (Appendix~\ref{app:logart}).

\paragraph{Broader impact.}
Multiplier-free LLM inference can reduce the energy consumption and carbon footprint of large-scale deployment. However, cheaper inference may also lower barriers to harmful applications; this work does not introduce new capabilities and applies to existing public models with slightly degraded accuracy.

\paragraph{Limitations.}
Efficiency projections use analytical cost models anchored to published RTL synthesis~\citep{przewlocka2022pot}; open-source logic synthesis (Yosys~\citep{wolf2013yosys}) confirms the structural ordering and bit-width scaling (Appendix~\ref{app:hw_yosys}), and end-to-end RTL/ASIC validation is left for future work. These figures are per weight MAC, so the end-to-end gain is largest in compute-bound prefill and smaller in memory-bound decode (Appendix~\ref{app:mac-breakdown}). FPoT depends on PTQ preprocessing. Without it, W4A4KV4 FPoT reaches PPL $181.3$ (Appendix~\ref{app:preprocessing}). Uniform INT at W4A4KV4 also relies on such preprocessing to handle activation outliers~\citep{ashkboos2024quarot, liu2024spinquant}, and the factorial in Appendix~\ref{app:factorial} shows that FPoT keeps its advantage over PoT under each preprocessing target. Preprocessing training requires float32 weight copies, limiting single-GPU (80\,GB) experiments to ${\leq}14$B parameters. The $k{=}2$ optimum is 4-bit specific (8-bit shifts to ${\approx}19$).

%% file: sections/07_conclusion.tex
\section{Conclusion}

We introduced FPoT---a half-octave fractional PoT weight quantization scheme that, combined with uniform INT activation quantization and state-of-the-art LLM PTQ preprocessing, achieves near-lossless accuracy while enabling multiplier-free shift-add inference. Three contributions enable this result. First, a Gaussian-proxy analysis identifies base-$\sqrt{2}$ as the optimal integer-resolution grid at 4-bit, halving the worst-case relative error of standard PoT ($41.4\%{\rightarrow}18.9\%$). Second, approximate-grid quantization absorbs the 1-adder $\sqrt{2}$ hardware error via weight calibration, yielding a PE whose compute unit is ${\sim}0.50{\times}$ the INT4 multiplier gate count (vs.\ ${\sim}0.38{\times}$ for standard PoT). Third, across twenty-one configurations spanning seven models and three quantization settings (W4A4KV4 / W4A8KV8 / W8A8KV8), 
FPoT recovers zero-shot accuracy to within $1.0\%$p of the uniform INT baseline in $20{/}21$ configurations (mean gap $-0.17\%$p; all $14$ W4 configurations within $0.5\%$p), and integrates with state-of-the-art LLM PTQ frameworks at comparable overhead (Appendix~\ref{app:duquant}).

These results suggest that FPoT is a promising pathway toward shift-based LLM inference hardware with near-lossless accuracy, with future work including scaling to larger models, exploring higher bit-widths, and end-to-end RTL/ASIC validation.

%% file: sections/A_appendix_preliminaries.tex
\section{Preliminaries}
\label{app:prelim}

\subsection{Uniform Quantization}

Standard uniform (linear) quantization maps a floating-point tensor to a finite set of evenly-spaced levels. For symmetric $b$-bit quantization with per-token dynamic scale $s$:
\begin{equation}
    Q_{\mathrm{lin}}(x) = \mathrm{clamp}\!\left( \left\lfloor \frac{x}{s} \right\rceil, -2^{b-1}, 2^{b-1}{-}1 \right) \cdot s,
\end{equation}
where $s = \|x\|_\infty / (2^{b-1} - 1)$. In this work, activations and the KV cache use uniform INT quantization in all configurations (in its asymmetric form, with a zero point), while weights use logarithmic (FPoT) quantization (\S\ref{sec:method}).

\subsection{Power-of-Two Quantization}
\label{app:pot-background}

PoT quantization restricts each weight to zero or $\pm 2^{n}$ times a scale. The product of an integer activation $x$ with $2^n$ is a barrel shift of $x$ by $n$ bits, so a MAC becomes a shift followed by an integer addition into the accumulator and needs no multiplier array. The idea was developed for CNNs. DeepShift~\citep{elhoushi2021deepshift} trains networks whose weights are single signed powers of two. APoT~\citep{li2020apot} represents each weight as a sum of $n$ PoT terms and pays one shift-add per term. DenseShift~\citep{li2023denseshift} removes the zero level so that every code carries a nonzero value. Base-$\sqrt{2}$ grids appear in logarithmic accelerators, which approximate $\sqrt{2}$ by $2^0 + 2^{-1}$~\citep{xu2018multibase, xu2023aslog}, and in LogART~\citep{logart2026}. With one level per octave, a 4-bit PoT grid rounds with up to $41.4\%$ relative error. CNNs trained or fine-tuned for the grid tolerate this, but LLM weights quantized after training do not. Standard PoT adds $+1.43$ PPL on Llama-3-8B W4A4KV4 even with preprocessing (Table~\ref{tab:main_results}). FPoT halves the octave, so one extra adder buys a twice finer grid.

\subsection{State-of-the-Art LLM PTQ Preprocessing}
\label{sec:ost}

Recent LLM PTQ preprocessing methods (QuaRot~\citep{ashkboos2024quarot}, SpinQuant~\citep{liu2024spinquant}, OSTQuant~\citep{ostquant2025}, DuQuant~\citep{duquant2024}; see Section~\ref{sec:rw} for taxonomy) apply learned transformations---orthogonal rotations, per-channel scaling, or their composition---to redistribute channel-concentrated outliers before quantization. For input activation $X \in \mathbb{R}^{n \times d}$ and weight $W \in \mathbb{R}^{d \times m}$, these methods rewrite the linear operation as
\begin{equation}
    Y = XW = (XT)(T^{-1} W) = \hat{X}\hat{W},
\end{equation}
where $T$ is an invertible transform (orthogonal rotation, scaling matrix, or their composition). This reduces kurtosis and produces smoother distributions amenable to the logarithmic PoT grid. We primarily build on OSTQuant, which jointly trains a Hadamard rotation with per-channel smooth scales via Riemannian gradient descent on the Stiefel manifold (400 steps with STE). The transform adds only ${\approx}0.02\%$ overhead in MAC operations (${\sim}49$\,K additions per layer at $d{=}4{,}096$ vs.\ $218$\,M weight MACs). Our FPoT method is compatible with any such preprocessing; we primarily evaluate with OSTQuant and confirm DuQuant portability in Appendix~\ref{app:duquant}.

%% file: sections/B_appendix_grid_matching_theory.tex
\section{Grid-Matching Theory}
\label{app:theory}

\subsection{Normalized Bin Entropy: Properties and Proofs}
\label{app:nbe-properties}

We establish five properties of Normalized Bin Entropy (NBE, Eq.~\eqref{eq:nbe}) that justify its use as a grid-level design metric.

\paragraph{Property 1 (Bounds).}
For any probability vector $\mathbf{p} \in \Delta^{N-1}$, $0 \leq \mathrm{NBE}(\mathbf{p}) \leq 1$, with $\mathrm{NBE} = 1$ iff $p_i = 1/N$ for all $i$ (uniform occupancy), and $\mathrm{NBE} = 0$ iff $p_i = 1$ for some single level $i$.

\textit{Proof.} By the non-negativity of Shannon entropy, $H(\mathbf{p}) \geq 0$ with equality iff $\mathbf{p}$ is a point mass. The upper bound $H(\mathbf{p}) \leq \log_2 N$ is the maximum-entropy property of the uniform distribution over $N$ outcomes, proven by Jensen's inequality applied to $-x \log x$. Dividing by $\log_2 N$ yields the stated bounds. \qed

\paragraph{Property 2 (Information-theoretic interpretation).}
Let $L^\star(\mathbf{p})$ denote the expected code length of an optimal prefix-free (Shannon--Fano or Huffman) code for the occupancy distribution $\mathbf{p}$. Then
\begin{equation*}
  \log_2 N \cdot \mathrm{NBE}(\mathbf{p}) \;=\; H(\mathbf{p}) \;\leq\; L^\star(\mathbf{p}) \;<\; H(\mathbf{p}) + 1.
\end{equation*}
That is, $\log_2 N \cdot \mathrm{NBE}$ is the tightest achievable expected code length per weight, and NBE measures the fraction of the fixed-length budget $\log_2 N$ that carries information under the data distribution.

\textit{Proof.} Shannon's source coding theorem states that for any distribution $\mathbf{p}$, $H(\mathbf{p}) \leq L^\star(\mathbf{p}) < H(\mathbf{p}) + 1$ bits per symbol. Dividing by $\log_2 N$ and rearranging yields the claim. \qed

\paragraph{Property 3 (Affine invariance of level values).}
NBE depends only on the occupancy probabilities $\mathbf{p}$, not on the level values $\{\ell_i\}$. Consequently, affine relabeling $\ell_i' = a \ell_i + b$ ($a \neq 0$) leaves NBE unchanged, provided the induced decision boundaries assign the same weight population to each level.

\textit{Proof.} Eq.~\eqref{eq:nbe} depends only on $\mathbf{p}$. Under affine relabeling with unchanged decision topology, each $p_i$ is preserved. \qed

This property ensures that NBE captures the \emph{shape} of the level-mass mapping independently of level magnitudes, distinguishing it from value-dependent metrics such as MSE.

\paragraph{Property 4 (Grid-structure sensitivity under fixed dynamic range).}
Let $Q_1, Q_2$ be two quantizers with identical $N$ and identical dynamic range $[-R, R]$ but different level placements $\{\ell_i^{(1)}\}, \{\ell_i^{(2)}\}$ (e.g., uniform vs.\ logarithmic). For a non-uniform weight distribution $p_W$ concentrated near $0$, generally $\mathrm{NBE}(Q_1; p_W) \neq \mathrm{NBE}(Q_2; p_W)$.

\textit{Proof sketch.} Different level placements yield different decision regions, which partition $p_W$ into different mass vectors $\mathbf{p}^{(1)} \neq \mathbf{p}^{(2)}$. While Shannon entropy can coincidentally coincide on distinct probability vectors (e.g., under permutation), structurally different partitions of a log-concave $p_W$---such as those induced by uniform vs.\ logarithmic grids---yield mass vectors that differ not merely by permutation but in their degree of concentration, producing distinct entropies. For $\mathcal{N}(0,1)$ weights with $N{=}15$, uniform INT4 yields NBE $= 0.9026$ while \fpotk{2} yields NBE $= 0.9364$ (Table~\ref{tab:mse-nbe}), quantitatively confirming this claim. \qed

This is the property that volumetric utilization metrics (e.g., QSUR~\citep{ostquant2025}) lack: they depend only on the bounding hypercube volume, not on the internal level structure.

\paragraph{Property 5 (Correlation with MSE for log-concave distributions).}
For log-concave symmetric weight distributions (satisfied by the Gaussian proxy and, after PTQ preprocessing, empirically by LLM weights), the quantizer base $\beta = 2^{1/k}$ that maximizes NBE lies within a narrow neighborhood of the base that minimizes MSE. Specifically, for $\mathcal{N}(0,1)$ weights with 4-bit budget ($N = 15$), both optima satisfy $\beta^\star \in [2^{0.50}, 2^{0.60}]$ (equivalently, $k^\star \in [1.67, 2.00]$).

\textit{Empirical justification.} The MSE of a log-base-$\beta$ quantizer on $p_W$ decomposes into a tail-clipping term (decreasing in $\beta$, as a larger $\beta^{N-1}$ reduces clipped tail mass) and an in-range rounding term (increasing in $\beta$, as coarser bins accept larger within-bin errors); the interior optimum balances these effects. NBE exhibits the analogous trade-off: too-small $\beta$ concentrates mass in tail-proximal bins, too-large leaves high-density bins over-populated. Numerical sweep on $\mathcal{N}(0,1)$ with $N{=}15$ confirms both optima in the stated interval (Table~\ref{tab:mse-nbe}, Fig.~\ref{fig:mse-nbe-vs-base}). A formal rate-distortion analysis is left to future work.

\textit{Remark.} Properties 1--4 are exact; Property 5 is empirical for our operating regime and serves to justify why NBE-based grid design yields near-MSE-optimal quantizers in practice.

\subsection{Log-Magnitude Distribution Derivation}
\label{app:log-magnitude}
Let $u = \log_2|w|$ for $w \sim \mathcal{N}(0,\sigma^2)$. Its probability density is
\begin{equation}
  f_U(u) = \frac{2\ln 2}{\sqrt{2\pi}\,\sigma}\, 2^{u}\, \exp\!\Bigl(-\frac{2^{2u}}{2\sigma^2}\Bigr).
  \label{eq:log-mag-pdf}
\end{equation}
Substituting $v = u - \log_2\sigma$ and expanding near $v{=}0$ gives $f_U \propto \exp(-v^2)$, so the log-magnitude is approximately Gaussian in log-space with standard deviation $1/(\sqrt{2}\ln 2) \approx 1.02$ octaves ($2^{1.02}$). Consequently, ${\sim}95\%$ of probability mass concentrates within ${\sim}4$ octaves around the mode. This ${\sim}4$-octave concentration is not unique to exact Gaussians: any symmetric, outlier-suppressed distribution with comparable kurtosis exhibits a similar log-domain spread, which is the condition that PTQ preprocessing enforces.

\subsection{MSE and NBE Across Integer \texorpdfstring{$k$}{k}}
\label{app:mse-nbe-table}
\input{tables/tab_mse_nbe}

Table~\ref{tab:mse-nbe} reports the 4-bit MSE and NBE under the Gaussian proxy across $k$. The optimal base $\beta^{\star} \approx 2^{0.55}$ yields an effective $k^{\star} \approx 1.8$ (MSE $= 1.23{\times}$ INT4, NBE $= 0.9417$); integer $k{=}2$ is the closest practical choice.

\paragraph{Bit-width dependence.}
At 8-bit (127 magnitudes per side), the abundance of levels permits much finer resolution without dynamic-range starvation, shifting the optimal to $k^{\star} \approx 19$. \fpotk{2} thus does not generalize as the preferred grid at higher bit-widths.

%% file: tables/tab_mse_nbe.tex
\begin{table}[t]
\centering
\caption{MSE and NBE of 4-bit PoT quantizers applied to $\mathcal{N}(0,1)$ weights ($N{=}15$ levels: 14~non-zero symmetric + 1~zero, log-domain rounding). MSE ratio is relative to uniform INT4.}
\label{tab:mse-nbe}
\small
\begin{tabular}{lccc}
\toprule
Method & MSE & NBE & MSE ratio \\
\midrule
Uniform INT4 & 0.0128 & 0.9026 & $1.00{\times}$ \\
\midrule
PoT & 0.0411 & 0.8773 & $3.21{\times}$ \\
\textbf{\fpotk{2}} & \textbf{0.0162} & \textbf{0.9364} & $\mathbf{1.27{\times}}$ \\
\fpotk{3} & 0.0368 & 0.8762 & $2.87{\times}$ \\
\fpotk{4} & 0.0688 & 0.8115 & $5.37{\times}$ \\
\midrule
Optimal ($k^{\star}{\approx}1.8$) & 0.0158 & 0.9417 & $1.23{\times}$ \\
\bottomrule
\end{tabular}
\end{table}

%% file: sections/C_appendix_empirical_bin_occupancy.tex
\section{Empirical Bin Occupancy}
\label{app:empirical}
\label{sec:appendix-bin-occupancy}

While the Gaussian-proxy bin occupancy (Fig.~\ref{fig:bin-occupancy-gaussian-body}, \S\ref{sec:fpot-motivation}) establishes the grid-matching mechanism analytically, we validate this on real LLM weights here. Fig.~\ref{fig:fpot-grid} compares the quantization grids of uniform INT4, PoT, and FPoT on the number line. Fig.~\ref{fig:bin-occupancy-llama3-8b} shows the aggregated empirical bin occupancy for Llama-3-8B (W4A4KV4).

\input{figures/fig_fpot_grid_float}

\input{figures/fig_bin_occupancy_llama3-8b_float}

We present the bin occupancy distributions of RTN-quantized weights (W4A4KV4) for each model evaluated in this work. These empirical distributions complement the theoretical Gaussian analysis (Fig.~\ref{fig:bin-occupancy-gaussian-body}) and the Llama-3-8B aggregated results (Fig.~\ref{fig:bin-occupancy-llama3-8b}).

Across all models, the empirical bin occupancy closely matches the theoretical Gaussian prediction: (a)~uniform INT4 exhibits a bell-shaped occupancy with ${\sim}14\%$ zero bin; (b)~PoT concentrates weights in the mid-exponent bins ($\pm 16$, $\pm 32$) with only ${\sim}3\%$ zero; (c)~FPoT achieves more uniform non-zero bin occupancy but maps ${\sim}20$--$22\%$ of weights to zero due to the $k$-adaptive zero threshold $\tau_2 = 2^{-1/4} \approx 0.841$ (Section~\ref{sec:fpot-zero}). The consistency across model families (Llama-2, Llama-3, Qwen2.5) and scales (1.5B--14B) confirms that recent LLM PTQ preprocessing produces weight distributions with similar log-domain concentration, validating the use of the Gaussian proxy analysis for grid design.

\subsection{Per-Projection Breakdown}

\begin{figure}[ht]
    \centering
    \includegraphics[width=\linewidth, height=0.9\textheight, keepaspectratio]{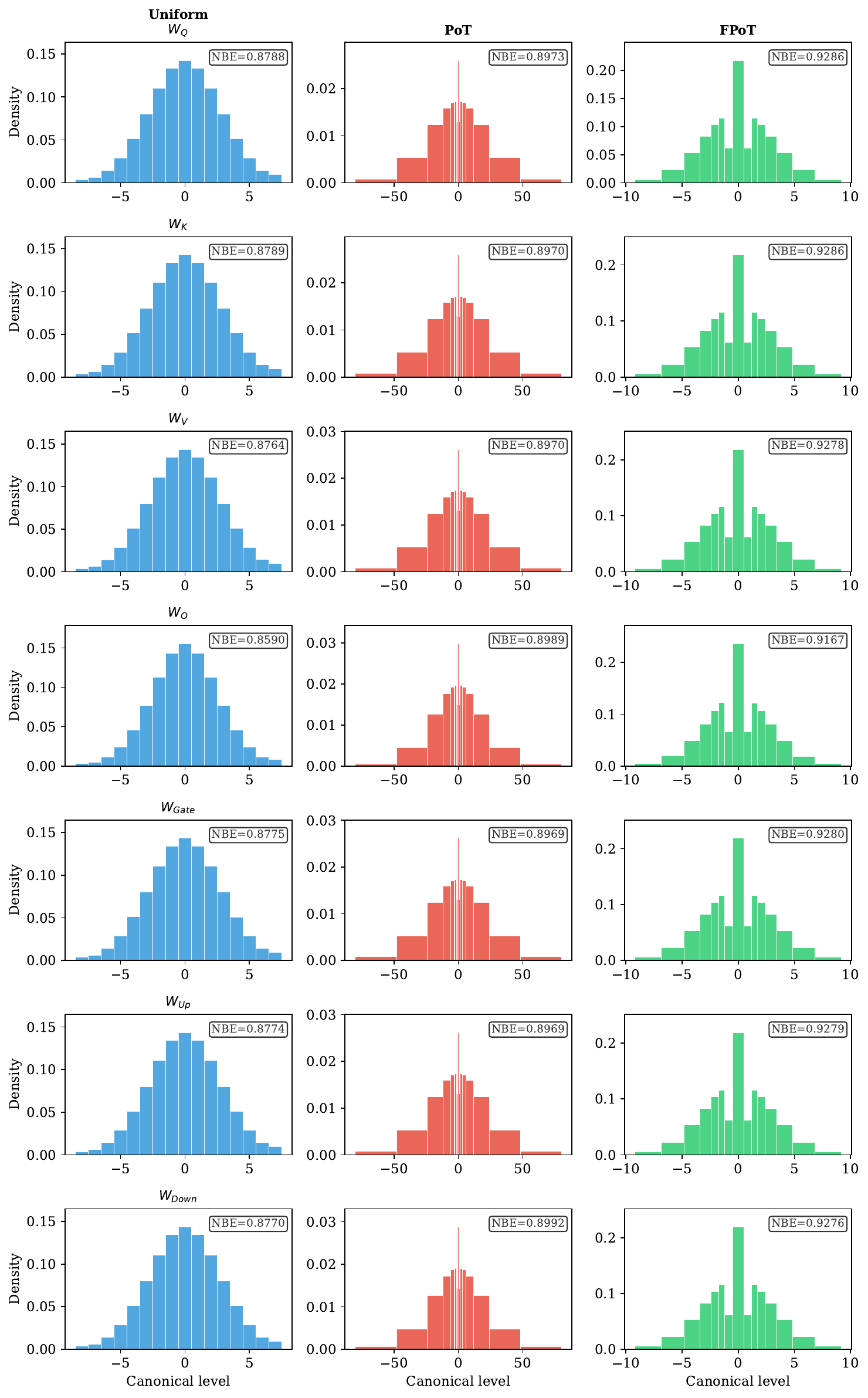}
    \caption{Per-projection bin occupancy of RTN-quantized Llama-3-8B weights (W4A4KV4). Rows: $W_Q$, $W_K$, $W_V$, $W_O$, $W_{Gate}$, $W_{Up}$, $W_{Down}$. Columns: Uniform, PoT, FPoT. The occupancy pattern is consistent across all seven projections, with FPoT achieving NBE $> 0.91$ uniformly (0.9167--0.9286), substantially above PoT and uniform INT4.}
    \label{fig:bin-occupancy-llama3-8b-per-proj}
\end{figure}

Fig.~\ref{fig:bin-occupancy-llama3-8b-per-proj} shows the per-projection bin occupancy for Llama-3-8B, complementing the aggregated view in Fig.~\ref{fig:bin-occupancy-llama3-8b}.

\FloatBarrier

\subsection{Effect of PTQ Preprocessing on Bin Occupancy}

\begin{figure}[ht]
    \centering
    \includegraphics[width=\linewidth]{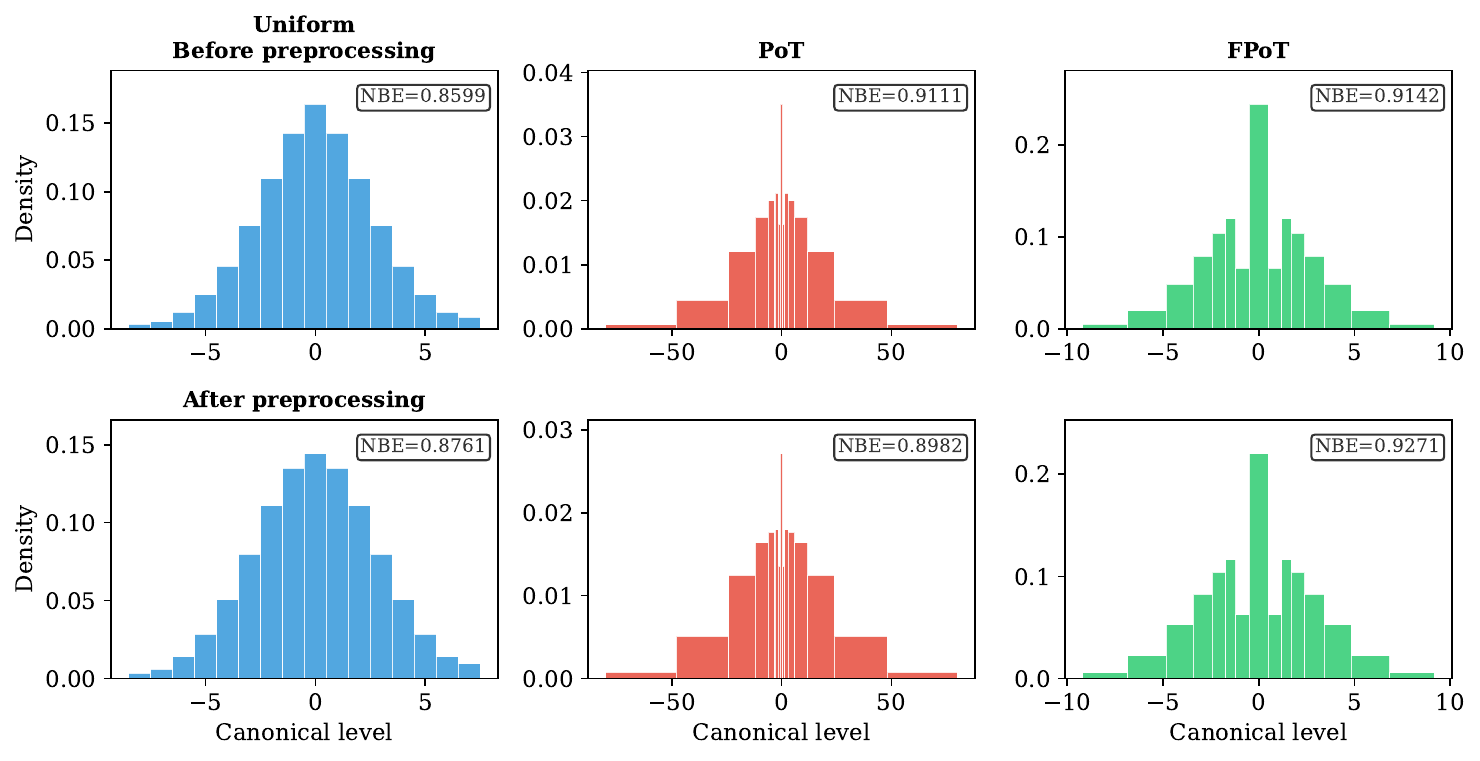}
    \caption{Bin occupancy before (top) and after (bottom) PTQ preprocessing for Llama-3-8B (W4A4KV4). uniform INT4 NBE rises from 0.8599 to 0.8761 and FPoT from 0.9142 to 0.9271; PoT drops slightly (0.9111 $\to$ 0.8982).}
    \label{fig:bin-occupancy-before-after-llama3-8b}
\end{figure}

\begin{figure}[ht]
    \centering
    \includegraphics[width=\linewidth]{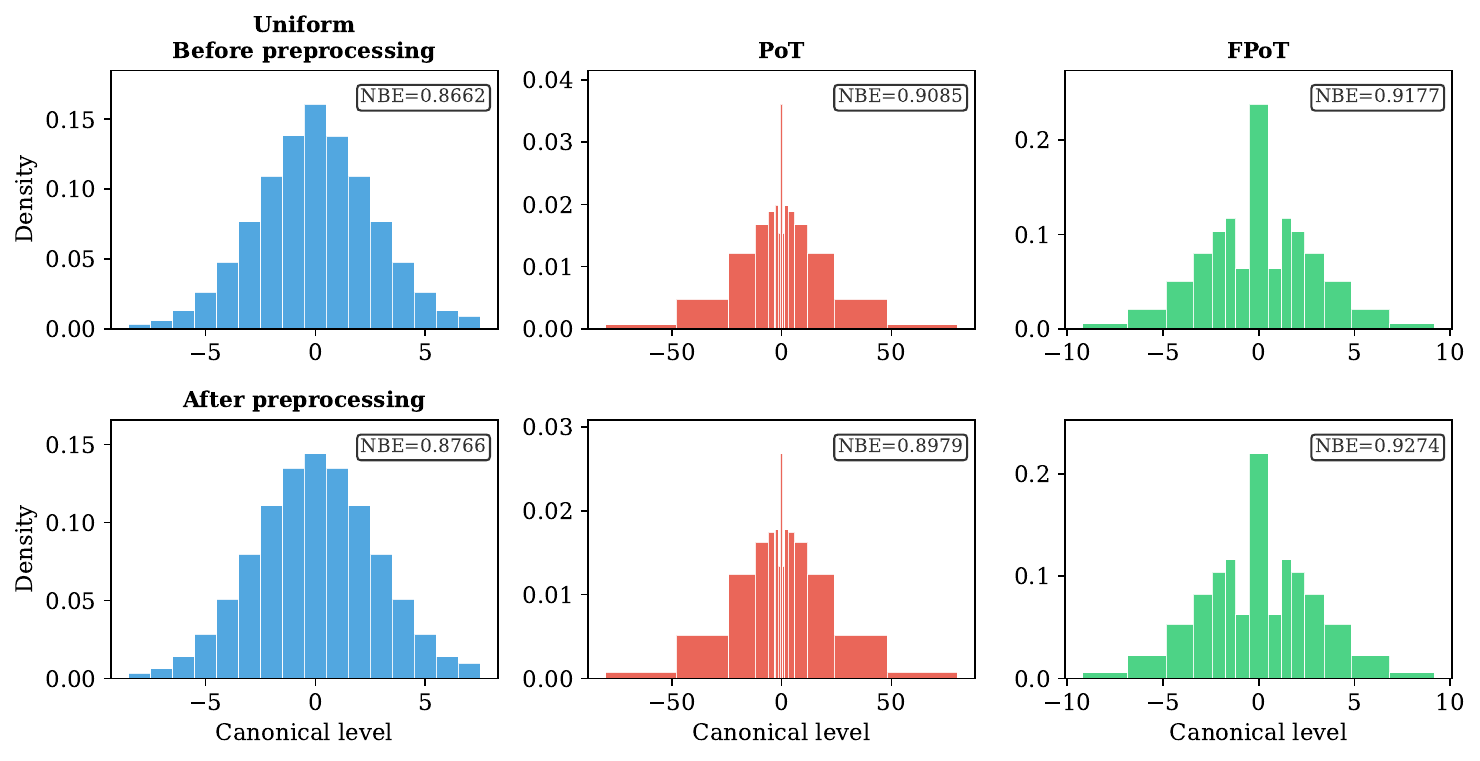}
    \caption{Bin occupancy before (top) and after (bottom) PTQ preprocessing for Llama-2-7B (W4A4KV4). NBE shifts: uniform INT4 0.8662 $\to$ 0.8766; FPoT 0.9177 $\to$ 0.9274.}
    \label{fig:bin-occupancy-before-after-llama2-7b}
\end{figure}

\begin{figure}[ht]
    \centering
    \includegraphics[width=\linewidth]{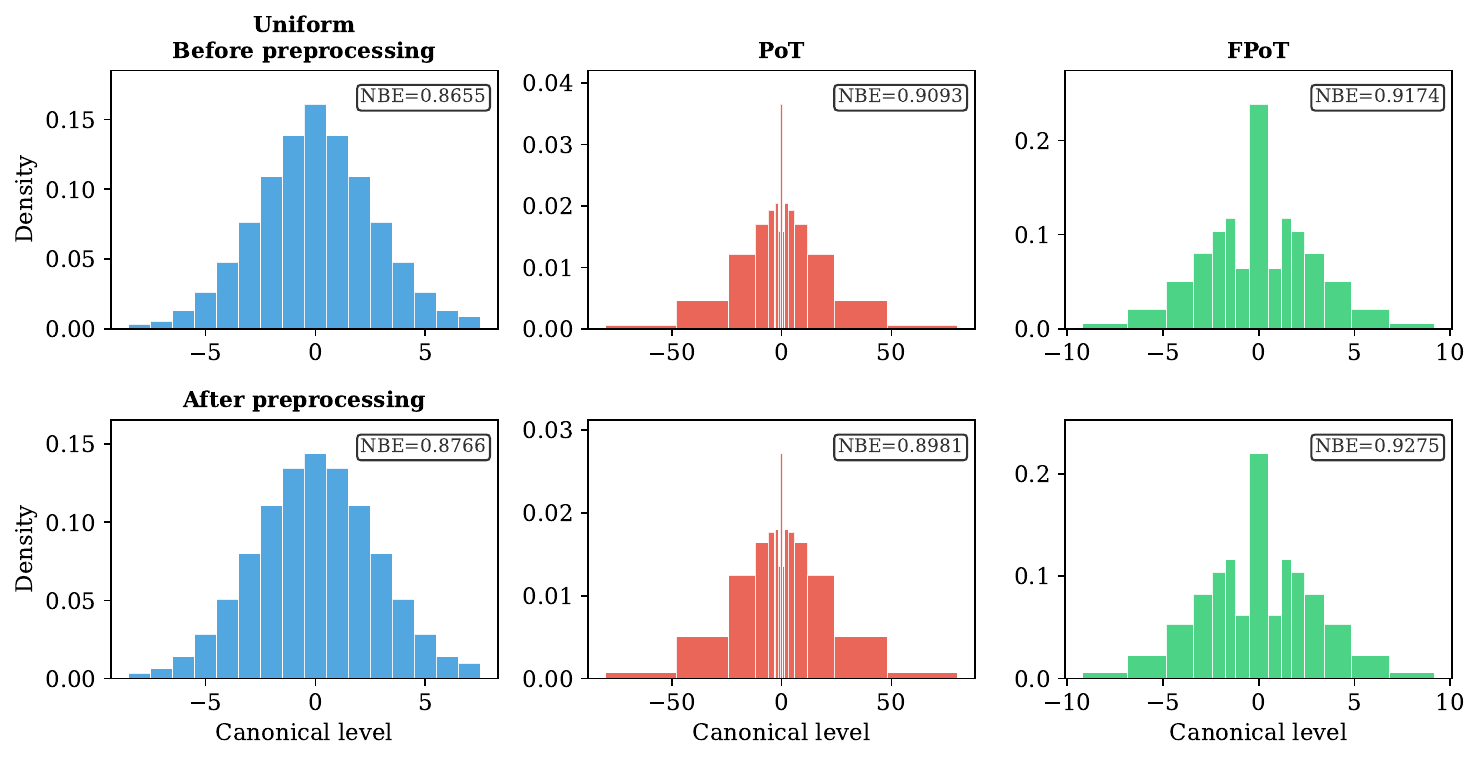}
    \caption{Bin occupancy before (top) and after (bottom) PTQ preprocessing for Llama-2-13B (W4A4KV4). NBE shifts: uniform INT4 0.8655 $\to$ 0.8766; FPoT 0.9174 $\to$ 0.9275.}
    \label{fig:bin-occupancy-before-after-llama2-13b}
\end{figure}

\begin{figure}[ht]
    \centering
    \includegraphics[width=\linewidth]{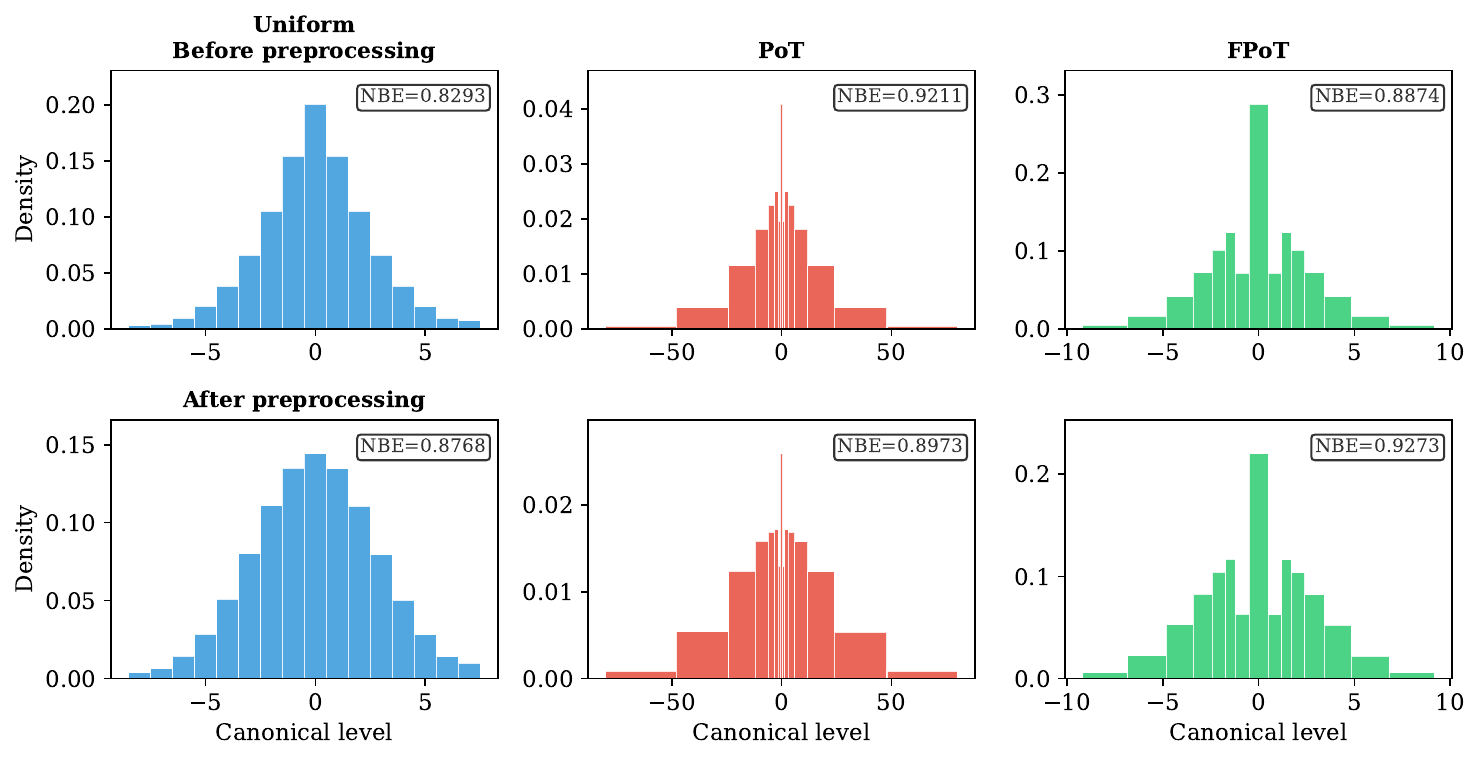}
    \caption{Bin occupancy before (top) and after (bottom) PTQ preprocessing for Qwen2.5-1.5B (W4A4KV4). NBE shifts are large (uniform INT4 0.8293 $\to$ 0.8768; FPoT 0.8874 $\to$ 0.9273), reflecting strong raw-weight outlier concentration.}
    \label{fig:bin-occupancy-before-after-qwen2.5-1.5b}
\end{figure}

\begin{figure}[ht]
    \centering
    \includegraphics[width=\linewidth]{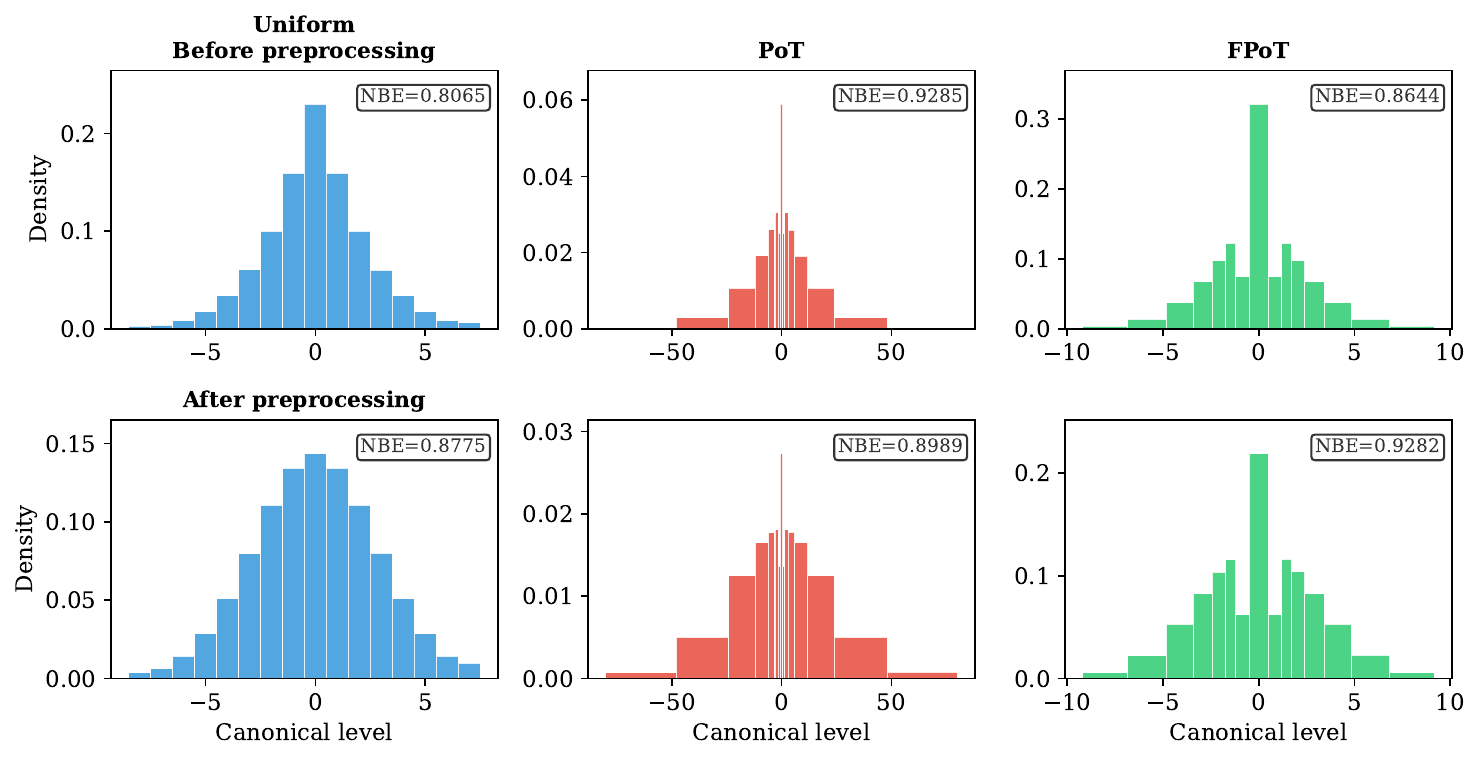}
    \caption{Bin occupancy before (top) and after (bottom) PTQ preprocessing for Qwen2.5-7B (W4A4KV4). NBE shifts: uniform INT4 0.8065 $\to$ 0.8775; FPoT 0.8644 $\to$ 0.9282.}
    \label{fig:bin-occupancy-before-after-qwen2.5-7b}
\end{figure}

\begin{figure}[ht]
    \centering
    \includegraphics[width=\linewidth]{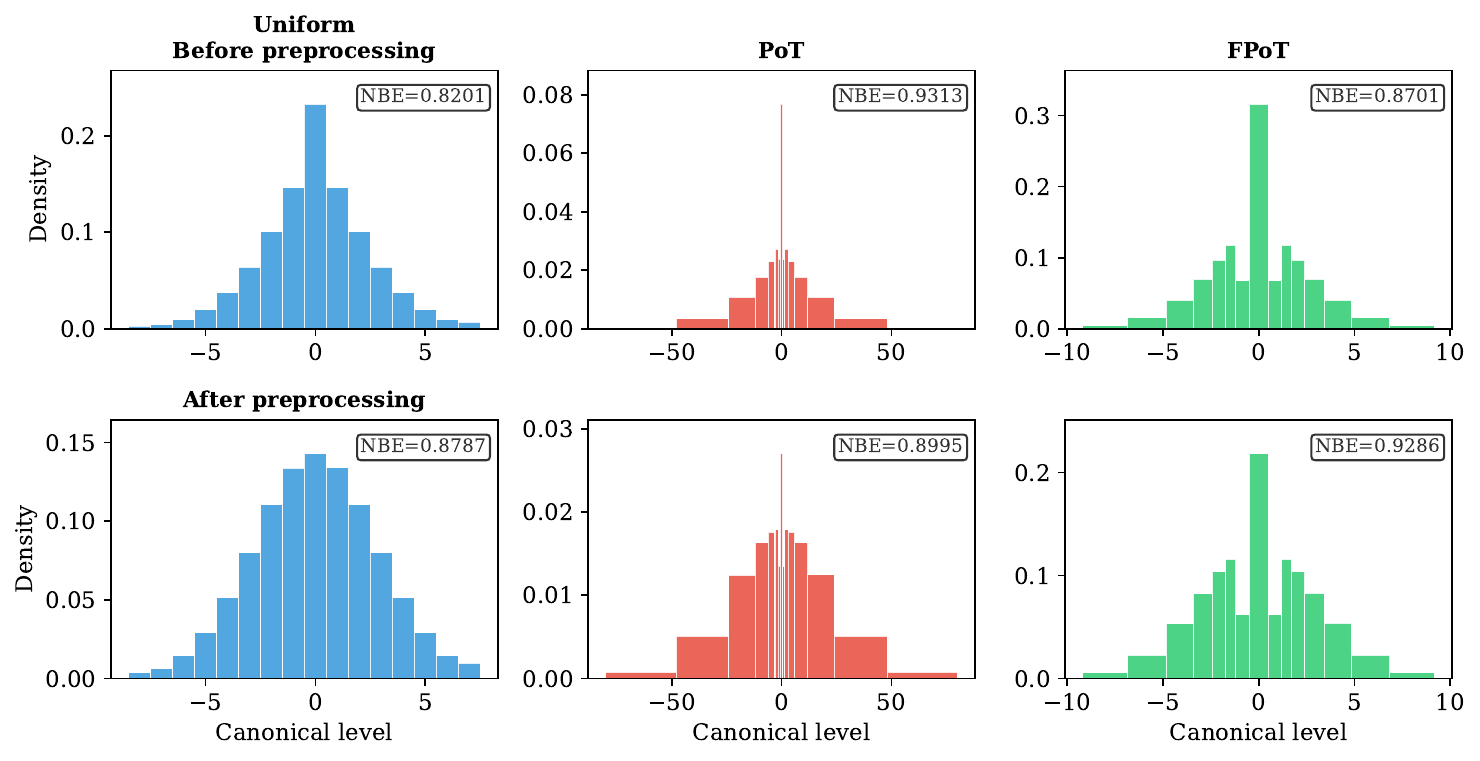}
    \caption{Bin occupancy before (top) and after (bottom) PTQ preprocessing for Qwen2.5-14B (W4A4KV4). NBE shifts: uniform INT4 0.8201 $\to$ 0.8787; FPoT 0.8701 $\to$ 0.9286.}
    \label{fig:bin-occupancy-before-after-qwen2.5-14b}
\end{figure}

Fig.~\ref{fig:bin-occupancy-before-after-llama3-8b}--\ref{fig:bin-occupancy-before-after-qwen2.5-14b} compare the bin occupancy \emph{before} and \emph{after} PTQ preprocessing. A consistent pattern emerges: preprocessing increases NBE for uniform INT4 ($+0.01$ to $+0.07$) and FPoT ($+0.01$ to $+0.06$), while slightly \emph{decreasing} NBE for PoT ($-0.01$ to $-0.03$). This confirms that distribution redistribution specifically benefits grids whose dynamic range is matched to the redistributed weight spread (${\sim}4$ octaves), while the excessively wide PoT grid ($64{:}1$) cannot exploit the narrower post-preprocessing distribution. The effect is most pronounced for Qwen2.5 models, whose raw weights exhibit stronger outlier concentration.

%% file: figures/fig_fpot_grid_float.tex
\begin{figure}[ht]
    \centering
    \includegraphics[width=0.7\linewidth]{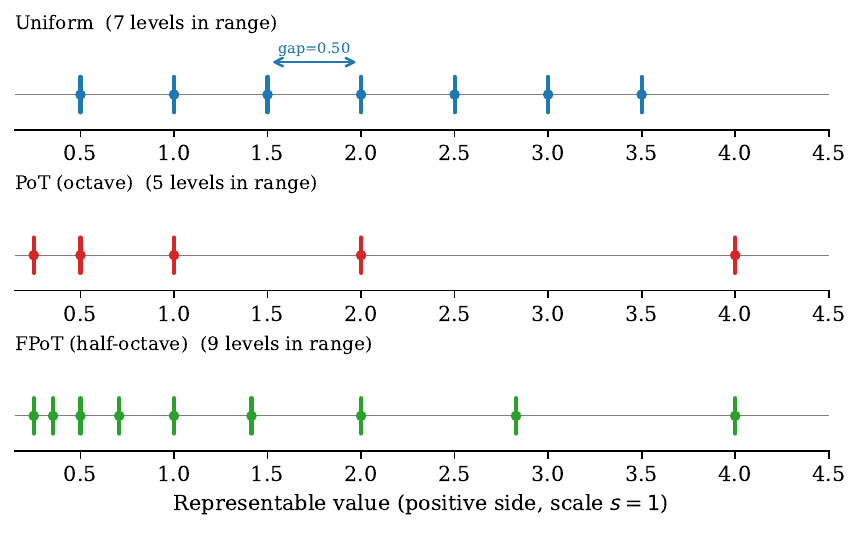}
    \caption{Quantization grid comparison: Uniform, PoT (octave), and FPoT (half-octave). The FPoT grid places $2{\times}$ more levels within the high-density region of the Gaussian weight distribution.}
    \label{fig:fpot-grid}
\end{figure}

%% file: figures/fig_bin_occupancy_llama3-8b_float.tex
\begin{figure}[ht]
    \centering
    \includegraphics[width=\linewidth]{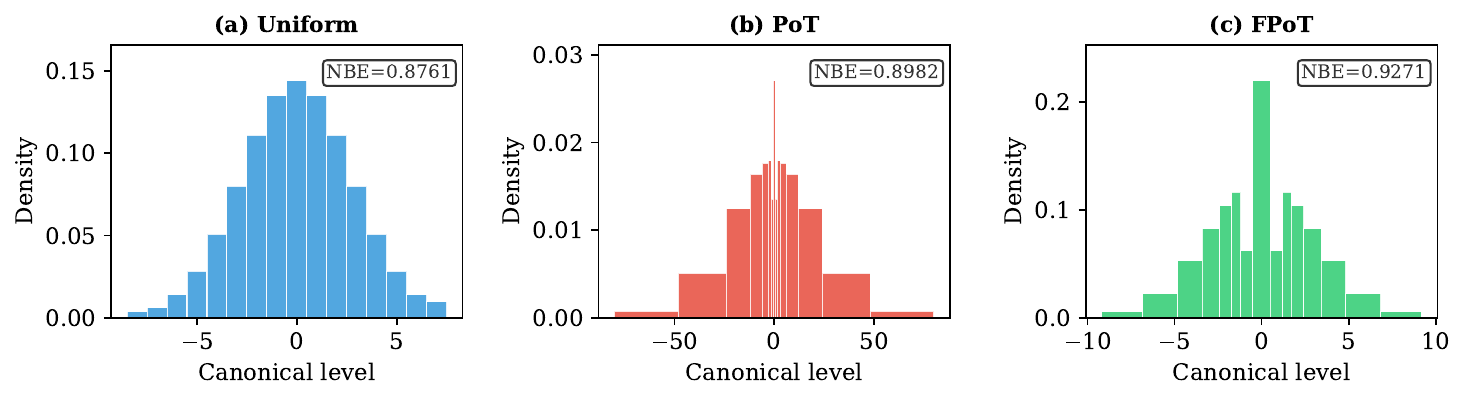}
    \caption{Empirical bin occupancy for Llama-3-8B (W4A4KV4, aggregated across all projections). The pattern closely matches the Gaussian proxy prediction in Fig.~\ref{fig:bin-occupancy-gaussian-body}.}
    \label{fig:bin-occupancy-llama3-8b}
\end{figure}

%% file: sections/D_appendix_per_projection_sensitivity_analysis.tex
\section{Per-Projection Sensitivity Analysis}
\label{app:sensitivity}
\label{app:mac-breakdown}

\subsection{Per-Projection MAC Breakdown}

Table~\ref{tab:mac-breakdown} presents the per-projection MAC breakdown for Llama-3-8B (32 layers, single token); the FFN projections (Gate, Up, Down) dominate at $80.8\%$ of weight MACs.

\input{tables/tab_mac_breakdown}

\paragraph{Prefill versus decode.}
The shares above count weight MACs only, and the unquantized LM head is not counted. Attention adds $2 d_\text{model} c$ MACs per token and layer for QK$^\top$ and PV, where $c$ is the number of attended positions ($c{=}L/2$ on average in causal prefill of length $L$, and the KV length in decode). With this count, linear projections carry $96.3\%$ of all MACs for Llama-3-8B in prefill at $L{=}2$K ($93.7$--$96.9\%$ across the seven models), $86.7\%$ at $L{=}8$K ($78.8$--$88.8\%$), and $76.5\%$ in decode with an 8K KV cache ($65.0$--$79.9\%$). Applying the $0.42$--$0.46{\times}$ per-weight-MAC energy of \S\ref{sec:hw_energy} to this share and keeping attention MACs at INT cost gives a total-MAC energy of $0.44$--$0.48{\times}$ in 2K prefill and $0.56$--$0.59{\times}$ in decode with an 8K KV cache. In memory-bound decode, arithmetic makes up a smaller share of system energy than in prefill, and weight traffic matches uniform INT4 because both store 4-bit weights.

\subsection{Per-Projection FPoT Sensitivity}
\label{sec:fpot-results}
\label{sec:sensitivity}

Table~\ref{tab:fpot_k2} presents the per-projection FPoT sensitivity on Llama-3-8B (W4A4KV4). Each row applies weight-only FPoT to the indicated projection(s) while the rest use uniform INT. The non-PoT baseline achieves PPL\,$= 7.444$ and 0-shot$^9 = 64.98$\%.

\input{tables/tab_fpot_k2}

\textbf{FPoT achieves near-lossless quantization on every individual projection.}
All single-projection entries achieve PPL within $+0.06$ of the non-PoT baseline (largest $+0.052$, Down), with 0-shot$^9$ within $0.4$\%p. Even when applied to all seven projections simultaneously, the total overhead is only $+0.12$ PPL and $-0.7$\%p 0-shot$^9$.

\textbf{Degradation is not additive.} Five projections individually land slightly below the baseline, yet all seven together (PPL\,$= 7.563$) add $+0.12$, so the single-projection entries rank sensitivity rather than form an error budget.

\subsection{Design Implications}

Per-projection sensitivity guides future aggressive compression designs. The Value projection requires uniform INT activations: full FPoT on $W_V$ incurs $\Delta$PPL\,$= +0.30$ versus $+0.04$ for weight-only FPoT (Table~\ref{tab:fpot_k2}), a $7{\times}$ gap driven by post-softmax activation noise. For future full-PoT activation designs, the Output projection is the safest candidate, followed by FFN projections; Value should be last.

%% file: tables/tab_mac_breakdown.tex
\begin{table}[ht]
\caption{MAC Breakdown per Token for Llama-3-8B (32 Layers)}
\label{tab:mac-breakdown}
\centering
\begin{tabular}{lrr}
\hline
\textbf{Projection} & \textbf{MACs (M)} & \textbf{\% of Total} \\
\hline
$W_Q$ & 537 & 7.7\% \\
$W_K$ & 134 & 1.9\% \\
$W_V$ & 134 & 1.9\% \\
$W_O$ & 537 & 7.7\% \\
Gate  & 1{,}879 & 26.9\% \\
Up    & 1{,}879 & 26.9\% \\
Down  & 1{,}879 & 26.9\% \\
\hline
\textit{Attn total} & \textit{1{,}342} & \textit{19.2\%} \\
\textit{FFN total}  & \textit{5{,}637} & \textit{80.8\%} \\
\hline
\textbf{Total} & \textbf{6{,}979} & \textbf{100\%} \\
\hline
\end{tabular}
\end{table}

%% file: tables/tab_fpot_k2.tex
\begin{table}[ht]
\centering
\caption{%
  Per-projection FPoT weight-only ablation on Llama-3-8B (W4A4KV4, 400~steps, single seed).
  Each row applies FPoT to the indicated projection(s); all others use uniform INT.
  Activations use uniform INT4 throughout.
}
\label{tab:fpot_k2}
\setlength{\tabcolsep}{4pt}
\begin{tabular}{lrc}
\toprule
FPoT Applied To & PPL${\downarrow}$ & 0-shot$^9{\uparrow}$ (\%) \\
\midrule
None (non-PoT baseline)  &  7.444 & 64.98 \\
\midrule
$W_Q$ only              &  7.379 & 65.59 \\
$W_K$ only              &  7.393 & 64.77 \\
$W_V$ only              &  7.487 & 65.45 \\
$W_O$ only              &  7.337 & 65.21 \\
Gate only               &  7.381 & 65.45 \\
Up only                 &  7.386 & 65.77 \\
Down only               &  7.496 & 64.59 \\
\midrule
Attention ($W_Q$+$W_K$+$W_V$+$W_O$) &  7.419 & 65.13 \\
FFN (Gate+Up+Down)       &  7.530 & 64.74 \\
\textbf{All 7 projections} & \textbf{7.563} & \textbf{64.28} \\
\bottomrule
\end{tabular}
\end{table}

%% file: sections/E_appendix_hardware_efficiency_analysis_details.tex
\section{Hardware Efficiency Analysis Details}
\label{app:hw_details}

This appendix documents the methodology underlying the hardware cost estimates in Section~\ref{sec:hw-analysis} and presents per-model accuracy--cost trade-offs.

\subsection{Na\"{\i}ve Component-Based Cost Model}
\label{app:hw_derivation}

We adopt a simple component-counting model with two unit costs: $c_{\mathrm{fa}} = 1.0$ (full-adder equivalent) for arithmetic logic, and $c_{\mathrm{mux}} = 0.5 \times c_{\mathrm{fa}}$ for 2-to-1 multiplexer stages (reflecting the lower transistor count of a pass-gate MUX vs.\ a full adder). All costs are per-bit of the datapath width $A$ (activation bits).

\paragraph{INT multiplier.}
A $W{\times}A$ signed multiplier generates $W{\times}A$ partial products and reduces them through an adder tree. Complexity grows \emph{quadratically}:
\[
C_{\text{INT}} = W \times A.
\]

\paragraph{PoT barrel shifter.}
A barrel shifter with $(W{-}1)$-bit shift control implements $(W{-}1)$ cascaded MUX stages, each containing $A$ two-input multiplexers:
\[
C_{\text{PoT}} = (W{-}1) \times A \times c_{\mathrm{mux}} = (W{-}1) \times A \times 0.5.
\]
Complexity grows \emph{linearly} in both $W$ and $A$, which is the fundamental source of PoT's hardware advantage.

\paragraph{FPoT shifter + adder.}
FPoT uses a barrel shifter with one fewer stage (shift amount $= \lfloor e/2 \rfloor$, requiring $W{-}2$ control bits), plus a 1-adder $\sqrt{2}$ approximation unit ($x + x{\gg}1 = 1.5x$, 6.1\% error). The constant right-shift ($\gg 1$) is a hard-wired reconnection requiring zero logic gates. The 6.1\% coefficient error is absorbed by \emph{approximate-grid quantization} (Section~\ref{sec:sqrt2-approx}):
\[
C_{\text{FPoT}} = \underbrace{(W{-}2) \times A \times 0.5}_{\text{barrel shifter}} + \underbrace{1 \times A}_{\sqrt{2}\text{ adder}} = 0.5 \cdot A \cdot W.
\]

\paragraph{Accumulator.}
All three modes share an identical $(W{+}A)$-bit accumulator adder:
\[
C_{\text{ACC}} = (W{+}A) \times c_{\mathrm{fa}} = W + A.
\]

\begin{table}[h]
\centering
\small
\caption{Na\"{\i}ve component-based cost model for INT, PoT, and FPoT
compute units and full MACs (Section~\ref{sec:hw-analysis}).
Values are normalized to W4A4 INT compute unit~$= 1.00{\times}$,
using $c_{\mathrm{mux}} = 0.5{\times}c_{\mathrm{fa}}$
(full derivation in \S\ref{app:hw_derivation}).
The accumulator ($C_{\text{ACC}} = W{+}A$) is identical across all three
modes and added to obtain the full-MAC cost.}
\label{tab:hw_analytical}
\setlength{\tabcolsep}{6pt}
\begin{tabular}{lccc cc}
\toprule
\textbf{Config} & \textbf{INT} & \textbf{PoT} & \textbf{FPoT} & \textbf{PoT/INT} & \textbf{FPoT/INT} \\
\midrule
\multicolumn{6}{l}{\emph{Compute unit only}} \\
W4A4 & 1.00 & 0.38 & 0.50 & $0.38{\times}$ & $0.50{\times}$ \\
W4A8 & 2.00 & 0.75 & 1.00 & $0.38{\times}$ & $0.50{\times}$ \\
W8A8 & 4.00 & 1.75 & 2.00 & $0.44{\times}$ & $0.50{\times}$ \\
\midrule
\multicolumn{6}{l}{\emph{Full MAC (compute unit + accumulator)}} \\
W4A4 & 1.50 & 0.88 & 1.00 & $0.58{\times}$ & $0.67{\times}$ \\
W4A8 & 2.75 & 1.50 & 1.75 & $0.55{\times}$ & $0.64{\times}$ \\
W8A8 & 5.00 & 2.75 & 3.00 & $0.55{\times}$ & $0.60{\times}$ \\
\bottomrule
\end{tabular}
\end{table}

The INT multiplier grows as $O(WA)$ (quadratic), while both PoT and FPoT grow as $O(W + A)$ (linear). The cost advantage therefore \emph{increases} at higher bit-widths.

\begin{remark}[Assumptions and Limitations]\label{app:hw_limitations}
The na\"{\i}ve model makes several simplifying assumptions: (1)~unit cost approximation ($c_{\mathrm{mux}} = 0.5 \times c_{\mathrm{fa}}$); (2)~no synthesis optimization; (3)~no energy modeling (gate counts as area proxy only); (4)~technology-node independence; (5)~omitted common logic (sign negation, zero-gating), which are shared across all three modes and implemented in the accumulator control path rather than the compute unit, so their omission does not affect the CU cost ratios reported above; (6)~the FPoT expression omits the even/odd output MUX and counts the $\sqrt{2}$ adder at activation width $A$, although the shifted operand it adds is wider. Assumption~(6) makes the analytical FPoT cost, and with it the FPoT/INT ratio, optimistic, and \S\ref{app:hw_yosys} reports the synthesized $0.57$--$0.59{\times}$ at W4. Despite these simplifications, the model correctly captures the fundamental scaling argument: multiplier cost is quadratic in bit-width while shifter + adder cost is linear, confirmed by the synthesis results.
\end{remark}

\subsection{Open-source Synthesis Results}
\label{app:hw_yosys}
To corroborate the structural claims of the na\"{\i}ve model, we synthesize
the three compute units using Yosys~0.9~\citep{wolf2013yosys} with ABC generic-gate mapping
(\texttt{abc~-g~gates}).
Sign negation and zero-gating are implemented in the accumulator control
path---consistent with Remark~\ref{app:hw_limitations}, assumption~(5)---and
are therefore excluded from compute-unit gate counts.
Table~\ref{tab:yosys_gates} reports gate counts for all nine
(config\,$\times$\,mode) combinations.

Three observations are consistent with the analytical model.
First, the structural ordering INT\,$>$\,FPoT\,$>$\,PoT holds for every
configuration, confirming that the 1-adder $\sqrt{2}$ unit adds only moderate
overhead relative to the INT multiplier.
Second, the INT gate count grows faster with bit-width than FPoT:
scaling from W4A4 to W8A8 (compute unit), INT grows $5.2{\times}$ while
FPoT grows $3.8{\times}$, confirming that FPoT's cost advantage
\emph{widens} at higher bit-widths.
Third, the full-MAC PoT/INT ratio decreases from $0.50{\times}$ at W4A4 to
$0.37{\times}$ at W8A8, consistent with the quadratic-versus-linear scaling
prediction.

Absolute ratios differ from Table~\ref{tab:hw_analytical} in two respects.
PoT/INT at W4A4 (compute unit) measures $0.24{\times}$ vs.\ the analytical
$0.38{\times}$: the ABC optimizer aggressively shares gates in the
barrel-shifter MUX chain, making the analytical model a \emph{conservative}
estimate of PoT's advantage.
FPoT/INT at W4 (compute unit) measures $0.57$--$0.59{\times}$ against the
analytical $0.50{\times}$. Ablative synthesis locates the difference. At W4A4
the FPoT compute unit splits into barrel shifter $10$, 1-adder $26$ and
even/odd output MUX $8$ gates, summing to the $44$ of
Table~\ref{tab:yosys_gates} (W4A8: $18$/$55$/$14$, sum $87$).
The expression for $C_{\text{FPoT}}$ in \S\ref{app:hw_derivation} has no term
for this output MUX. Without it the unit has $36$ gates at W4A4, below the
$37.5$ implied by the analytical $0.50{\times}$ (W4A8: $73$ against $77$),
so the MUX rather than the shift path separates the measured ratio from the
analytical one.
The Yosys results therefore support the core structural claim of the
analytical model---the INT-vs.-FPoT cost gap is real and widens with
bit-width---while absolute values should be interpreted with the caveats
of Remark~\ref{app:hw_limitations}.
Quantitatively, the cosine similarity between the analytical and
Yosys-measured cost vectors (all 18 entries of Table~\ref{tab:hw_analytical}
and Table~\ref{tab:yosys_gates}, normalized to W4A4 INT CU) is
$\mathbf{0.988}$, confirming near-perfect agreement in structural ordering
and bit-width scaling direction despite differences in absolute ratios.

\input{tables/tab_yosys_gates}

\subsection{MAC Energy Ratio: Published Synthesis Anchor}

The energy estimates in \S\ref{sec:hw_energy} are anchored to the only published RTL-level ASIC synthesis comparing INT and PoT MAC units at identical conditions: \citet{przewlocka2022pot} (Samsung 5\,nm, 250\,MHz, TT corner, 0.6\,V, 25\textdegree C, 50\% toggle rate). Relative to an 8$\times$8 uniform MAC baseline, they report W4A8 uniform MAC at $2.5{\times}$ energy reduction and W4A8 PoT MAC at $6.0{\times}$. The PoT-to-INT ratio for the \emph{same} W4A8 precision is therefore:
\[
\frac{E_{\text{PoT,W4A8}}}{E_{\text{INT,W4A8}}} = \frac{2.5}{6.0} \approx \mathbf{0.42}{\times}.
\]

FPoT adds the $\times\!\sqrt{2}$ unit (one adder, active for ${\sim}50\%$ of MACs). We report the FPoT/INT energy ratio as a range: \textbf{$0.42$--$0.46{\times}$}, where the lower bound assumes negligible $\sqrt{2}$ switching energy and the upper bound assumes proportional contribution.

\paragraph{Applicability to LLMs.}
Although \citet{przewlocka2022pot} synthesize CNN-oriented MAC units (ResNet/VGG benchmarks), the reported per-MAC energy ratio is model-agnostic: the synthesized block---a $4{\times}8$ shift-add unit vs.\ a $4{\times}8$ uniform multiplier---is structurally identical to the per-weight MAC in our LLM datapath. Model-level differences (sequence length, attention, layer count) affect only the aggregate MAC count, not the per-MAC energy, so the $0.42{\times}$ ratio transfers directly to LLM inference. Note that this anchor reports per-MAC energy of the arithmetic unit alone; total LLM inference energy additionally includes memory-access, accumulator, and activation-path contributions, the relative weight of which varies with batch size and sequence length. The per-MAC ratio therefore sets the achievable upper bound on weight-arithmetic energy savings rather than the total system-level reduction.

\subsection{Per-Model Accuracy--Cost Trade-offs}
\label{app:hw_tradeoff}

The per-model figures below present accuracy--cost Pareto trade-offs across compute unit (CU) cost and full MAC cost (CU + accumulator), normalized to W4A4 Uniform INT $= 1.0{\times}$.

The figures reveal a consistent pattern across all seven models: (1)~FPoT consistently approaches the Uniform INT accuracy at $0.50{\times}$ CU cost; (2)~the shared accumulator compresses the advantage at full-MAC level ($0.67{\times}$ at W4A4); (3)~as bit-widths increase, the quadratic/linear scaling gap widens, increasing FPoT's advantage; (4)~all models share the same $x$-axis coordinates (hardware is model-independent), with variation only in accuracy ($y$-axis). Per-model Pareto curves for Llama-3-8B, Llama-2-7B/13B, Mistral-7B, and Qwen2.5-1.5B/7B/14B are shown in Figs.~\ref{fig:tradeoff-llama3-8b}--\ref{fig:tradeoff-qwen2.5-14b}, respectively.

\input{figures/fig_tradeoff_all}

%% file: tables/tab_yosys_gates.tex
\begin{table}[h]
\centering
\small
\caption{Gate counts from Yosys\,+\,ABC open-source synthesis
(generic-gate library, \texttt{abc~-g~gates}),
with relative cost in parentheses normalized to
W4A4 INT compute unit~$= 1.00{\times}$ (matching Table~\ref{tab:hw_analytical}).
Sign negation and zero-gating reside in the accumulator control path
and are excluded from compute-unit counts
(Remark~\ref{app:hw_limitations}, assumption~5).}
\label{tab:yosys_gates}
\setlength{\tabcolsep}{6pt}
\begin{tabular}{l ccc cc}
\toprule
\textbf{Config}
  & \textbf{INT} & \textbf{PoT} & \textbf{FPoT}
  & \textbf{PoT/INT} & \textbf{FPoT/INT} \\
\midrule
\multicolumn{6}{l}{\emph{Compute unit only}} \\
W4A4 & 75\ {\small(1.00)} & 18\ {\small(0.24)} &  44\ {\small(0.59)} & $0.24{\times}$ & $0.59{\times}$ \\
W4A8 & 154\ {\small(2.05)} & 30\ {\small(0.40)} &  87\ {\small(1.16)} & $0.19{\times}$ & $0.57{\times}$ \\
W8A8 & 391\ {\small(5.21)} & 92\ {\small(1.23)} & 167\ {\small(2.23)} & $0.24{\times}$ & $0.43{\times}$ \\
\midrule
\multicolumn{6}{l}{\emph{Full MAC (compute unit + accumulator)}} \\
W4A4 & 119\ {\small(1.59)} &  60\ {\small(0.80)} &  91\ {\small(1.21)} & $0.50{\times}$ & $0.77{\times}$ \\
W4A8 & 216\ {\small(2.88)} &  95\ {\small(1.27)} & 161\ {\small(2.15)} & $0.44{\times}$ & $0.75{\times}$ \\
W8A8 & 487\ {\small(6.49)} & 180\ {\small(2.40)} & 271\ {\small(3.61)} & $0.37{\times}$ & $0.56{\times}$ \\
\bottomrule
\end{tabular}
\end{table}

%% file: figures/fig_tradeoff_all.tex
\begin{figure*}[ht]
    \centering
    \includegraphics[width=\linewidth]{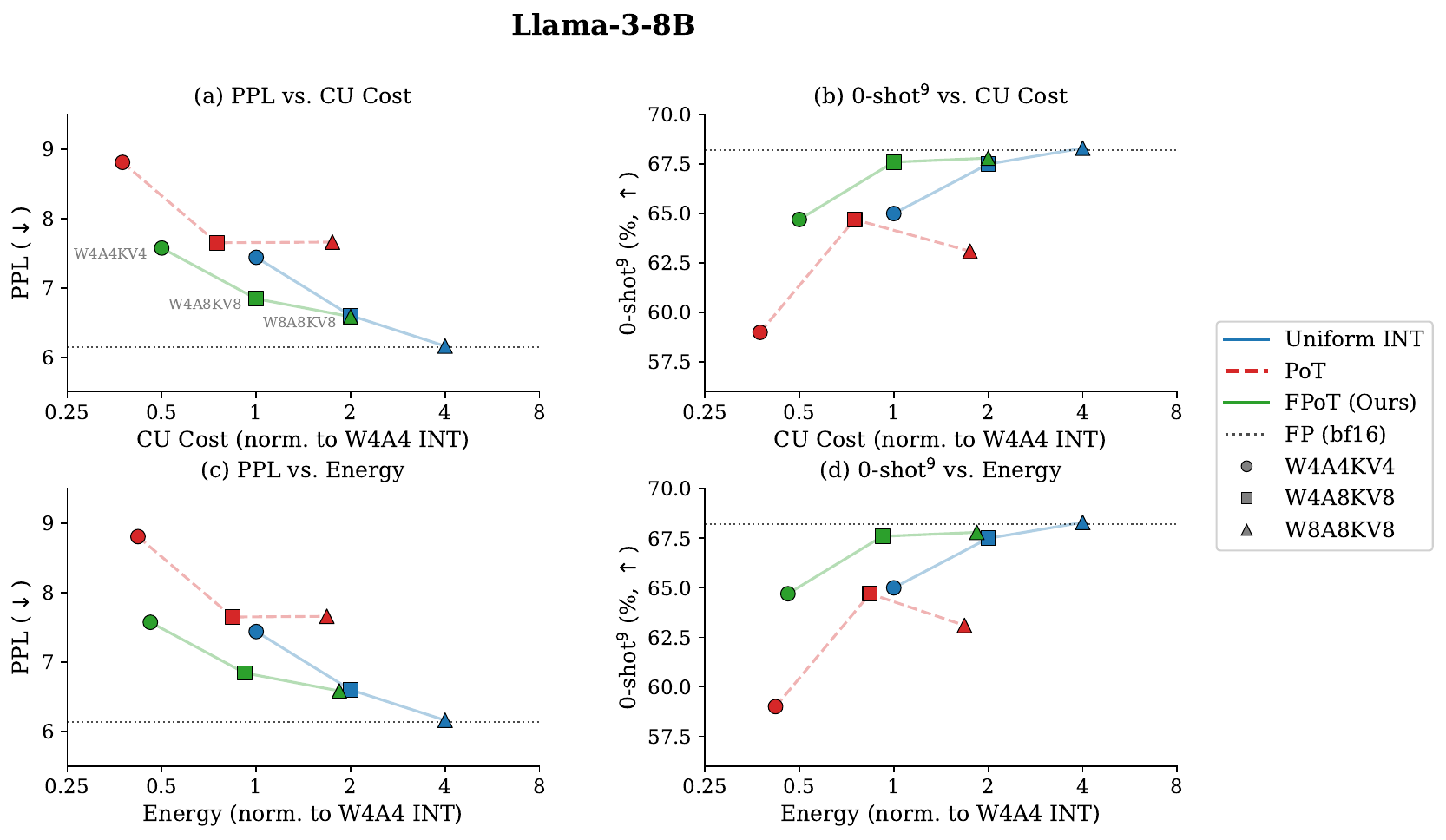}
    \caption{Accuracy--cost Pareto trade-off: Llama-3-8B.}
    \label{fig:tradeoff-llama3-8b}
\end{figure*}

\begin{figure*}[ht]
    \centering
    \includegraphics[width=\linewidth]{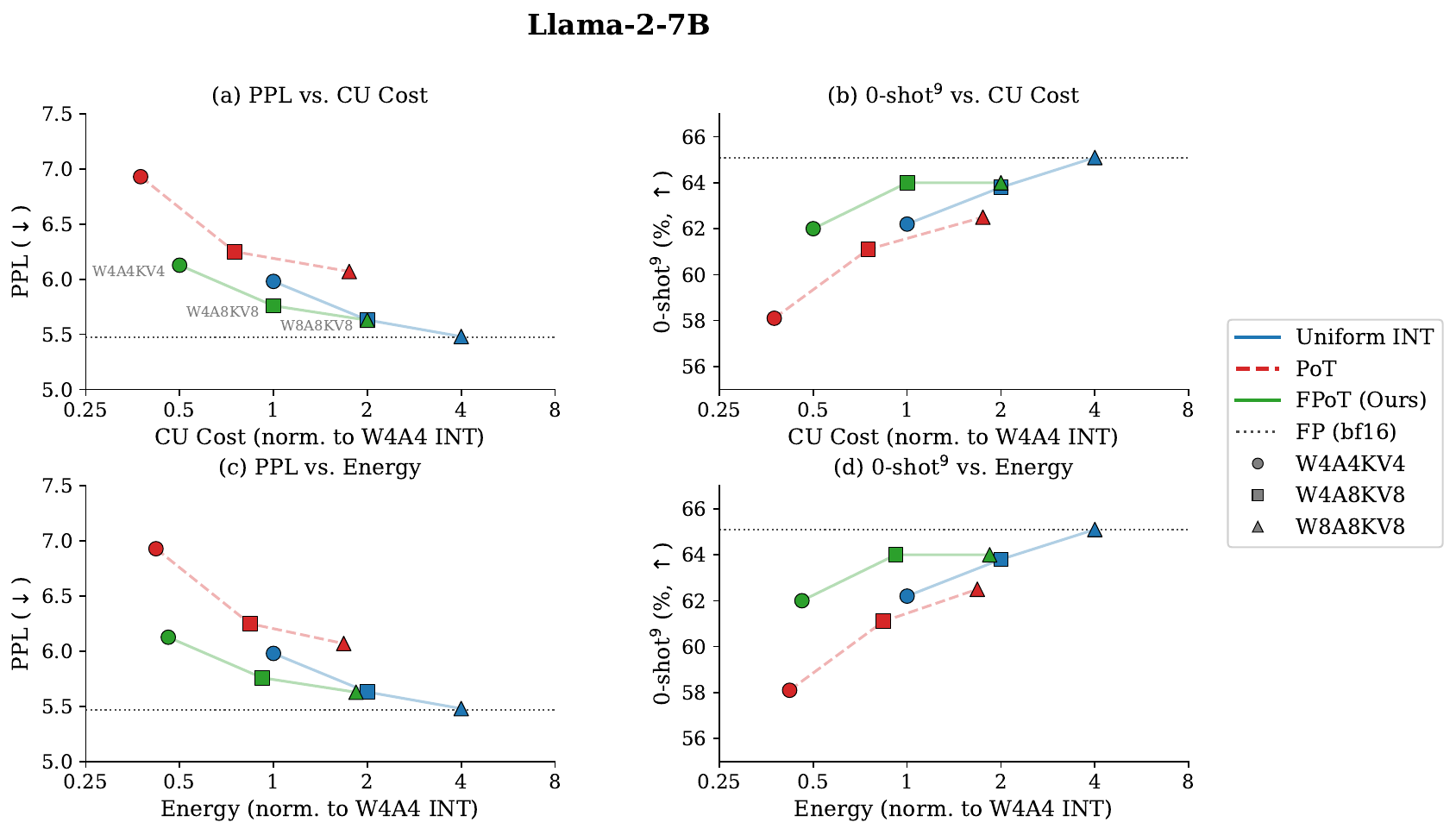}
    \caption{Accuracy--cost Pareto trade-off: Llama-2-7B.}
    \label{fig:tradeoff-llama2-7b}
\end{figure*}

\begin{figure*}[ht]
    \centering
    \includegraphics[width=\linewidth]{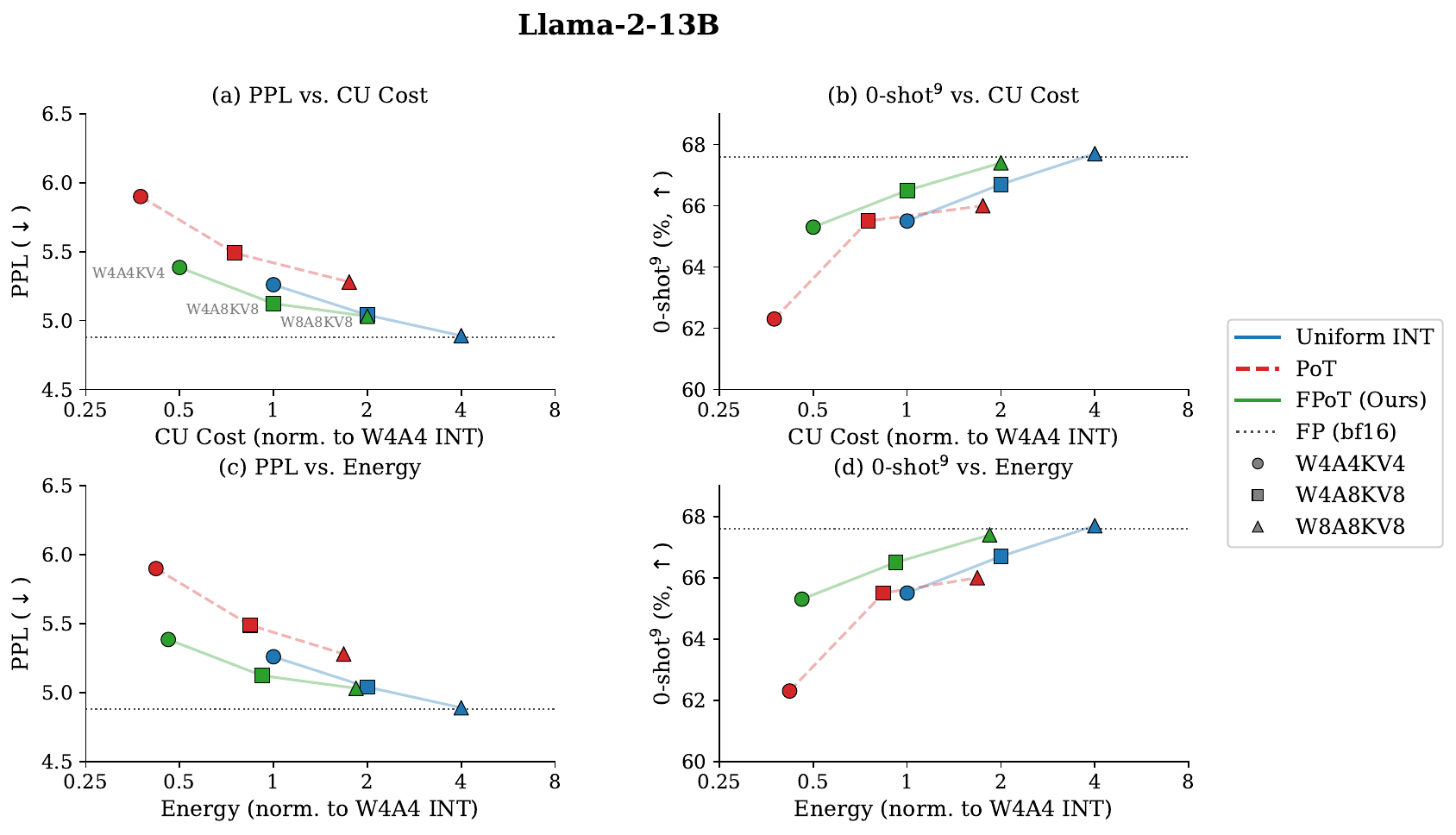}
    \caption{Accuracy--cost Pareto trade-off: Llama-2-13B.}
    \label{fig:tradeoff-llama2-13b}
\end{figure*}

\begin{figure*}[ht]
    \centering
    \includegraphics[width=\linewidth]{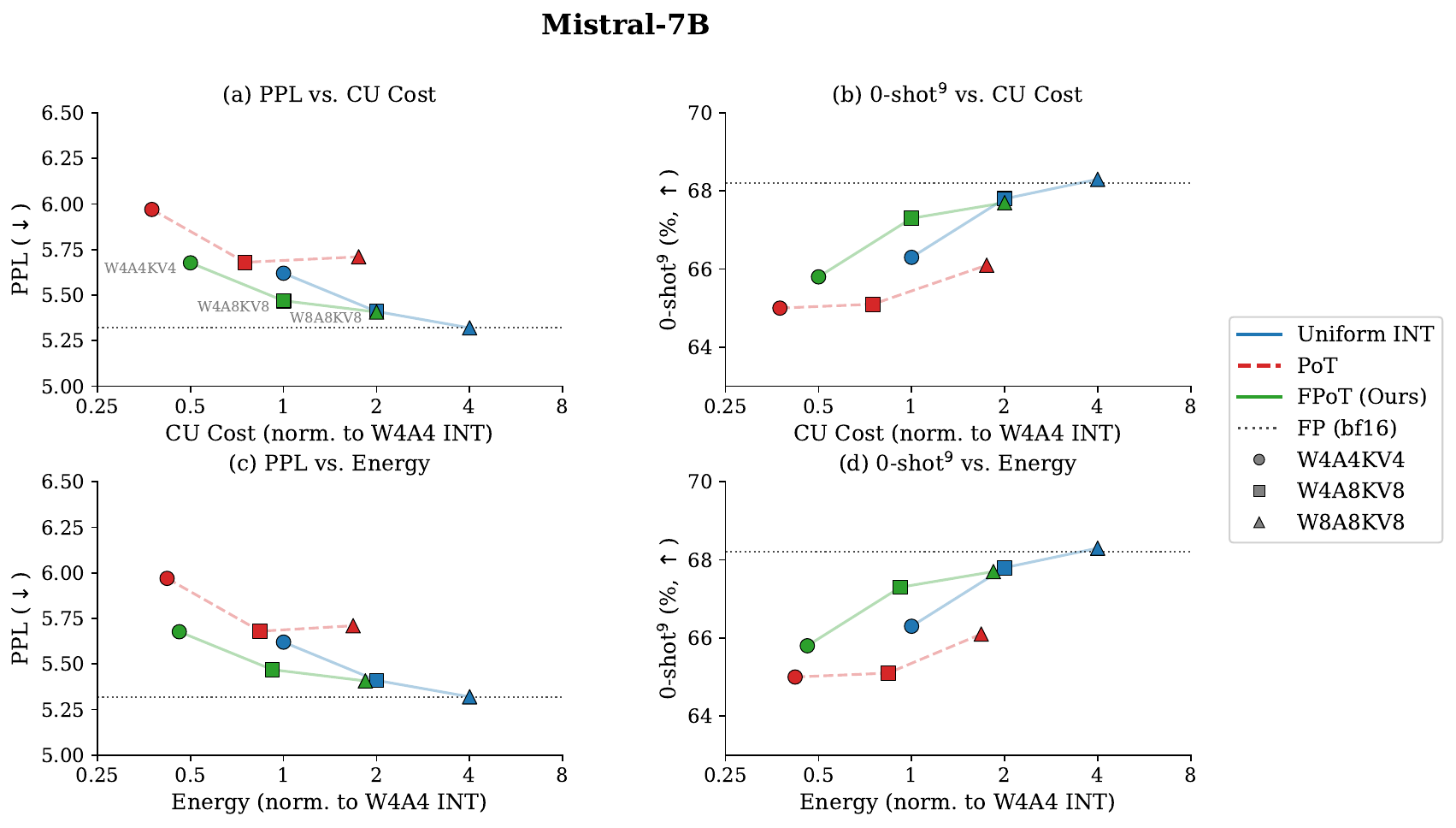}
    \caption{Accuracy--cost Pareto trade-off: Mistral-7B.}
    \label{fig:tradeoff-mistral-7b}
\end{figure*}

\begin{figure*}[ht]
    \centering
    \includegraphics[width=\linewidth]{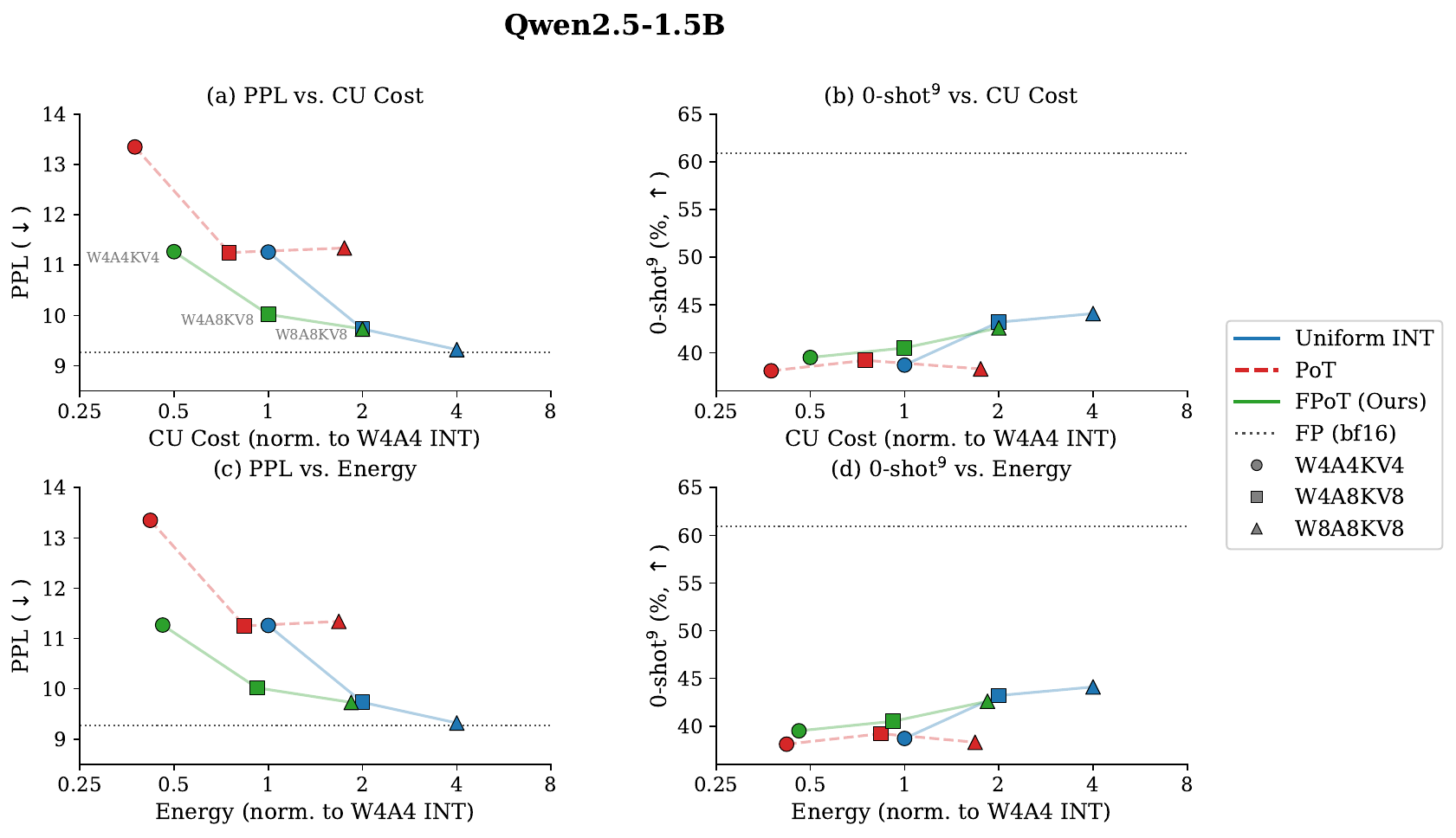}
    \caption{Accuracy--cost Pareto trade-off: Qwen2.5-1.5B.}
    \label{fig:tradeoff-qwen2.5-1.5b}
\end{figure*}

\begin{figure*}[ht]
    \centering
    \includegraphics[width=\linewidth]{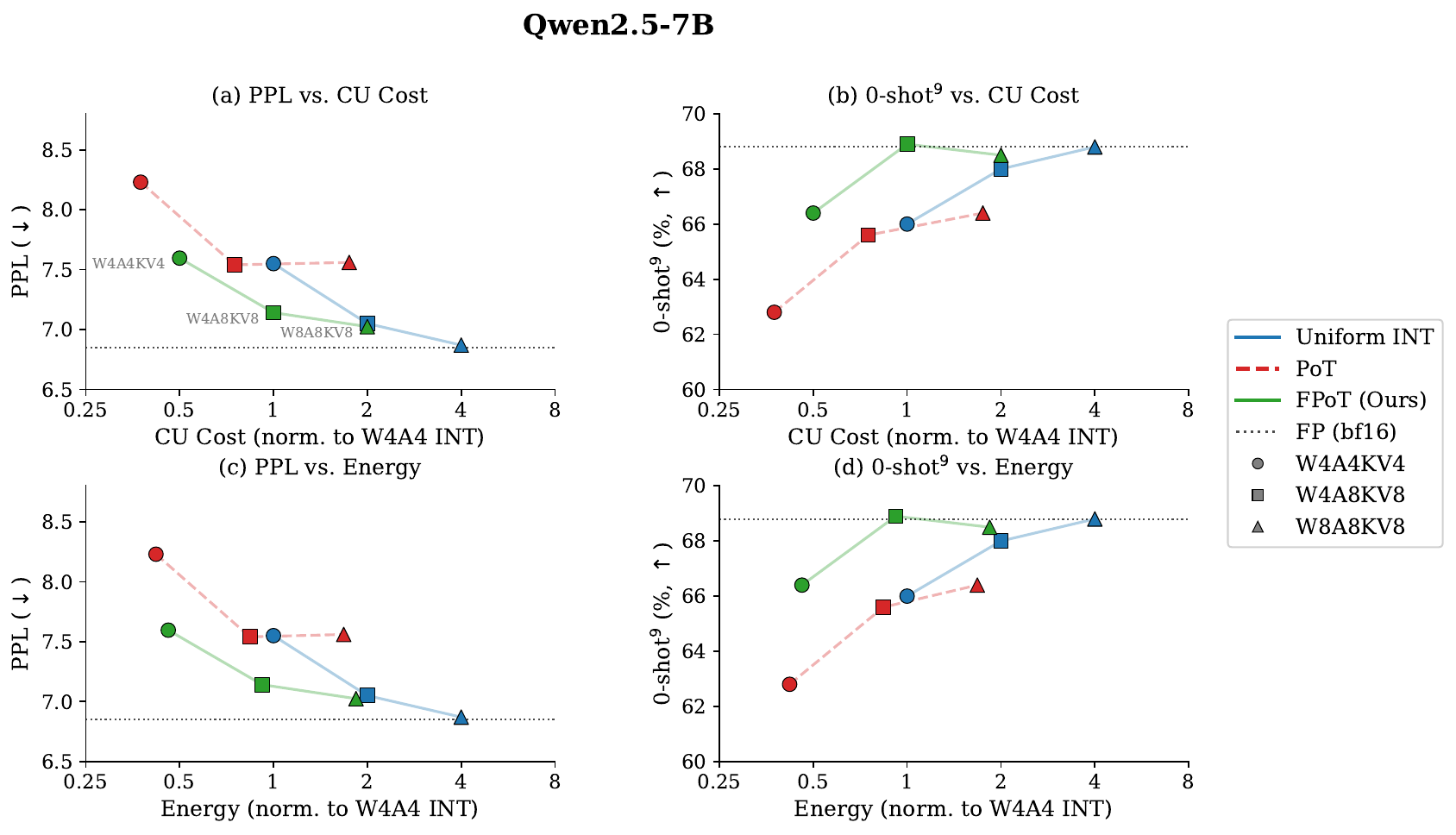}
    \caption{Accuracy--cost Pareto trade-off: Qwen2.5-7B.}
    \label{fig:tradeoff-qwen2.5-7b}
\end{figure*}

\begin{figure*}[ht]
    \centering
    \includegraphics[width=\linewidth]{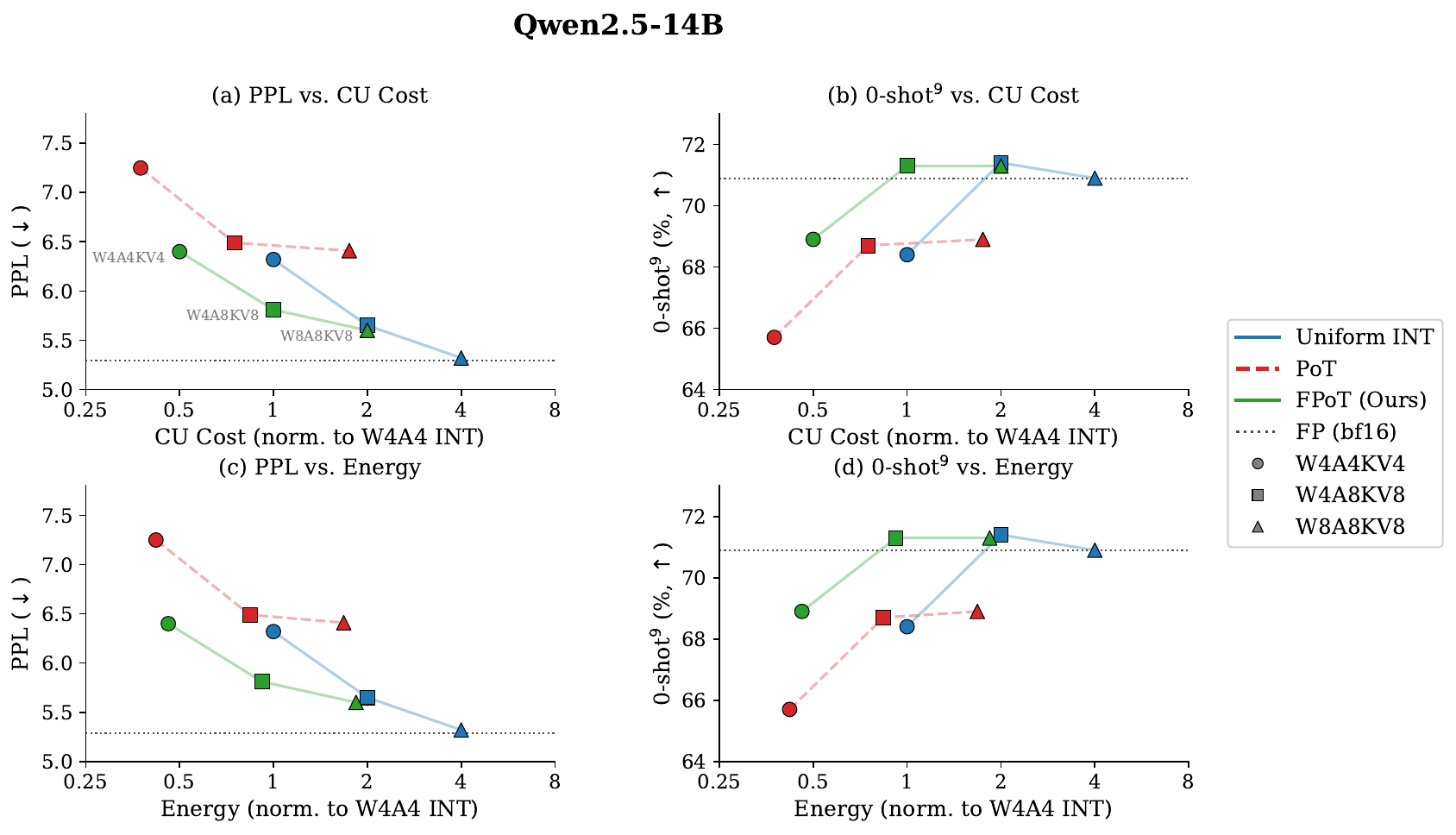}
    \caption{Accuracy--cost Pareto trade-off: Qwen2.5-14B.}
    \label{fig:tradeoff-qwen2.5-14b}
\end{figure*}

%% file: sections/F_appendix_fused_triton_kernel.tex
\section{Fused Triton Kernel for FPoT Training}
\label{app:triton-kernel}

The FPoT fake-quantization pipeline (quantize $\to$ dequantize $\to$ STE) generates ${\sim}8$ intermediate tensors per weight matrix when implemented in standard PyTorch---sign, absolute value, log$_2$ result, exponent, zero/kill masks, and dequantized output---each the same size as the weight.
During gradient-checkpointed backward passes, these intermediates are regenerated, creating a memory peak that exceeds single-GPU capacity for models above ${\sim}13$B parameters on 80\,GB GPUs.

We implement a \textbf{fused Triton kernel} that performs the entire pipeline in a single kernel launch:

\begin{enumerate}
    \item \textbf{Quantize}: compute $|x/s|$, apply zero threshold ($2^{-0.5/k}$), compute $\text{round}(\log_2(\cdot) \times k)$, clamp exponents, determine sign and kill masks.
    \item \textbf{Dequantize}: compute $s \cdot \text{sign} \cdot 2^{e/k}$.
    \item \textbf{STE}: output $x_{\text{fake}}$; backward passes gradient through unchanged via a custom \texttt{torch.autograd.Function}.
\end{enumerate}

All intermediate values reside in GPU registers and are never materialized as global-memory tensors.
The kernel handles per-channel broadcast of scale, $n$, and $m$ parameters natively through stride-based indexing, avoiding \texttt{expand\_as().contiguous()} copies that would otherwise duplicate these tensors to full weight size.

\paragraph{Memory savings.}
On the largest single-projection weight in Qwen2.5-14B (\texttt{gate\_proj}: $18{,}944 \times 5{,}120$), the fused kernel reduces peak intermediate memory from 1.07\,GB to 0.19\,GB---an \textbf{82\% reduction}.
This enables FPoT training of 14B-parameter models on a single A100-80GB with batch size 4, previously infeasible due to OOM during gradient checkpoint recomputation.

\paragraph{Numerical fidelity.}
The kernel performs all arithmetic in float32, matching the precision-critical $\log_2$ and $2^{(\cdot)}$ operations.
On a full-size weight tensor, $99.6\%$ of elements match the PyTorch reference exactly; the remaining $0.4\%$ differ by at most one quantization level due to bf16$\to$float32 promotion order differences, with the Triton path being strictly more precise.
A pure-PyTorch fallback is provided for environments without Triton.

\paragraph{Pseudocode.}
Algorithm~\ref{alg:fpot-quantize} details the complete FPoT quantize/dequantize/STE pipeline.

\input{figures/alg_fpot_quantize}

%% file: figures/alg_fpot_quantize.tex
\begin{figure}[t]
\begin{algorithm}[H]
\caption{Signed FPoT Quantization with $k$-Adaptive Zero Threshold}
\label{alg:fpot-quantize}
\begin{algorithmic}[1]
\Require Weight tensor $\mathbf{W} \in \mathbb{R}^{C_{\mathrm{out}} \times C_{\mathrm{in}}}$, bit budget $b$, resolution parameter $k$, clip ratio $\alpha \in (0, 1]$
\Ensure Quantized weight codes $\mathbf{M} \in \{0,\ldots,2^{b-1}{-}1\}^{C_{\mathrm{out}} \times C_{\mathrm{in}}}$, signs $\mathbf{S} \in \{-1,0,+1\}$, per-channel scales $\mathbf{s} \in \mathbb{R}^{C_{\mathrm{out}}}$
\Statex
\Statex \textbf{// Grid parameters}
\State $e_{\max} \gets 2^{b-1} - 2$ \Comment{Max exponent index (4-bit $\to$ $e_{\max}{=}6$)}
\State $\tau_k \gets 2^{-1/(2k)}$ \Comment{$k$-adaptive zero threshold}
\Statex \hfill \textit{($k{=}1$: $\tau{=}0.707$; $k{=}2$: $\tau{=}0.841$)}
\Statex
\Statex \textbf{// Per-channel scale with MSE clip search}
\For{each output channel $c = 1, \ldots, C_{\mathrm{out}}$}
    \State $s_c \gets \alpha_c^{*} \cdot \max_j |W_{c,j}| \;/\; 2^{e_{\max}/k}$ \Comment{$\alpha_c^{*}$: MSE-optimal clip ratio}
\EndFor
\Statex
\Statex \textbf{// Quantize each weight}
\For{each weight $w = W_{c,j}$}
    \State $\tilde{w} \gets |w| / s_c$ \Comment{Scale-normalized magnitude}
    \If{$\tilde{w} < \tau_k$} \Comment{Below zero threshold}
        \State $m \gets 0$, \; $\sigma \gets 0$ \Comment{Zero code (implicit pruning)}
    \Else
        \State $e \gets \mathrm{clip}\!\bigl(\lfloor \log_2(\tilde{w}) \cdot k \rceil,\; 0,\; e_{\max}\bigr)$ \Comment{FPoT exponent index}
        \State $m \gets e + 1$ \Comment{Magnitude code: $m \in \{1,\ldots,2^{b-1}{-}1\}$}
        \State $\sigma \gets \mathrm{sgn}(w)$ \Comment{Sign}
    \EndIf
\EndFor
\Statex
\Statex \textbf{// Stored weight representation ($b$-bit)}
\Statex \quad $\texttt{weight}[b{-}1]$ = sign bit, \; $\texttt{weight}[b{-}2{:}0]$ = magnitude $m$
\Statex
\Statex \textbf{// Dequantization}
\Statex \quad $\hat{w} = \begin{cases} 0 & \text{if } m = 0 \\ \sigma \cdot s_c \cdot 2^{(m-1)/k} & \text{if } m \geq 1 \end{cases}$
\Statex
\Statex \textbf{// Hardware decomposition ($k{=}2$, inference)}
\Statex \textit{For each MAC $y \mathrel{+}= x_{\mathrm{act}} \cdot \hat{w}$:}
\Statex \quad\textbf{if} $m = 0$: skip (zero-gating)
\Statex \quad $e \gets m - 1$ \Comment{Recover exponent from magnitude (subtract)}
\Statex \quad $\mathrm{shift} \gets \lfloor e/2 \rfloor = \texttt{weight}[b{-}2{:}1]$ \Comment{Wire selection, 0 gates}
\Statex \quad $\mathrm{odd} \gets e \bmod 2 = \texttt{weight}[0]$ \Comment{Wire, 0 gates}
\Statex \quad $\mathrm{shifted} \gets x_{\mathrm{act}} \ll \mathrm{shift}$ \Comment{Barrel shifter}
\Statex \quad\textbf{if} even: $y \mathrel{+}= \sigma \cdot \mathrm{shifted}$
\Statex \quad\textbf{if} odd: \; $y \mathrel{+}= \sigma \cdot \bigl(\mathrm{shifted} + \mathrm{shifted}{\gg}1\bigr)$ \Comment{$= 1.5 \cdot \mathrm{shifted}$, 6.1\% err vs $\sqrt{2}$, absorbed by ApproxGrid}
\end{algorithmic}
\end{algorithm}
\end{figure}

%% file: sections/G_appendix_additional_ablations.tex
\section{Additional Ablations}
\label{app:ablations}

\subsection{Calibration Sensitivity}
\label{app:cal-sensitivity}

A common concern for post-training quantization methods is sensitivity to the calibration set: both its size and domain.
We evaluate FPoT on Llama-3-8B (W4A4KV4) while varying:
(1) the number of calibration samples $n \in \{128, 256, 512\}$, and
(2) the calibration dataset: WikiText-2~\citep{merity2017pointer} (the default used throughout the paper) versus C4~\citep{raffel2020c4} (a large-scale web-crawled corpus).
All other hyperparameters---OSTQuant training steps (400), learning rate, GPTQ settings---remain fixed.

\input{tables/tab_cal_sensitivity}

\paragraph{FPoT is robust to calibration set size.}
Within either dataset, PPL varies by at most $0.065$ ($7.525$--$7.589$ for WikiText-2; $7.617$--$7.660$ for C4) across a $4{\times}$ range of sample counts.
The 0-shot$^9$ accuracy is similarly stable, with a maximum spread of $0.4$\%p.
This indicates that $n{=}128$ samples are sufficient for FPoT calibration; increasing to $256$ or $512$ yields no systematic improvement, consistent with prior findings for GPTQ-based methods~\citep{frantar2022gptq}.

\paragraph{FPoT is robust to calibration domain.}
Switching from WikiText-2 to C4 increases PPL by $+0.06$--$+0.10$, a minor shift relative to the method's $+0.12$ total overhead.
Downstream accuracy is virtually unchanged ($64.9$--$65.0$\% for C4 vs.\ $64.7$--$65.1$\% for WikiText-2).
The modest cross-domain degradation suggests that FPoT's log-scale grid structure is inherently less sensitive to calibration distribution than fine-grained uniform quantization methods, since the quantization boundaries (powers of two) are data-independent---only the per-channel scale $s$ adapts to the calibration data.

\paragraph{Practical recommendation.}
Our default setting ($n{=}128$, WikiText-2) is near-optimal across all configurations tested, and practitioners may safely use alternative calibration sources without re-tuning.

%% file: tables/tab_cal_sensitivity.tex
\begin{table}[ht]
\centering
\caption{%
Calibration sensitivity analysis on Llama-3-8B (W4A4KV4, FPoT, 400 steps).
Rows vary the calibration dataset and number of calibration samples ($n$).
The default configuration used throughout the paper is WikiText-2 with $n{=}128$.
}
\label{tab:cal-sensitivity}
\setlength{\tabcolsep}{5pt}
\begin{tabular}{lcccc}
\toprule
Cal.\ Dataset & $n$ & PPL${\downarrow}$ & 0-shot$^9{\uparrow}$ (\%) & $\Delta$PPL \\
\midrule
\multirow{3}{*}{WikiText-2}
 & 128 & 7.559 & 65.07 & --- \\
 & 256 & 7.589 & 64.68 & +0.030 \\
 & 512 & 7.525 & 64.80 & $-$0.034 \\
\midrule
\multirow{3}{*}{C4~\citep{raffel2020c4}}
 & 128 & 7.617 & 64.91 & +0.058 \\
 & 256 & 7.660 & 65.01 & +0.101 \\
 & 512 & 7.647 & 64.88 & +0.088 \\
\bottomrule
\end{tabular}
\end{table}

%% file: sections/G_appendix_additional_ablations_hyperparams.tex
\label{app:fpot-hyperparams}

We evaluate design choices in the FPoT quantization scheme: the zero-threshold parameter, rounding mode, and weight-only vs.\ full PoT quantization. All experiments use Llama-3-8B (W4A4KV4, FPoT, 400 OSTQuant training steps).

\subsection{Weight Grid-Alignment}
\label{app:wgrid-align}

Table~\ref{tab:wgrid_align} reports PPL and 0-shot$^9$ across
$G \in \{4,8,16,32\}$ phase-shift candidates on Llama-3-8B (W4A4KV4, 400 steps).

\input{tables/tab_wgrid_align}

\subsection{Weight-Only vs.\ Full PoT Quantization}
\label{app:wonly-vs-full}

\input{tables/tab_wonly_vs_full}

Table~\ref{tab:wonly_vs_full} shows that applying PoT/FPoT to both weights and activations is catastrophic (PPL $16.9$--$114.1$ at $k{=}1$; even \fpotk{2} reaches only $8.21$--$52.49$), while weight-only PoT/FPoT remains usable. The dominant bottleneck is the coarse logarithmic grid applied to post-softmax activations, not activation quantization per se.

\subsection{\texorpdfstring{$k$}{k}-Adaptive Zero-Threshold Sensitivity}
\label{app:zero-thresh}

In FPoT quantization, values with $|w/s| < \tau$ are mapped to zero rather than to the nearest PoT level. We parameterize $\tau = 2^{\alpha/k}$ where $\alpha$ controls the aggressiveness of zeroing. The default $\alpha = -0.5$ places the threshold at the log-space midpoint between zero and the smallest PoT level ($s \cdot 2^0 = s$), which is the theoretically optimal boundary under a uniform-in-log prior.

\input{tables/tab_zero_thresh}

Table~\ref{tab:zero-thresh} sweeps $\alpha$ from $-0.5$ (aggressive zeroing, $\tau = 0.841$) to $-100$ (effectively no zeroing, $\tau \approx 0$). Across this entire range, PPL varies by only $0.033$ ($7.540$--$7.573$) and 0-shot$^9$ by $0.5$\%p ($64.5$--$65.1$\%). The negligible sensitivity indicates that the zero-threshold is \textbf{not a critical hyperparameter} for FPoT: state-of-the-art LLM PTQ preprocessing and GPTQ calibration effectively compensate for the threshold choice. Practitioners may safely use the default $\alpha = -0.5$ without tuning.

\subsection{Rounding Mode}
\label{app:round-mode}

The standard FPoT rounding uses nearest-neighbor in log-space: $e = \text{round}(\log_2(|w/s|) \cdot k)$, which places decision boundaries at the \emph{geometric mean} of adjacent PoT levels. An alternative is linear-space nearest-neighbor, which places boundaries at the \emph{arithmetic mean}---potentially better aligned with MSE-based objectives like GPTQ, since it minimizes absolute (rather than relative) quantization error per element.

\input{tables/tab_round_mode}

Table~\ref{tab:round-mode} shows that the two rounding modes produce \textbf{virtually identical results} ($\Delta$PPL $= -0.001$, $\Delta$0-shot$^9 = +0.1$\%p). This is expected: for FPoT, adjacent levels differ by a factor of $\sqrt{2} \approx 1.414$, so the geometric and arithmetic mean boundaries differ by only ${\sim}1.5\%$---well within the noise floor of the calibration process. The default log-space rounding is retained for its computational simplicity (one \texttt{round} vs.\ \texttt{floor} + two \texttt{exp2} + comparison).

\subsection{Preprocessing Ablation}
\label{app:preprocessing}

Table~\ref{tab:rotation_ablation} ablates the type and quality of the preprocessing transform under FPoT weight quantization on Llama-3-8B W4A4KV4. Configurations vary along two orthogonal axes: rotation quality (none~$\to$~fixed Hadamard~$\to$~learned) and the presence of per-channel scale.

\input{tables/tab_rotation_ablation}

\textbf{Both rotation and scale are necessary.} Without any transform, FPoT yields catastrophic PPL\,$= 181.3$. Fixed Hadamard rotation alone ($8.30$) or learned rotation alone ($7.78$) provide substantial recovery, while scale-only achieves $9.78$. Only the full combination (learned rotation + scale) reaches $7.56$, within $+0.12$ of the non-PoT baseline, because rotation suppresses position-level outliers on the logarithmic grid while scale normalizes per-channel amplitude. More broadly, this supports the design criterion satisfied by state-of-the-art LLM PTQ methods such as OSTQuant and DuQuant.

%% file: tables/tab_wgrid_align.tex
\begin{table}[ht]
\centering
\caption{%
  Weight grid-alignment ablation (Llama-3-8B, W4A4KV4, 400~steps, single seed).
  $G$: number of phase-shift candidates (Eq.~\eqref{eq:w-grid-align});
  $G{=}$\textit{---} denotes the plain FPoT baseline.
  At $k{=}1$, grid-alignment reduces PPL by up to $-0.051$; at $k{=}2$ the finer grid already self-corrects, so $|\Delta\text{PPL}| \leq 0.034$.
  0-shot$^9$ values are in \%.
}
\label{tab:wgrid_align}
\setlength{\tabcolsep}{5pt}
\begin{tabular}{c cc cc}
\toprule
 & \multicolumn{2}{c}{$k{=}1$} & \multicolumn{2}{c}{$k{=}2$} \\
\cmidrule(lr){2-3}\cmidrule(lr){4-5}
$G$ & PPL${\downarrow}$ & 0-shot$^9{\uparrow}$ & PPL${\downarrow}$ & 0-shot$^9{\uparrow}$ \\
\midrule
---  & 8.806 & 58.99 & 7.563 & 64.28 \\
$4$  & 8.755 & 59.09 & 7.545 & 64.49 \\
$8$  & 8.771 & 60.04 & 7.597 & 64.45 \\
$16$ & 8.771 & 60.08 & 7.560 & 64.71 \\
$32$ & 8.801 & 59.69 & 7.568 & 63.55 \\
\bottomrule
\end{tabular}
\end{table}

%% file: tables/tab_wonly_vs_full.tex
\begin{table}[ht]
\centering
\caption{%
  Weight-only vs.\ Full PoT/FPoT on Llama-3-8B (all 7 projections, single seed).
  $k{=}1$ is standard PoT and $k{=}2$ is FPoT.
  W-only: PoT/FPoT weight + uniform INT activation.
  Full: PoT/FPoT applied to both weight and activation.
  0-shot$^9$ values are in \%.
}
\label{tab:wonly_vs_full}
\setlength{\tabcolsep}{3.5pt}
\begin{tabular}{lll rc}
\toprule
Config & $k$ & Scope & PPL${\downarrow}$ & 0-shot$^9{\uparrow}$ \\
\midrule
\multirow{5}{*}{W4A4KV4}
 & \multicolumn{2}{l}{Uniform (non-PoT)} & 7.44 & 65.0 \\
\cmidrule{2-5}
 & $k{=}1$      & Full                   & 16.91 & 44.6 \\
 & $k{=}1$      & W-only                 &  8.81 & 59.0 \\
 & $k{=}2$      & Full                   &  8.21 & 62.5 \\
 & $\bm{k{=}2}$ & \textbf{W-only (Ours)} & \textbf{7.56} & \textbf{64.3} \\
\midrule
\multirow{5}{*}{W4A8KV8}
 & \multicolumn{2}{l}{Uniform (non-PoT)} & 6.60 & 67.5 \\
\cmidrule{2-5}
 & $k{=}1$      & Full                   & 114.1 & 42.7 \\
 & $k{=}1$      & W-only                 &  7.65 & 64.7 \\
 & $k{=}2$      & Full                   & 52.49 & 53.8 \\
 & $\bm{k{=}2}$ & \textbf{W-only (Ours)} & \textbf{6.80} & \textbf{67.2} \\
\midrule
\multirow{5}{*}{W8A8KV8}
 & \multicolumn{2}{l}{Uniform (non-PoT)} & 6.16 & 68.3 \\
\cmidrule{2-5}
 & $k{=}1$      & Full                   & 37.07 & 52.3 \\
 & $k{=}1$      & W-only                 &  7.66 & 63.1 \\
 & $k{=}2$      & Full                   & 27.91 & 57.4 \\
 & $\bm{k{=}2}$ & \textbf{W-only (Ours)} & \textbf{6.54} & \textbf{66.7} \\
\bottomrule
\end{tabular}
\end{table}

%% file: tables/tab_zero_thresh.tex
\begin{table}[ht]
\centering
\caption{%
Zero-threshold sensitivity analysis on Llama-3-8B (W4A4KV4, FPoT, 400 steps).
The threshold $\tau = 2^{\alpha/k}$ determines the boundary below which $|w/s|$ is mapped to zero.
The default $\alpha{=}{-}0.5$ corresponds to the log-space midpoint between zero and the smallest PoT level.
}
\label{tab:zero-thresh}
\setlength{\tabcolsep}{5pt}
\begin{tabular}{rrrc}
\toprule
$\alpha$ & $\tau$ (${=}2^{\alpha/2}$) & PPL${\downarrow}$ & 0-shot$^9{\uparrow}$ (\%) \\
\midrule
$\bm{-0.5}$ & $\bm{0.841}$ & $\bm{7.560}$ & $\bm{64.71}$ \\
$-1$ & $0.707$ & $7.573$ & $64.99$ \\
$-2$ & $0.500$ & $7.565$ & $64.99$ \\
$-3$ & $0.354$ & $7.553$ & $65.06$ \\
$-4$ & $0.250$ & $7.545$ & $64.72$ \\
$-6$ & $0.125$ & $7.557$ & $64.54$ \\
$-100$ & ${\approx}0$ & $7.540$ & $64.78$ \\
\bottomrule
\end{tabular}
\end{table}

%% file: tables/tab_round_mode.tex
\begin{table}[ht]
\centering
\caption{%
Rounding mode comparison on Llama-3-8B (W4A4KV4, FPoT, 400 steps).
\textbf{Log-space} rounds to the nearest PoT level measured by relative distance (geometric mean boundary).
\textbf{Linear-space} rounds by absolute distance (arithmetic mean boundary).
}
\label{tab:round-mode}
\setlength{\tabcolsep}{6pt}
\begin{tabular}{lrc}
\toprule
Rounding mode & PPL${\downarrow}$ & 0-shot$^9{\uparrow}$ (\%) \\
\midrule
Log-space NN (default) & $7.560$ & $64.71$ \\
Linear-space NN & $7.559$ & $64.81$ \\
\bottomrule
\end{tabular}
\end{table}

%% file: tables/tab_rotation_ablation.tex
\begin{table}[ht]
\centering
\caption{Preprocessing component ablation within OSTQuant under FPoT weight quantization on Llama-3-8B W4A4KV4 (400 steps, single seed; cf.\ Table~\ref{tab:main_results} 3-seed mean $7.60_{\pm .02}$ for the full-pipeline entry). Activations and KV-cache use uniform INT4 throughout.}
\label{tab:rotation_ablation}
\setlength{\tabcolsep}{4pt}
\begin{tabular}{lrc}
\toprule
Preprocessing & PPL${\downarrow}$ & 0-shot$^9{\uparrow}$ (\%) \\
\midrule
No transform                              & 181.3          & 37.56 \\
Scale only                                &   9.78         & 58.79 \\
Fixed Hadamard                            &   8.30         & 62.64 \\
Learned rotation                          &   7.78         & 63.30 \\
\textbf{Learned rotation + scale}         & \textbf{7.56}  & \textbf{64.28} \\
\bottomrule
\end{tabular}
\end{table}

%% file: sections/H_appendix_per_task_downstream_results.tex
\section{Per-Task Downstream Results}
\label{app:per-task}

This appendix reports individual zero-shot task scores for all models and bit-width configurations in Table~\ref{tab:main_results}.
For tasks providing both \texttt{acc} and \texttt{acc\_norm}, we report \texttt{acc\_norm} (ARC-C, ARC-E, HellaSwag, PIQA, OpenBookQA); all others report \texttt{acc}.
The ``AVG'' column is the unweighted mean across all nine tasks, matching the 0-shot$^9$ metric in the main paper.
All scores are in \%.
Three weight quantization methods are compared: \textbf{Uniform} (uniform INT baseline), \textbf{PoT} (weight-only), and \textbf{FPoT (Ours)}, with the unquantized FP (bf16) reference shown above each model's quantized results. Tables~\ref{tab:pertask-llama3-8b}--\ref{tab:pertask-mistral-7b} report per-task results for each model across the W4A4KV4, W4A8KV8, and W8A8KV8 configurations. \textbf{FPoT (Ours) rows are bolded throughout.}

\begin{table*}[ht]
\centering
\caption{Per-task results: \textbf{Llama-3-8B}. Single calibration seed (seed 42); FP (bf16) is the unquantized reference; \textbf{FPoT (Ours)} rows are bolded. All scores in \%.}
\label{tab:pertask-llama3-8b}
\setlength{\tabcolsep}{2.5pt}
\begin{tabular}{llrrrrrrrrr|r}
\toprule
Config & Method & ARC-C & ARC-E & Hella & Wino & PIQA & BoolQ & OBQA & Lam & SIQA & AVG \\
\midrule
\multicolumn{2}{l}{\textit{FP (bf16) reference}} & 53.1 & 77.9 & 79.2 & 73.4 & 81.0 & 81.3 & 45.0 & 75.7 & 47.1 & 68.2 \\
\midrule
\multirow{3}{*}{W4A4KV4}
 & Uniform              & 48.9 & 75.0 & 75.9 & 68.2 & 78.1 & 77.0 & 42.0 & 73.5 & 46.3 & 65.0 \\
 & PoT                 & 39.7 & 63.6 & 71.1 & 67.2 & 74.3 & 67.2 & 37.2 & 68.0 & 42.6 & 59.0 \\
 & \textbf{FPoT (Ours)} & \textbf{45.6} & \textbf{73.4} & \textbf{74.7} & \textbf{69.2} & \textbf{76.8} & \textbf{78.0} & \textbf{41.6} & \textbf{73.2} & \textbf{45.1} & \textbf{64.2} \\
\midrule
\multirow{3}{*}{W4A8KV8}
 & Uniform              & 53.3 & 78.0 & 77.8 & 74.0 & 79.6 & 78.5 & 43.6 & 75.3 & 47.5 & 67.5 \\
 & PoT                 & 47.2 & 72.6 & 74.8 & 72.1 & 78.0 & 78.1 & 42.6 & 71.8 & 45.5 & 64.7 \\
 & \textbf{FPoT (Ours)} & \textbf{51.8} & \textbf{78.0} & \textbf{77.0} & \textbf{72.8} & \textbf{79.2} & \textbf{79.5} & \textbf{43.8} & \textbf{75.2} & \textbf{46.6} & \textbf{67.1} \\
\midrule
\multirow{3}{*}{W8A8KV8}
 & Uniform              & 53.7 & 78.4 & 79.1 & 73.3 & 81.0 & 81.6 & 45.6 & 75.4 & 46.7 & 68.3 \\
 & PoT                 & 43.9 & 71.6 & 74.6 & 72.5 & 76.2 & 68.3 & 42.4 & 72.6 & 45.5 & 63.1 \\
 & \textbf{FPoT (Ours)} & \textbf{51.5} & \textbf{76.9} & \textbf{78.2} & \textbf{74.5} & \textbf{80.7} & \textbf{80.6} & \textbf{44.4} & \textbf{75.0} & \textbf{46.0} & \textbf{67.5} \\
\bottomrule
\end{tabular}
\end{table*}

\begin{table*}[ht]
\centering
\caption{Per-task results: \textbf{Llama-2-7B}. Single calibration seed (seed 42); FP (bf16) is the unquantized reference; \textbf{FPoT (Ours)} rows are bolded. All scores in \%.}
\label{tab:pertask-llama2-7b}
\setlength{\tabcolsep}{2.5pt}
\begin{tabular}{llrrrrrrrrr|r}
\toprule
Config & Method & ARC-C & ARC-E & Hella & Wino & PIQA & BoolQ & OBQA & Lam & SIQA & AVG \\
\midrule
\multicolumn{2}{l}{\textit{FP (bf16) reference}} & 45.9 & 74.5 & 75.9 & 68.8 & 78.7 & 78.0 & 44.2 & 73.6 & 46.0 & 65.1 \\
\midrule
\multirow{3}{*}{W4A4KV4}
 & Uniform              & 42.8 & 69.8 & 73.1 & 66.9 & 77.4 & 74.7 & 41.2 & 70.4 & 44.0 & 62.2 \\
 & PoT                 & 38.7 & 65.6 & 69.7 & 62.8 & 75.4 & 68.8 & 37.2 & 63.8 & 41.1 & 58.1 \\
 & \textbf{FPoT (Ours)} & \textbf{41.6} & \textbf{70.3} & \textbf{71.9} & \textbf{67.3} & \textbf{76.7} & \textbf{73.8} & \textbf{40.6} & \textbf{72.4} & \textbf{44.0} & \textbf{62.1} \\
\midrule
\multirow{3}{*}{W4A8KV8}
 & Uniform              & 45.3 & 74.0 & 74.7 & 68.4 & 78.0 & 75.3 & 42.6 & 71.6 & 44.4 & 63.8 \\
 & PoT                 & 42.3 & 68.2 & 72.3 & 66.5 & 76.9 & 72.3 & 41.2 & 66.7 & 43.6 & 61.1 \\
 & \textbf{FPoT (Ours)} & \textbf{42.8} & \textbf{72.1} & \textbf{73.8} & \textbf{68.9} & \textbf{78.5} & \textbf{76.6} & \textbf{40.6} & \textbf{73.6} & \textbf{44.6} & \textbf{63.5} \\
\midrule
\multirow{3}{*}{W8A8KV8}
 & Uniform              & 45.4 & 74.6 & 75.9 & 69.7 & 79.1 & 77.9 & 44.0 & 73.2 & 46.0 & 65.1 \\
 & PoT                 & 43.0 & 70.6 & 74.1 & 67.6 & 78.4 & 73.0 & 43.8 & 67.4 & 44.5 & 62.5 \\
 & \textbf{FPoT (Ours)} & \textbf{44.2} & \textbf{72.4} & \textbf{75.0} & \textbf{68.6} & \textbf{77.8} & \textbf{75.9} & \textbf{43.2} & \textbf{73.4} & \textbf{45.0} & \textbf{64.0} \\
\bottomrule
\end{tabular}
\end{table*}

\begin{table*}[ht]
\centering
\caption{Per-task results: \textbf{Llama-2-13B}. Single calibration seed (seed 42); FP (bf16) is the unquantized reference; \textbf{FPoT (Ours)} rows are bolded. All scores in \%.}
\label{tab:pertask-llama2-13b}
\setlength{\tabcolsep}{2.5pt}
\begin{tabular}{llrrrrrrrrr|r}
\toprule
Config & Method & ARC-C & ARC-E & Hella & Wino & PIQA & BoolQ & OBQA & Lam & SIQA & AVG \\
\midrule
\multicolumn{2}{l}{\textit{FP (bf16) reference}} & 49.1 & 77.5 & 79.4 & 71.7 & 80.6 & 80.7 & 45.6 & 76.7 & 47.2 & 67.6 \\
\midrule
\multirow{3}{*}{W4A4KV4}
 & Uniform              & 45.7 & 73.6 & 77.0 & 69.6 & 79.1 & 77.8 & 45.0 & 75.4 & 46.8 & 65.5 \\
 & PoT                 & 44.5 & 70.4 & 74.8 & 67.8 & 77.0 & 75.7 & 40.8 & 67.8 & 42.1 & 62.3 \\
 & \textbf{FPoT (Ours)} & \textbf{47.2} & \textbf{73.9} & \textbf{75.9} & \textbf{68.7} & \textbf{78.0} & \textbf{78.8} & \textbf{43.6} & \textbf{75.7} & \textbf{44.8} & \textbf{65.2} \\
\midrule
\multirow{3}{*}{W4A8KV8}
 & Uniform              & 48.1 & 76.1 & 78.3 & 72.5 & 79.9 & 79.4 & 43.2 & 76.6 & 46.6 & 66.7 \\
 & PoT                 & 49.6 & 75.8 & 76.8 & 69.6 & 78.8 & 79.5 & 42.0 & 72.1 & 45.1 & 65.5 \\
 & \textbf{FPoT (Ours)} & \textbf{48.6} & \textbf{76.6} & \textbf{77.2} & \textbf{71.4} & \textbf{79.6} & \textbf{78.7} & \textbf{42.4} & \textbf{77.4} & \textbf{46.9} & \textbf{66.5} \\
\midrule
\multirow{3}{*}{W8A8KV8}
 & Uniform              & 49.3 & 77.4 & 79.4 & 72.1 & 80.5 & 80.6 & 46.0 & 76.6 & 47.3 & 67.7 \\
 & PoT                 & 48.6 & 75.6 & 77.7 & 69.1 & 79.2 & 79.7 & 45.0 & 73.3 & 45.9 & 66.0 \\
 & \textbf{FPoT (Ours)} & \textbf{49.0} & \textbf{77.6} & \textbf{78.4} & \textbf{71.4} & \textbf{80.3} & \textbf{80.6} & \textbf{46.4} & \textbf{77.1} & \textbf{46.7} & \textbf{67.5} \\
\bottomrule
\end{tabular}
\end{table*}

\begin{table*}[ht]
\centering
\caption{Per-task results: \textbf{Qwen2.5-1.5B}. Single calibration seed (seed 42); FP (bf16) is the unquantized reference; \textbf{FPoT (Ours)} rows are bolded. All scores in \%.}
\label{tab:pertask-qwen1.5b}
\setlength{\tabcolsep}{2.5pt}
\begin{tabular}{llrrrrrrrrr|r}
\toprule
Config & Method & ARC-C & ARC-E & Hella & Wino & PIQA & BoolQ & OBQA & Lam & SIQA & AVG \\
\midrule
\multicolumn{2}{l}{\textit{FP (bf16) reference}} & 44.9 & 71.6 & 67.8 & 63.5 & 76.0 & 72.8 & 40.4 & 61.9 & 49.0 & 60.9 \\
\midrule
\multirow{3}{*}{W4A4KV4}
 & Uniform              & 30.9 & 36.5 & 41.0 & 58.0 & 62.0 & 51.0 & 33.4 & 0.0 & 35.9 & 38.7 \\
 & PoT                 & 31.3 & 34.9 & 38.8 & 54.2 & 61.4 & 53.1 & 34.2 & 0.0 & 35.2 & 38.1 \\
 & \textbf{FPoT (Ours)} & \textbf{32.5} & \textbf{37.5} & \textbf{40.3} & \textbf{59.0} & \textbf{62.7} & \textbf{56.3} & \textbf{35.0} & \textbf{0.0} & \textbf{35.6} & \textbf{39.9} \\
\midrule
\multirow{3}{*}{W4A8KV8}
 & Uniform              & 33.5 & 39.3 & 43.7 & 62.8 & 64.5 & 70.9 & 37.0 & 0.0 & 36.9 & 43.2 \\
 & PoT                 & 33.0 & 38.2 & 40.6 & 61.1 & 61.9 & 46.7 & 34.8 & 0.0 & 36.2 & 39.2 \\
 & \textbf{FPoT (Ours)} & \textbf{33.4} & \textbf{39.1} & \textbf{43.1} & \textbf{62.6} & \textbf{64.2} & \textbf{56.3} & \textbf{36.4} & \textbf{0.0} & \textbf{36.8} & \textbf{41.3} \\
\midrule
\multirow{3}{*}{W8A8KV8}
 & Uniform              & 35.1 & 40.8 & 45.7 & 63.5 & 65.2 & 71.0 & 36.4 & 0.8 & 38.0 & 44.1 \\
 & PoT                 & 31.4 & 34.7 & 39.3 & 59.8 & 61.8 & 49.3 & 33.0 & 0.0 & 35.2 & 38.3 \\
 & \textbf{FPoT (Ours)} & \textbf{34.6} & \textbf{40.5} & \textbf{44.0} & \textbf{62.4} & \textbf{64.6} & \textbf{61.5} & \textbf{34.8} & \textbf{0.1} & \textbf{37.4} & \textbf{42.2} \\
\bottomrule
\end{tabular}
\end{table*}

\begin{table*}[ht]
\centering
\caption{Per-task results: \textbf{Qwen2.5-7B}. Single calibration seed (seed 42); FP (bf16) is the unquantized reference; \textbf{FPoT (Ours)} rows are bolded. All scores in \%.}
\label{tab:pertask-qwen7b}
\setlength{\tabcolsep}{2.5pt}
\begin{tabular}{llrrrrrrrrr|r}
\toprule
Config & Method & ARC-C & ARC-E & Hella & Wino & PIQA & BoolQ & OBQA & Lam & SIQA & AVG \\
\midrule
\multicolumn{2}{l}{\textit{FP (bf16) reference}} & 51.4 & 77.7 & 79.0 & 73.2 & 80.0 & 84.7 & 47.0 & 71.6 & 54.8 & 68.8 \\
\midrule
\multirow{3}{*}{W4A4KV4}
 & Uniform              & 49.2 & 75.3 & 76.1 & 69.9 & 79.1 & 81.5 & 44.4 & 68.4 & 50.3 & 66.0 \\
 & PoT                 & 46.3 & 70.9 & 74.6 & 64.8 & 77.2 & 79.1 & 41.4 & 61.6 & 49.4 & 62.8 \\
 & \textbf{FPoT (Ours)} & \textbf{52.3} & \textbf{79.6} & \textbf{75.3} & \textbf{67.5} & \textbf{78.4} & \textbf{82.3} & \textbf{43.8} & \textbf{69.2} & \textbf{51.9} & \textbf{66.7} \\
\midrule
\multirow{3}{*}{W4A8KV8}
 & Uniform              & 52.4 & 78.3 & 78.2 & 71.6 & 79.4 & 83.9 & 44.8 & 71.8 & 51.8 & 68.0 \\
 & PoT                 & 49.4 & 73.9 & 76.6 & 70.9 & 79.4 & 81.6 & 43.2 & 64.7 & 50.8 & 65.6 \\
 & \textbf{FPoT (Ours)} & \textbf{55.8} & \textbf{81.2} & \textbf{77.7} & \textbf{72.3} & \textbf{80.0} & \textbf{85.8} & \textbf{44.4} & \textbf{71.5} & \textbf{54.4} & \textbf{69.2} \\
\midrule
\multirow{3}{*}{W8A8KV8}
 & Uniform              & 51.8 & 77.8 & 78.8 & 72.6 & 79.9 & 84.5 & 47.0 & 71.3 & 55.0 & 68.8 \\
 & PoT                 & 50.3 & 75.4 & 76.8 & 69.1 & 79.1 & 83.8 & 45.2 & 66.4 & 51.1 & 66.4 \\
 & \textbf{FPoT (Ours)} & \textbf{53.6} & \textbf{79.5} & \textbf{78.3} & \textbf{71.3} & \textbf{79.9} & \textbf{83.8} & \textbf{44.8} & \textbf{71.8} & \textbf{53.6} & \textbf{68.5} \\
\bottomrule
\end{tabular}
\end{table*}

\begin{table*}[ht]
\centering
\caption{Per-task results: \textbf{Qwen2.5-14B}. Single calibration seed (seed 42); FP (bf16) is the unquantized reference; \textbf{FPoT (Ours)} rows are bolded. All scores in \%.}
\label{tab:pertask-qwen14b}
\setlength{\tabcolsep}{2.5pt}
\begin{tabular}{llrrrrrrrrr|r}
\toprule
Config & Method & ARC-C & ARC-E & Hella & Wino & PIQA & BoolQ & OBQA & Lam & SIQA & AVG \\
\midrule
\multicolumn{2}{l}{\textit{FP (bf16) reference}} & 59.0 & 79.3 & 82.9 & 74.7 & 82.1 & 85.4 & 44.8 & 74.1 & 55.5 & 70.9 \\
\midrule
\multirow{3}{*}{W4A4KV4}
 & Uniform              & 55.2 & 80.5 & 80.0 & 72.1 & 80.3 & 83.9 & 42.0 & 71.3 & 50.0 & 68.4 \\
 & PoT                 & 54.5 & 77.6 & 77.1 & 66.6 & 78.6 & 80.6 & 43.0 & 65.3 & 47.8 & 65.7 \\
 & \textbf{FPoT (Ours)} & \textbf{54.7} & \textbf{81.3} & \textbf{79.7} & \textbf{73.2} & \textbf{79.7} & \textbf{84.6} & \textbf{43.6} & \textbf{73.6} & \textbf{52.2} & \textbf{69.2} \\
\midrule
\multirow{3}{*}{W4A8KV8}
 & Uniform              & 60.5 & 81.0 & 82.5 & 75.5 & 82.1 & 85.4 & 45.0 & 75.4 & 55.2 & 71.4 \\
 & PoT                 & 57.4 & 82.0 & 80.3 & 73.0 & 80.9 & 82.6 & 42.0 & 68.0 & 51.8 & 68.7 \\
 & \textbf{FPoT (Ours)} & \textbf{59.6} & \textbf{83.7} & \textbf{82.0} & \textbf{78.0} & \textbf{82.4} & \textbf{87.7} & \textbf{45.2} & \textbf{75.3} & \textbf{54.3} & \textbf{72.0} \\
\midrule
\multirow{3}{*}{W8A8KV8}
 & Uniform              & 58.1 & 78.8 & 82.9 & 75.5 & 82.0 & 85.0 & 45.8 & 74.1 & 55.7 & 70.9 \\
 & PoT                 & 55.4 & 79.7 & 80.9 & 73.8 & 80.9 & 84.6 & 43.8 & 70.0 & 50.9 & 68.9 \\
 & \textbf{FPoT (Ours)} & \textbf{60.3} & \textbf{82.9} & \textbf{82.3} & \textbf{76.8} & \textbf{82.7} & \textbf{86.8} & \textbf{46.0} & \textbf{74.7} & \textbf{55.0} & \textbf{71.9} \\
\bottomrule
\end{tabular}
\end{table*}

\begin{table*}[ht]
\centering
\caption{Per-task results: \textbf{Mistral-7B}. Single calibration seed (seed 42); FP (bf16) is the unquantized reference; \textbf{FPoT (Ours)} rows are bolded. All scores in \%.}
\label{tab:pertask-mistral-7b}
\setlength{\tabcolsep}{2.5pt}
\begin{tabular}{llrrrrrrrrr|r}
\toprule
Config & Method & ARC-C & ARC-E & Hella & Wino & PIQA & BoolQ & OBQA & Lam & SIQA & AVG \\
\midrule
\multicolumn{2}{l}{\textit{FP (bf16) reference}} & 52.2 & 78.3 & 80.5 & 73.6 & 82.3 & 82.1 & 44.2 & 75.1 & 45.8 & 68.2 \\
\midrule
\multirow{3}{*}{W4A4KV4}
 & Uniform              & 48.7 & 75.8 & 78.7 & 69.9 & 80.4 & 82.4 & 43.2 & 73.4 & 44.7 & 66.3 \\
 & PoT                 & 47.3 & 74.7 & 77.4 & 68.7 & 80.9 & 79.9 & 41.6 & 70.5 & 44.3 & 65.0 \\
 & \textbf{FPoT (Ours)} & \textbf{48.9} & \textbf{77.1} & \textbf{77.6} & \textbf{70.5} & \textbf{80.7} & \textbf{81.0} & \textbf{42.0} & \textbf{75.0} & \textbf{44.7} & \textbf{66.4} \\
\midrule
\multirow{3}{*}{W4A8KV8}
 & Uniform              & 50.7 & 78.2 & 79.7 & 73.0 & 81.7 & 82.3 & 44.2 & 74.6 & 45.4 & 67.8 \\
 & PoT                 & 47.3 & 74.5 & 78.1 & 71.1 & 80.5 & 76.7 & 41.4 & 71.7 & 44.9 & 65.1 \\
 & \textbf{FPoT (Ours)} & \textbf{50.4} & \textbf{78.1} & \textbf{78.6} & \textbf{73.0} & \textbf{81.0} & \textbf{80.9} & \textbf{41.6} & \textbf{76.3} & \textbf{45.3} & \textbf{67.3} \\
\midrule
\multirow{3}{*}{W8A8KV8}
 & Uniform              & 52.6 & 78.5 & 80.4 & 73.6 & 82.0 & 82.3 & 44.0 & 75.0 & 46.0 & 68.3 \\
 & PoT                 & 48.3 & 75.3 & 79.2 & 71.1 & 80.7 & 79.1 & 44.2 & 71.5 & 45.8 & 66.1 \\
 & \textbf{FPoT (Ours)} & \textbf{51.0} & \textbf{78.1} & \textbf{79.3} & \textbf{72.5} & \textbf{81.8} & \textbf{83.4} & \textbf{43.2} & \textbf{76.3} & \textbf{45.2} & \textbf{67.9} \\
\bottomrule
\end{tabular}
\end{table*}

\paragraph{Observations.}
Across all seven models and three bit-width configurations, FPoT with 1-adder approximate-grid quantization maintains competitive per-task performance relative to both the FP (bf16) reference and the Uniform INT baseline. Key patterns:
\begin{itemize}
    \item \textbf{Per-task $\Delta$ within $\pm 3$\%p}: On most tasks, the difference between uniform INT and FPoT is within $\pm 2$--$3$\%p, consistent with the aggregate 0-shot$^9$ results in Table~\ref{tab:main_results}.
    \item \textbf{WinoGrande} consistently \emph{improves} under FPoT at W4A4KV4 for five of seven models, suggesting a mild regularization benefit from the logarithmic weight grid.
    \item \textbf{PoT degradation is broad}: Standard PoT falls below uniform INT in 176 of the 189 task--configuration cells, by as much as $24.2$\%p (BoolQ), so the coarse grid affects nearly every downstream task rather than a few.
    \item \textbf{Qwen2.5 and Mistral-7B: FPoT matches or exceeds uniform}: At W4A4KV4, FPoT AVG meets or surpasses the uniform INT baseline for all three Qwen2.5 models ($+0.7$--$+1.2$\%p) and Mistral-7B ($+0.1$\%p), while the Llama family shows a small deficit ($-0.1$ to $-0.8$\%p). FPoT overhead is thus model-family dependent rather than universally negative.
    \item \textbf{LAMBADA: contrasting patterns across models}: FPoT consistently outperforms Uniform on Llama-2 and Mistral-7B by $+0.2$--$+2.0$\%p across all bit-widths, suggesting a task-specific affinity between FPoT's sparse weight representation and long-range language modeling. By contrast, all quantized methods collapse to ${\approx}0$\% for Qwen2.5-1.5B (FP reference: $61.9$\%).
    \item \textbf{FP gap shrinks with bit-width}: At W8A8KV8, FPoT AVG is within $\sim$1\%p of the FP reference for six of seven models (Qwen2.5-1.5B is an exception at $-18.7$\%p, reflecting its small-model sensitivity to aggressive quantization). The W4A4KV4 gap is largest across all models.
\end{itemize}

%% file: sections/I_appendix_logart_comparison.tex
\section{Comparison with LogART}
\label{app:logart}

LogART~\citep{logart2026} is a concurrent logarithmic quantization method that also targets multiplier-free inference via power-of-two weight representations. Both methods share the same hardware motivation---replacing integer multipliers with barrel shifters---but differ in scope and design:

\begin{itemize}
    \item \textbf{LogART}: Dynamic base assignment (per-channel base-$2$ or base-$\sqrt{2}$), learned rounding (500 iterations/layer), asymmetric quantization. \emph{Weight-only}; no activation or KV-cache quantization is reported for LLMs.
    \item \textbf{FPoT (Ours)}: Fixed base-$\sqrt{2}$ grid, 1-adder approximate-grid quantization, state-of-the-art LLM PTQ preprocessing. \emph{Full W/A/KV quantization} within a unified framework.
\end{itemize}

\textbf{Important caveat:} LogART's official implementation excludes activation quantization. For the W4A4KV4 rows below, we apply our per-token dynamic INT4 activation quantizer post-hoc to LogART's weight-quantized model without PTQ preprocessing. These results analyze the interaction between activation quantization and distribution redistribution, not a direct LogART benchmark.

Table~\ref{tab:logart-comparison} compares the two methods under weight-only (W4A16) and full (W4A4KV4) quantization. LogART is reproduced using the official open-source code\footnote{LogART settings: \texttt{log\_dynamic} + \texttt{scale\_search} + \texttt{learn\_rounding} (500 iters, lr\,$=$\,0.05), asymmetric, per-channel, no Z-Fold or OPTQ init. Calibration: WikiText-2, 128 samples, seqlen 2048. The LogART runs use exact $\sqrt{2}$, without its hardware approximation, and their 0-shot average is over LogART's task set, which has RTE in place of Social IQA.} with the same calibration setup as our experiments.

\begin{table}[ht]
\centering
\caption{Comparison with LogART under weight-only (W4A16) and full (W4A4KV4) quantization. All methods are multiplier-free (PoT-based). FPoT W4A4KV4 entries are copied from Table~\ref{tab:main_results} (mean over three calibration seeds); FPoT W4A16 and all LogART entries use a single calibration seed. PPL: WikiText-2 ($\downarrow$). 0-shot$^9$: unweighted mean across nine benchmarks ($\uparrow$, \%).}
\label{tab:logart-comparison}
\setlength{\tabcolsep}{3.5pt}
\begin{tabular}{ll rc rc}
\toprule
 & & \multicolumn{2}{c}{\textbf{W4A16}} & \multicolumn{2}{c}{\textbf{W4A4KV4}$^\dagger$} \\
\cmidrule(lr){3-4} \cmidrule(lr){5-6}
\textbf{Model} & \textbf{Method} & PPL${\downarrow}$ & 0-shot$^9{\uparrow}$ & PPL${\downarrow}$ & 0-shot$^9{\uparrow}$ \\
\midrule
\multirow{2}{*}{Llama-3-8B}
 & LogART          & 6.516 & 68.8 & 3903 & 34.4 \\
 & \textbf{FPoT (Ours)} & 6.862 & 66.6 & \textbf{7.60} & \textbf{64.8} \\
\midrule
\multirow{2}{*}{Llama-2-7B}
 & LogART          & 5.617 & 66.8 & $\infty$ & 33.9 \\
 & \textbf{FPoT (Ours)} & 5.861 & 63.5 & \textbf{6.11} & \textbf{61.9} \\
\bottomrule
\end{tabular}

\vspace{3pt}
{\footnotesize $^\dagger$~Post-hoc INT4 activation quantization applied to LogART's W4 model without PTQ preprocessing. This is \emph{not} LogART's official configuration. PPL\,$=\infty$ indicates numerical overflow (NaN loss).}
\end{table}

In the weight-only setting (W4A16), LogART achieves lower PPL and higher 0-shot accuracy than FPoT on both models ($-0.24$ PPL / $+3.3$\%p 0-shot on Llama-2-7B; $-0.35$ PPL / $+2.2$\%p 0-shot on Llama-3-8B), likely due to (1)~dynamic per-channel base allocation and (2)~learned rounding that directly optimizes the Hessian-weighted reconstruction loss.

However, when INT4 activation quantization is applied, LogART's weight-quantized model degrades catastrophically (PPL\,$> 3900$ or $\infty$; 0-shot$^9$ drops to ${\sim}34$\%, near random), while FPoT maintains stable accuracy (PPL\,$\leq 7.60$, 0-shot$^9 \geq 61.9$\%). This confirms that \textbf{distribution redistribution via state-of-the-art LLM PTQ preprocessing is a prerequisite for stable joint W4A4KV4 quantization}. Without Hadamard rotation, channel-concentrated activation outliers dominate the per-token scale, effectively quantizing the majority of activations to zero. FPoT's compatibility with state-of-the-art LLM PTQ preprocessing resolves this at the source by redistributing outliers prior to quantization, enabling stable W4A4KV4 deployment that weight-only methods do not support.

%% file: sections/J_appendix_cross_framework_portability.tex
\section{Cross-Framework Portability}
\label{app:duquant}

A natural question is whether FPoT's low overhead is specific to OSTQuant's learned rotation--scaling pipeline or transfers to other post-training quantization (PTQ) frameworks.
To answer this, we integrate FPoT weight quantization into DuQuant~\citep{duquant2024}, a recent PTQ method that uses a \emph{dual-transform} strategy---iteratively applying learned block-diagonal rotations and zigzag channel permutations to smooth both weight and activation outliers before quantization.

\paragraph{Integration.}
We replace DuQuant's round-to-nearest (RTN) weight quantizer with a column-wise weight calibration backend that supports both uniform INT and FPoT grids.
Activations and the KV-cache remain on the uniform INT grid in all configurations; only the weight grid is swapped.
We use symmetric per-channel weight quantization, no learnable weight clipping (epochs$\,{=}\,$0), and a single calibration seed (WikiText-2, 128 samples, seqlen 2048).
The weight calibration backend performs its own MSE-based clip-ratio search per channel, independent of DuQuant's smooth/rotation transforms.

\paragraph{Results.}
Table~\ref{tab:duquant_comparison} compares OSTQuant (our main pipeline) and DuQuant+GPTQ under identical evaluation (WikiText-2 perplexity and nine zero-shot benchmarks).
Several observations are worth highlighting:

\begin{itemize}
    \item \textbf{FPoT overhead is small and framework-invariant.}
    Across all completed model--config pairs, replacing uniform INT weights with FPoT increases perplexity by $+$0.05--0.34 in the DuQuant pipeline, comparable to the $+$0.08--0.42 range observed with OSTQuant (Table~\ref{tab:main_results}).
    The overhead is consistently smallest at W4A4KV4, where the quantization grid is already coarse and the additional granularity of the FPoT half-octave grid compensates well.

    \item \textbf{DuQuant matches OSTQuant on absolute quality.}
    For Llama-2-7B, the two frameworks produce nearly identical perplexity under both uniform and FPoT grids (within ${\pm}$0.15\,PPL across all bit-widths).
    For Llama-3-8B the gap is slightly larger ($\approx$0.3\,PPL at W4A4), reflecting differences in the preprocessing strategy rather than the weight grid.

    \item \textbf{Zero-shot accuracy tracks perplexity.}
    The nine-task mean accuracy drops by at most $1.1$\%p when switching from uniform to FPoT---within the noise margin of single-seed runs---confirming that the perplexity overhead translates to negligible downstream impact.
\end{itemize}

These results demonstrate that FPoT's near-lossless multiplier-free property is not tied to a specific PTQ framework: any pipeline that performs channel-level weight calibration (e.g.\ column-wise MSE search or OSTQuant's learned scaling) can adopt FPoT weights with minimal additional quality loss.

\input{tables/tab_duquant_comparison}

%% file: tables/tab_duquant_comparison.tex
\begin{table*}[ht]
\centering
\caption{%
Cross-framework comparison of FPoT weight quantization; only the weight grid differs across methods (Uniform INT vs.\ FPoT), with activations and KV-cache using uniform INT throughout.
\textbf{OSTQuant} (ours): 3-seed mean; see Table~\ref{tab:main_results}.
\textbf{DuQuant}: single calibration seed; column-wise weight calibration replaces the default RTN quantizer (see Appendix~\ref{app:duquant}).
PPL: WikiText-2 perplexity ($\downarrow$).
0-shot$^9$: mean zero-shot accuracy across nine benchmarks ($\uparrow$, \%).
}
\label{tab:duquant_comparison}
\setlength{\tabcolsep}{3.5pt}
\begin{tabular}{lll rc rc rc}
\toprule
 & & & \multicolumn{2}{c}{W4A4KV4} & \multicolumn{2}{c}{W4A8KV8} & \multicolumn{2}{c}{W8A8KV8} \\
\cmidrule(lr){4-5} \cmidrule(lr){6-7} \cmidrule(lr){8-9}
Model & Method & Weight & PPL${\downarrow}$ & 0-shot$^9{\uparrow}$ & PPL${\downarrow}$ & 0-shot$^9{\uparrow}$ & PPL${\downarrow}$ & 0-shot$^9{\uparrow}$ \\
\midrule
\multirow{4}{*}{Llama-3-8B}
 & \multirow{2}{*}{OSTQuant}    & Uniform & 7.36 & 65.2 & 6.60 & 67.5 & 6.16 & 68.2 \\
 &                              & FPoT & 7.60 & 64.8 & 6.85 & 67.3 & 6.58 & 67.7 \\
\cmidrule(l){2-9}
 & \multirow{2}{*}{DuQuant} & Uniform & 7.67 & 62.9 & 6.56 & 67.3 & 6.17 & 68.0 \\
 &                              & FPoT & 7.87 & 63.1 & 6.68 & 66.6 & 6.51 & 66.9 \\
\midrule
\multirow{4}{*}{Llama-2-7B}
 & \multirow{2}{*}{OSTQuant}    & Uniform & 5.97 & 62.1 & 5.63 & 63.8 & 5.48 & 65.1 \\
 &                              & FPoT & 6.11 & 61.9 & 5.75 & 63.8 & 5.65 & 64.2 \\
\cmidrule(l){2-9}
 & \multirow{2}{*}{DuQuant} & Uniform & 6.08 & 61.7 & 5.62 & 64.3 & 5.49 & 65.0 \\
 &                              & FPoT & 6.19 & 61.6 & 5.67 & 63.3 & 5.60 & 64.4 \\
\midrule
\multirow{4}{*}{Llama-2-13B}
 & \multirow{2}{*}{OSTQuant}    & Uniform & 5.26 & 65.6 & 5.03 & 67.1 & 4.89 & 67.7 \\
 &                              & FPoT & 5.37 & 65.2 & 5.11 & 66.6 & 5.03 & 67.2 \\
\cmidrule(l){2-9}
 & \multirow{2}{*}{DuQuant} & Uniform & 5.30 & 65.5 & 5.00 & 66.6 & 4.90 & 67.4 \\
 &                              & FPoT & 5.35 & 65.4 & 5.05 & 67.1 & 4.99 & 66.8 \\
\bottomrule
\end{tabular}
\end{table*}

%% file: sections/K_appendix_factorial.tex
\section{Preprocessing \texorpdfstring{$\times$}{x} Quantizer Factorial}
\label{app:factorial}

FPoT is always paired with learned preprocessing, so its gain over PoT could come from the grid itself or from preprocessing trained for that grid. Table~\ref{tab:factorial} separates the two on Llama-3-8B W4A4KV4. The OSTQuant rotation and scales are trained for one of three grids (rows), and GPTQ then quantizes to one of three grids (columns), with three seeds per cell. The weight quantizer dominates. Column means span $1.51$ PPL and $5.51$\%p accuracy, while row means span $0.15$ PPL and $0.97$\%p. The order INT, FPoT, PoT (best to worst) holds in every row for both metrics, so FPoT keeps its advantage over PoT whichever preprocessing it is paired with. On the diagonal, which holds the runs of Table~\ref{tab:main_results}, FPoT recovers $83.4\%$ of the PoT$\to$INT PPL gap and $92.3\%$ of the accuracy gap.

\input{tables/tab_factorial}

%% file: tables/tab_factorial.tex
\begin{table}[ht]
\centering
\small
\setlength{\tabcolsep}{4pt}
\caption{Preprocessing $\times$ weight-quantizer factorial on Llama-3-8B W4A4KV4 (mean$\pm$std over 3 seeds). Rows give the grid that the OSTQuant rotation and scales were trained for, and columns give the grid that GPTQ quantizes to. PPL is WikiText-2 perplexity ($\downarrow$) and 0-shot$^9$ is the mean zero-shot accuracy over nine tasks ($\uparrow$, \%). \textbf{Bold} marks the matched pairs (diagonal), which are the runs of Table~\ref{tab:main_results}. The last column averages each row, and the mean row of each block averages each column.}
\label{tab:factorial}
\begin{tabular}{llcccc}
\toprule
 & & \multicolumn{3}{c}{Weight quantizer} & \\
\cmidrule(lr){3-5}
Metric & Preprocessing target & Uniform & PoT & FPoT (Ours) & mean \\
\midrule
\multirow{4}{*}{PPL${\downarrow}$} & Uniform      & \textbf{7.358}\,{\tiny$\pm$.074} & 9.160\,{\tiny$\pm$.126} & 7.851\,{\tiny$\pm$.039} & 8.123\\
 & PoT          & 7.493\,{\tiny$\pm$.022} & \textbf{8.786}\,{\tiny$\pm$.037} & 7.833\,{\tiny$\pm$.042} & 8.037\\
 & FPoT (Ours)  & 7.437\,{\tiny$\pm$.023} & 8.886\,{\tiny$\pm$.061} & \textbf{7.596}\,{\tiny$\pm$.020} & 7.973\\
\cmidrule{2-6}
 & mean         & 7.430 & 8.944 & 7.760 & \\
\midrule
\multirow{4}{*}{0-shot$^9{\uparrow}$} & Uniform      & \textbf{65.22}\,{\tiny$\pm$.37} & 58.69\,{\tiny$\pm$.60} & 63.16\,{\tiny$\pm$.44} & 62.35\\
 & PoT          & 64.67\,{\tiny$\pm$.31} & \textbf{59.55}\,{\tiny$\pm$.58} & 63.39\,{\tiny$\pm$.44} & 62.54\\
 & FPoT (Ours)  & 65.03\,{\tiny$\pm$.38} & 60.15\,{\tiny$\pm$.91} & \textbf{64.78}\,{\tiny$\pm$.55} & 63.32\\
\cmidrule{2-6}
 & mean         & 64.97 & 59.46 & 63.78 & \\
\bottomrule
\end{tabular}
\end{table}

%% file: sections/L_appendix_generation.tex
\section{Generation and Instruction Following}
\label{app:generation}

Quantization error can accumulate over autoregressive decoding, which perplexity and zero-shot accuracy do not exercise. Table~\ref{tab:generation} evaluates two instruction-tuned models at W4A4KV4 on math, code, instruction following, safety and long-context retrieval. On math and code, standard PoT trails uniform INT by about 15 to 25 points. On Llama-3.1-8B-Instruct its HumanEval score falls from $66.5$ (bf16) to $27.6$, against $52.4$ for uniform INT, and FPoT restores it to $47.8$, recovering $81\%$ of the PoT$\to$INT gap. FPoT recovers $77$--$88\%$ of that gap on Qwen2.5-7B-Instruct math and code. On Llama it also closes the long-context gap ($79.4$ against INT $79.6$ and PoT $61.8$) and $60\%$ of the safety regression PoT introduces. Safety on Qwen is saturated (bf16 attack-success rate $0.00\%$), and on Qwen long-context retrieval FPoT does not improve over PoT.

\input{tables/tab_generation}

\paragraph{Protocol.}
Table~\ref{tab:generation} applies the preprocessing and GPTQ pipeline of the main results to Qwen2.5-7B-Instruct and Llama-3.1-8B-Instruct, with three calibration seeds (0, 42, 123) per quantized method. GSM8K uses the lm-eval-harness \texttt{gsm8k} task (5-shot, flexible-extract exact match, no chat template). IFEval reports prompt-level strict accuracy under the model's chat template. HumanEval prompts the model through its chat template, extracts the code block from the reply, and reports pass@1 against the unit tests. The 520 AdvBench prompts are also issued through the chat template, Llama-Guard-3-8B judges each reply, and the reported value is the fraction judged unsafe (attack-success rate). LongBench uses the English passage-retrieval task (40 samples) with inputs truncated to 10{,}000 tokens for Qwen and 8{,}000 for Llama.

%% file: tables/tab_generation.tex
\begin{table}[t]
\centering
\small
\setlength{\tabcolsep}{4pt}
\caption{Generation and instruction following at W4A4KV4 (mean$\pm$std over 3 seeds; FP (bf16) is the unquantized model). GSM8K exact match, HumanEval pass@1, IFEval prompt-level strict accuracy, AdvBench attack-success rate ($\downarrow$) judged by Llama-Guard-3-8B, and LongBench passage retrieval. \textbf{FPoT (Ours)} rows are bolded.}
\label{tab:generation}
\begin{tabular}{llccccc}
\toprule
Model & Method & GSM8K & HumanEval & IFEval & AdvBench$\downarrow$ & LongBench \\
\midrule
\multirow{4}{*}{Qwen2.5-7B-Instruct} & FP (bf16)   & 81.05 & 82.93 & 69.50 & 0.00 & 87.50 \\
\cmidrule{2-7}
 & Uniform           & 74.15\tiny{$\pm$0.7} & 73.78\tiny{$\pm$3.8} & 64.14\tiny{$\pm$1.1} & 0.58\tiny{$\pm$0.6} & 89.17\tiny{$\pm$2.9} \\
 & PoT               & 53.95\tiny{$\pm$4.4} & 58.94\tiny{$\pm$2.5} & 63.65\tiny{$\pm$1.9} & 0.83\tiny{$\pm$1.0} & 84.17\tiny{$\pm$2.9} \\
 & \textbf{FPoT (Ours)}& \textbf{71.65}\tiny{$\pm$3.1} & \textbf{70.33}\tiny{$\pm$0.7} & \textbf{66.91}\tiny{$\pm$1.0} & \textbf{0.32}\tiny{$\pm$0.1} & \textbf{84.17}\tiny{$\pm$3.8} \\
\midrule
\multirow{4}{*}{Llama-3.1-8B-Instruct} & FP (bf16)   & 78.85 & 66.46 & 72.27 & 3.85 & 82.50 \\
\cmidrule{2-7}
 & Uniform           & 60.78\tiny{$\pm$0.7} & 52.44\tiny{$\pm$1.8} & 68.21\tiny{$\pm$3.0} & 3.72\tiny{$\pm$1.2} & 79.58\tiny{$\pm$4.7} \\
 & PoT               & 44.58\tiny{$\pm$1.6} & 27.64\tiny{$\pm$1.5} & 57.48\tiny{$\pm$2.2} & 9.55\tiny{$\pm$3.4} & 61.77\tiny{$\pm$16.5} \\
 & \textbf{FPoT (Ours)}& \textbf{57.67}\tiny{$\pm$1.9} & \textbf{47.76}\tiny{$\pm$2.0} & \textbf{66.73}\tiny{$\pm$2.2} & \textbf{6.03}\tiny{$\pm$1.1} & \textbf{79.44}\tiny{$\pm$5.9} \\
\bottomrule
\end{tabular}
\end{table}